%% file: bkmshort.tex
\documentclass[twoside,11pt]{article}
\pdfoutput=1
\newif\iffrozen

\usepackage{booktabs}       

\usepackage{comment}
\usepackage{amsmath}
\usepackage{xcolor}
\usepackage{environ} 
\usepackage{tabularx} 
\usepackage{blindtext}

\usepackage{color}
\usepackage{colortbl}
\usepackage{multirow}
\usepackage{enumitem}
\frozenfalse

\newif\ifjmlr
\jmlrfalse

\usepackage{placeins}

\usepackage[
font=small
]{subfig}
\usepackage{graphbox} 

\usepackage{graphicx}
\usepackage{array,calc}
\usepackage[ruled,algonl,vlined]{algorithm2e}
\usepackage{makecell}
\usepackage{relsize}

\usepackage{jmlr2e}
\usepackage{cleveref}
\newcommand{\mimi}{\mkern-4.0mu}

\newcommand{\km}{\texttt{\texorpdfstring{$k$}{k}-means}} 
\newcommand{\Km}{\texttt{\texorpdfstring{$K$}{K}-means}} 
\newcommand{\KM}{\texttt{\texorpdfstring{$K$}{K}-Means}} 

\newcommand{\bkm}{\texttt{breathing \texorpdfstring{$\mimi k$}{k}-means}} 
\newcommand{\Bkm}{\texttt{Breathing \texorpdfstring{$\mimi k$}{k}-means}} 
 
\newcommand{\BKM}{\texttt{Breathing \texorpdfstring{$\mimi K$}{K}-Means}}

\newcommand{\mc}{\mathcal{C}}
\newcommand{\mx}{\mathcal{X}}

\newcommand{\kmp}{\texttt{\texorpdfstring{$k$}{k}-means++}}

\newcommand{\KMp}{\texttt{\texorpdfstring{$K$}{K}-Means++}}
\newcommand{\Kmp}{\texttt{\texorpdfstring{$K$}{K}-means++}}

\newcommand{\vkmp}{\texttt{vanilla \!\texorpdfstring{$k$}{k}-means++}}
\newcommand{\Vkmp}{\texttt{Vanilla\hspace{0.25em}\texorpdfstring{$k$}{k}-means++}}

\newcommand{\hawo}{\texttt{Hartigan-Wong}}

\newcommand{\gkmp}{\texttt{greedy \texorpdfstring{$\mimi k$}{k}-means++}}

\newcommand{\Gkmp}{\texttt{Greedy \texorpdfstring{$\mimi k$}{k}-means++}}
\newcommand{\GKMp}{\texttt{Greedy \texorpdfstring{\!\!$K$}{K}-Means++}}

\newcommand{\bkmp}{\texttt{better \texorpdfstring{$\mimi k$}{k}-means++}}

\newcommand{\Bkmp}{\texttt{Better \texorpdfstring{$\mimi k$}{k}-means++}}

\newcommand{\rs}{\texttt{random swap}}
\newcommand{\Rs}{\texttt{Random swap}}

\newcommand{\Rsc}{Random swap clustering}

\newcommand{\ga}{\texttt{genetic algorithm}}

\newcommand{\KMpr}{K-Means++}

\newcommand{\gla}{\texttt{GLA}}
\newcommand{\GLAS}{\mbox{GLA}}
\newcommand{\mdefault}{5}

\newcommand{\GLA}{\texttt{Generalized Lloyd Algorithm}}
\newcommand{\gLa}{\texttt{generalized Lloyd algorithm}}

\DeclareMathOperator*{\argmax}{arg\,max}
\DeclareMathOperator*{\argmin}{arg\,min} 

\newcommand{\skl}{\texttt{scikit-learn}}

\newcommand{\assign}{=}
\newcommand{\givenx}{$\mx \assign \{x_1,x_2,\dots,x_n\}, x_i \in \mathbb{R}^d$\tcc*[r]{data set}}

\newcommand{\meanx}{$\mc \assign \{c_1\}, \;\mbox{with} \; c_1=\overline{\mx}$\tcc*[r]{codebook consisting only of data set mean}}

\newcommand{\randomc}{$\mc \assign \{c_1,\ldots,c_k\}$\tcc*[r]{random seeding from $\mx$}}

{\begin{list}{}%
		{\setlength{\leftmargin}{#1}}%
		\item[]%
	}%
	{\end{list}}

\ifjmlr
  \jmlrheading{1}{202x}{xx-yy}{xx/24}{mm/yy}{fritzke24a}{Bernd Fritzke}
\fi

\ShortHeadings{Breathing $K$-Means}{Fritzke}

\makeatletter
\let\ftype@table\ftype@figure
\makeatother

\begin{document}

\NewEnviron{NORMAL}{%
    \scalebox{2}{$\BODY$} 
} 
\NewEnviron{LARGER}{%
    \scalebox{1.5}{$\BODY$} 
} 
 
\NewEnviron{HUGE}{%
    \scalebox{5}{$\BODY$} 
} 

\title{Breathing K-Means: Superior K-Means Solutions through Dynamic K-Values}
%

\author{\name Bernd Fritzke \email fritzke@web.de \\
	\addr 61381 Friedrichsdorf\\
    Lindenstr.~4\\
	Germany
}

\ifjmlr
\editor{First1 Last1 and First2 Last2}
\fi

\maketitle

\begin{abstract}%
    We introduce the \bkm{} algorithm, which on average significantly improves solutions obtained by the widely-known \gkmp{} algorithm, the default method for \km{} clustering in the \skl{} package. The improvements are achieved through a novel "breathing" technique, that cyclically increases and decreases the number of centroids based on local error and utility measures. We conducted experiments using \gkmp{} as a baseline, comparing it with \bkm{} and five other \km{} algorithms. Among the methods investigated, only \bkm{} and \bkmp{} consistently outperformed the baseline, with \bkm{} demonstrating a substantial lead. This superior performance was maintained even when comparing the best result of ten runs for all other algorithms to a single run of \bkm, highlighting its effectiveness and speed. Our findings indicate that the \bkm{} algorithm outperforms the other \km{} techniques, especially \gkmp{} with ten repetitions, which it dominates in both solution quality and speed. This positions \bkm{} (with the built-in initialization by a single run of \gkmp) as a superior alternative to running \gkmp{} on its own.
\end{abstract}


	\section{Introduction}
   This section defines the \km{} problem and describes the classic \GLA{} (a.k.a.~\km{} algorithm), the \kmp{} algorithm, and its widely-used variant, \gkmp. 
	\subsection{The \KM{} Problem}
	A common task in data analysis or compression is to describe an extensive data
	set consisting of numeric vectors by a smaller set of representative vectors,
	often called \emph{centroids}. This task is known as the \km{} problem. 
	
	We assume an integer $k$ and a set of $n$ data points $\mx
	\subset \mathbb{R}^d$. The \emph{\km{} problem} is to position a set $\mc =
	\{ c_1,c_2, \ldots\,,c_k\}$ of $k$ $d$-dimensional centroids such that the error
	function

	\begin{equation}
	\phi(\mc,\mx) = \sum_{x\in\mx}
	\min\limits_{c\in\mc} ||x-c||^2 \label{eqn:1}
	\end{equation}
	is minimized. 
We will also refer to $\phi(\mc,\mx)$ as
	\emph{Summed Squared Error} or shortly SSE. In the context of vector
	quantization, the centroid set $\mc$ is called a 
	\emph{codebook}, centroids are referred to as \emph{codebook vectors}, and 
	$\phi(\mc,\mx)$ is denoted as \emph{quantization error}.

    For each centroid $c_i$, one can determine its so-called \emph{Voronoi set},
		which is the  set  $C_i$ of data points for which $c_i$ is the nearest centroid:
		\begin{equation}\label{eqn:voro}
		C_i = \{x \in \mx\,|\; \|x-c_i\| < \|x-c_j\|\, \forall j\neq i\}
		\end{equation} \;

    A necessary but not sufficient condition for a solution $\mc$ to be optimal is the fulfillment of the \emph{centroid condition}: Each centroid $c_i \in \mc$ must be the mean of its Voronoi set $C_i$:
        \begin{equation}\label{eqn:centroid}
        c_i = \frac{1}{|C_i|}\sum_{x\in C_i} x
        \end{equation} \;

        While the term  \emph{centroid} typically refers to the mean of a Voronoi set, we will use it in this article to generally denote a codebook vector, even if is not yet the mean of its Voronoi set. We will also use the term \emph{codebook} instead of \emph{centroid set} for brevity.

	Finding the optimal solution to the \km{} problem is known to be NP-hard
	\citep{Aloise2009}. Therefore, approximation algorithms are used to find a solution with an SSE as low as possible.

	Please note: In this article, we are not concerned with the general clustering problem or whether solutions to the \km{} problem lead to
	``good'' or even ``correct'' clusterings. We also do not require the data to fulfill any pre-conditions or criteria beyond the above definition of the \km{} problem. We are exclusively interested in minimizing the SSE as defined in Equation~\eqref{eqn:1} for a given data set $\mx$ and a given value of $k$. 
	
	\subsection{The Generalized Lloyd Algorithm} \label{sec:gla}

    The \GLA{}, proposed by \cite{Linde1980}, is a multidimensional version of a scalar quantization method initially proposed by John Stuart Lloyd in a 1957 technical report and published 25 years later \citep{Lloyd1982}. It differs from the \km{} algorithm proposed by MacQueen (1967) and described in Section \ref{sec:MacQueen}. Despite common misconceptions, the \GLA{} is not synonymous with 'the' \km{} algorithm, as several \km{} algorithms exist.

Defined in Algorithm~\ref{alg:gla}, the \GLA{} starts with the \emph{seeding} step (the initial codebook choice), followed by repeated \emph{Lloyd iterations} as long as the SSE decreases. Alternatively, it can stop when the relative SSE improvement falls below a certain threshold.

\begin{algorithm}
    \DontPrintSemicolon
    \givenx
    $\mc = \{ c_1,c_2, \ldots\,,c_k\},\, c_i \in \mathbb{R}^d$.\tcc*[r]{seeding} 
    \Repeat 
    ( /* Lloyd Iteration */ )
    {$C$ $\mbox{no longer changes}$}{
        $\bullet\;$ Determine for each centroid $c_i$ its Voronoi set $C_i$.\;
            $C_i = \{x \in \mx\,|\; \|x-c_i\| < \|x-c_j\|\, \forall j\neq i\},\;\; \forall i, i \in \{1, \ldots,\,k\}$
            \; 
        $\bullet\;
        $ Move each centroid $c_i$ to the center of gravity of its Voronoi set:
		$c_i = \frac{1}{|C_i|} \sum\limits_{x\in C_i}x,\;\; \forall i \in \{1, \ldots,\,k\}$ 
    }

    \caption{The \GLA}\label{alg:gla}
\end{algorithm}

The algorithm is proven to converge in finite steps \citep{Selim1984}, but solution quality can vary greatly depending on seeding. Hence, it is common to perform multiple runs with different seedings and select the best result \citep{Franti2019b}.
	
	\subsection{\KMp} \label{sec:kmp}
	\citet{Arthur2007} proposed \kmp, a specific way of seeding the \linebreak \gLa. Centroids are sequentially added by randomly selecting from the data set. The probability of a data point x to be selected is proportional to its quadratic distance to the nearest centroid already in the current codebook (see Algorithm~\ref{alg:kmp}).

    \begin{algorithm}
        \DontPrintSemicolon
        \givenx
        $\mc = \{c_1\}$, with $c_1$ chosen at random from $\mx$\tcc*{start with one centroid}
        \Repeat{$|C| \assign k$}{
            $\bullet\;$ for each $x \in \mx$ let $D(x)$ be the distance of $x$ to the nearest centroid $c \in \mc$\;
            $\bullet\;$ Select a new centroid $q$, choosing $x \in \mx$ with
			probability $P(x) = \frac{D(x)^2}{\sum_{x \in \mx}D(x)^2}$.

            $\bullet\;$ $ \mc \assign \mc \cup \{q\}$\tcc*{add $q$ to the set of centroids}
        }
        $\mc = \mbox{GLA}(\mc,\mx)$ \tcc*{apply the \gLa{}}
        \caption{\Kmp}\label{alg:kmp}
    \end{algorithm}
	
     \citet{Arthur2007} proved the following theorem providing an upper bound for the expected error $E[\phi]$ of a \kmp{} seeding:\\
	
	\setlength\parindent{24pt} \textsc{Theorem.} For any set of data points, $E[\phi] \le 8(\mbox{log}\, k + 2)\phi_{OPT}$\\
	
	\setlength\parindent{0pt}
Thereby, $\phi_{OPT}$ is the error of the optimal solution. 
The theorem provides a significant theoretical improvement over random initialization, which lacks an upper bound for expected error. While seeding is only the algorithm's initial phase, subsequent Lloyd iterations often lead to substantial error reduction. Yet, there is a lack of theoretical evidence quantifying the expected error reduction during this post-seeding phase.
\subsection{\GKMp}\label{sec:gkmp}

\Gkmp, a variant of \kmp{} introduced by \cite{Arthur2007}, draws multiple centroid candidates in each step, choosing the one that maximizes overall error reduction. This reduces the chance of two closely located centroids, which could limit error reduction. Despite reporting improved solution quality, the $\mathcal{O}(\log{}k)$ approximation no longer holds, as confirmed by \cite{DBLP:conf/esa/BhattacharyaER020}.

The Python library, \skl, uses \gkmp{} as its default seeding method, making it the most commonly used seeding approach. It served as our baseline for experimental evaluations. The default number of candidates drawn per step in \skl{} is $\mbox{n\_local\_trials} = 2 + \lfloor \log(k)\rfloor$, resulting for example in 4 for $k=10$, 6 for $k=100$, and 8 for $k=1000$.

	\section{\BKM}\label{sec:bkm}
In this section, we motivate and define the core components of the proposed approach before presenting the complete algorithm.
	\subsection{Algorithm~Outline}

    The \gLa{} is deterministic and only performs local movements of its centroids (by moving them to the center of gravity of their associated data points). This makes this approach very dependent on the initial seeding. To overcome this locality, we added so-called "breathing cycles" consisting of the following steps which are executed after one initial execution of the \gLa:
	
	\begin{enumerate}
		\item Insert $m$ additional centroids (``breathe in'').
		\item Run the \gLa{} on the resulting enlarged codebook of size $k+m$.
		\item Delete $m$ centroids (``breathe out'').
		\item Run the \gLa{} on the resulting  codebook of size $k$.
	\end{enumerate}

    The purpose of a breathing cycle is to position the $m$ additional centroids to minimize the SSE, and subsequently remove $m$ centroids without significantly increasing the SSE thus leading to an improved solution with $k$ centroids. Usually, the removed centroids differ from the added ones, effectively leading to non-local movements of the centroids. 
	
	Because of the periodic changes in codebook size, we refer to the new algorithm as ``\bkm.'' Several questions must be addressed to complete the description of the approach, which are covered in the following sections: 
	\begin{itemize}
		\item Where should new centroids be inserted during the ``breathe in'' step?
		\item Which centroids should be deleted in the ``breathe out'' step?
		\item When should the algorithm terminate?
	\end{itemize}

	\subsection{Breathe In: Adding Centroids Based on High Error}
	An established strategy \citep{Fritzke93, Fritzke95} for minimizing error, regardless of the underlying data distribution, involves adding new centroids near those generating significant errors in quantizing their Voronoi sets.
	Let us denote with $d(x,\mc)$ 
the \emph{quantization error} made for data point $x$ using the
	codebook $\mc$, i.e.,\ the squared distance between $x$ and the nearest centroid in $\mc$, defined as
	
	\begin{equation*}
		d(x,\mc) = \min_{c_i\in \mc} \|x-c_i\|^2.
	\end{equation*}
	
	Given a codebook $\mc$ and a data set $\mx$, we define for each centroid $c_i \in \mc$ its associated error $\phi(c_i)$
	as 
	\begin{equation}
	\label{eqn:Err}
	\phi(c_i) = \sum_{x \in C_i} d(x,\mc),
	\end{equation} 
which is the sum of all $d(x,\mc)$-values over its Voronoi set $C_i$ (defined in Equation~\ref{eqn:voro}).

We can now define the set of those $m$ centroids, which will serve as  anchors for placing new centroids as 
\begin{equation}
\mathcal{M} = (\mbox{the $m$ centroids with the largest associated error $\phi(c)$})
\end{equation}
	
\newcommand{\rmse}{\mbox{RMSE}(\mc,\mx)}
	One new centroid will be inserted near the position of each centroid in $\mathcal{M}$, modified by adding a small random offset vector $v$ to ensure
	distinct centroid values. To be independent of the scaling of the data, 
	we set the length of these offset vectors proportional to the root-mean-square error $\rmse$, defined as 
	
	$$  \rmse = \sqrt{\phi(\mc,\mx) / |\mx|}.
	$$

	Accordingly, we compute each offset vector $v$ as 
	\begin{equation}
	v = \epsilon \rmse u\label{eqn:offset}
	\end{equation}
	with a small constant $\epsilon$ and a random vector $u$ drawn uniformly from the
	$d$-dimensional unit hypercube centered at the origin. This leads to the following set $\mathcal{D}^+$ of new centroids
\begin{equation}
\mathcal{D}^+ = \{c+v|c\in\mathcal{M}\} \;\mbox{with each $v$ being an offset vector according to \eqref{eqn:offset}}.
\end{equation}

The set $\mathcal{D}^+$ is added to the current codebook to finalize the ``breathe in'' step:
\begin{equation}
\mc \leftarrow \mc \cup \mathcal{D}^+
\end{equation}
	
	\subsection{Breathe Out: Removing Centroids Based on Low Utility}\label{sec:breatheout}
	Removing centroids inevitably increases the SSE. To minimize this effect, we select for removal the $m$ centroids causing the smallest
	error increase. Fortunately, the subsequent run of the \gLa{} will lower the resulting SSE again to some degree.
	
	Following \cite{Fritzke1997},  we define the \emph{Utility} $U(c_i)$ of a
	given centroid $c_i$ as 
	
	\begin{equation}\label{eqn:util}
	U(c_i) = \phi(\mc\setminus \{c_i\},\mx) -
	\phi(\mc,\mx).
	\end{equation} 
The utility measures the increase in the overall error caused by removing $c_i$ from the original codebook $\mc$. If this difference is significant, then $c_i$ is considered useful.
	
	Using the definition of $\phi(\mc,\mx)$ in Eq.~(\ref{eqn:1}), the utility of a centroid $c_i$ can be expressed as

	\begin{align}
	U(c_i) & = \sum_{x \in \mx} d(x,\mc\setminus\{c_i\})- d(x,\mc)\nonumber\\
	& = \sum_{x \in C_i} d(x,\mc\setminus\{c_i\})- d(x,\mc)
\;+\; \sum_{x \notin C_i} \underbrace{d(x,\mc\setminus\{c_i\})- d(x,\mc)}_{0} \label{eqn:vororeg}\\
& = \sum_{x \in C_i} d(x,\mc\setminus\{c_i\})- d(x,\mc)\label{eqn:utifinal}.
	\end{align}

The second sum in Equation~\eqref{eqn:vororeg} contains only zero summands since for any $x$ outside the Voronoi region $\mc_i$ the following holds (and makes the terms in the second sum to be zero):
$$\mathlarger{\mathlarger{\forall}}_{x\notin\mc_i} \,\mathlarger{\mathlarger{\exists}}_{j, j\ne i} : x\in\mc_j \land \, d(x,\mc\setminus\{c_i\}) =  \|x-c_j\| = d(x,\mc).$$

 Thus, the
	utility $U(c_i)$ of a centroid $c_i$ only depends on the data points in its Voronoi set $C_i$. Moreover, the utility is always non-negative. This follows from the fact that the expression $d(x,\mc\setminus\{c_i\})- d(x,\mc)$ inside the sum in Equation~\eqref{eqn:utifinal} is non-negative since $d(x,\mc\setminus\{c_i\} \ge d(x,\mc)$.

This expression inside the sum in Equation~\eqref{eqn:utifinal} can be denoted as the utility $U_x(c_i)$ of the centroid $c_i$ for a particular data point $x\in\mx$:
\begin{equation*}
U_x(c_i) = d(x,\mc\setminus\{c_i\})- d(x,\mc)
\end{equation*}
The overall utility can now be expressed as the sum of the utilities of the individual data points in the Voronoi region of $c_i$:
\begin{equation*}
U(c_i) = \sum_{x \in C_i} U_x(c_i)
\end{equation*}

The utility of a centroid, $U_x(c_i)$, only becomes zero when another centroid, $c_j$, is at the same distance from $x$. This happens when $x$ lies on the so-called \emph{bisecting normal hyperplane} of $c_i$ and $c_j$, an event with practically zero probability assuming random positions of $X$ and $C$. Similarly, the complete utility of a centroid $c_i$, $U(c_i)$, becomes zero only when all its associated data points lie on bisecting hyperplanes, another event of virtually zero probability.
Figure~\ref{fig:util_simple} illustrates the error and utility values for a simple \km{} problem.	
\newcommand{\bibi}{0.35}
	\begin{figure}[tb]%
		\centering
		\subfloat[Error values of the centroids]
		{\makebox[\bibi\linewidth][c]
			{{\includegraphics[align=c,width=\bibi\linewidth]{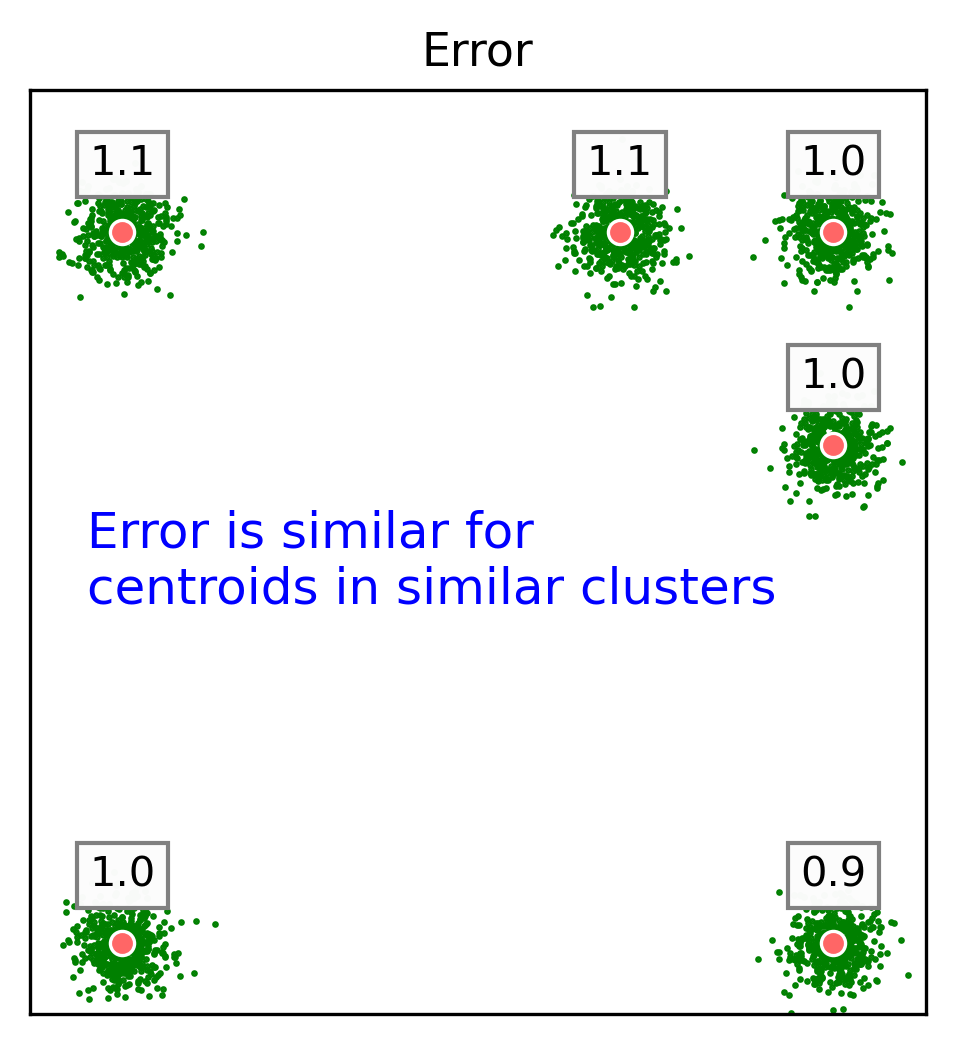} }}}
		~
		\subfloat[Utility values of the centroids]
		{\makebox[\bibi\linewidth][c]
			{{\includegraphics[align=c,width=\bibi\linewidth]{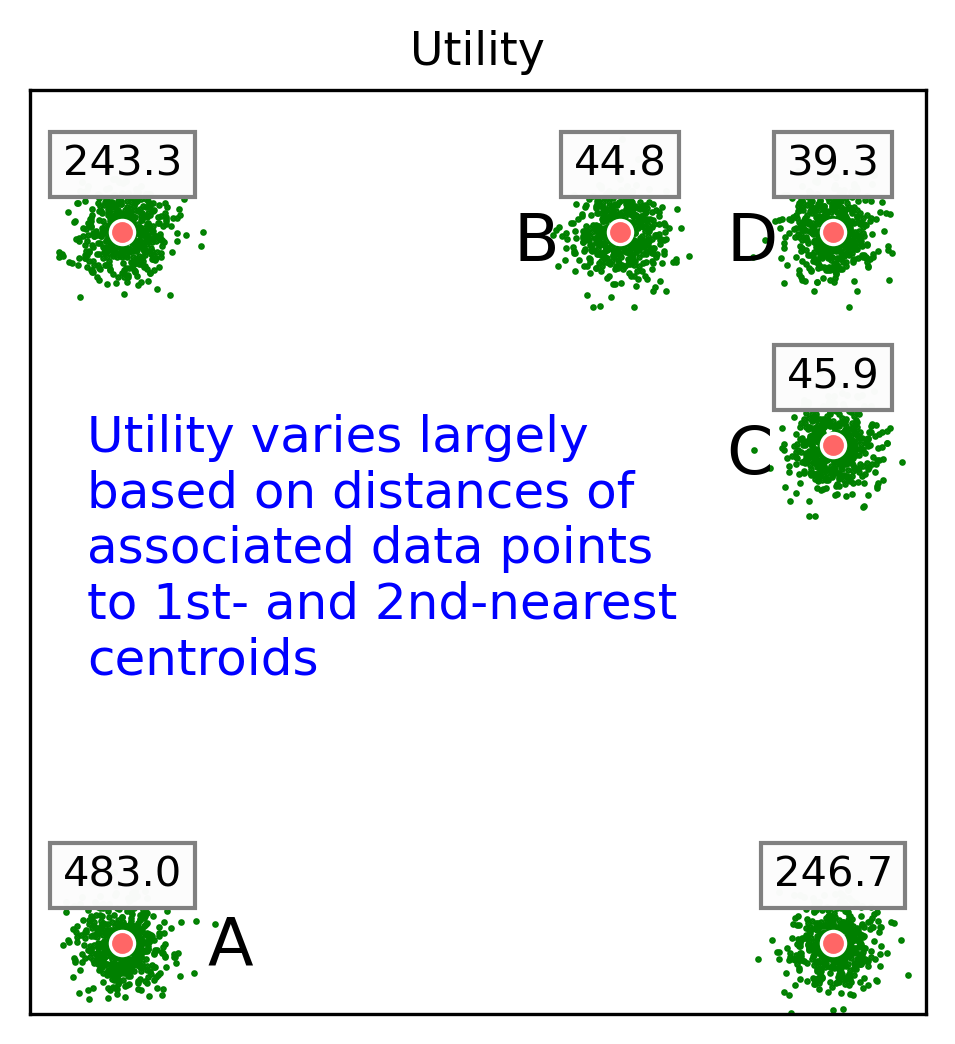} }}}
		\caption{   
Error and utility values are shown for a problem with data from six equal Gaussian kernels and $k=6$, each centroid placed at a cluster center. While error values are similar, the utilities of centroids differ based on the distance between the nearest and second-nearest centroids. The most useful centroid is in cluster A, while the least useful is in cluster D, followed by those in B and C.}
			\label{fig:util_simple}%
	\end{figure}

	To reduce the codebook back to its original size, one might consider deleting the $m$ centroids with the lowest utility values. However, there is a fundamental flaw in  this approach: If the
	distance between two centroids, $c_i$, and $c_j$, is small, also their utility values $U(c_i)$ and $U(c_j)$ are small because they mutually act as the second-nearest centroid for their
	Voronoi sets (see below). Both seem rather ``useless.'' However, removing both $c_i$ and $c_j$ can lead to a colossal error increase, as becomes evident further below.

Let us first calculate what happens to the utility values of two centroids approaching each other:
	\begin{align*}\lim_{c_i\rightarrow c_j} U(c_i) 
& = \lim_{c_i\rightarrow c_j} \sum_{x \in C_i} \underbrace{d(x,\mc\setminus\{c_i\})}_{\le \|x-c_j\|^2}- 
\underbrace{d(x,\mc)}_{\|x-c_i\|^2}\\
& \le \lim_{c_i\rightarrow c_j} \sum_{x \in C_i} \|x-c_j\|^2 - \|x-c_i\|^2 \\
& = 0 \;\;\;\mbox{for all}\, c_i,\,c_j \in \mc, i\ne j
\end{align*}
Since also $U(c_i) \ge 0$ holds and because of symmetry reasons, the following is fulfilled:

\begin{equation*}
\lim_{c_i\rightarrow c_j} U(c_i) = \lim_{c_j\rightarrow c_i} U(c_j) = 0\;\;\;\mbox{for all}\, c_i,\,c_j \in \mc, i\ne j
\end{equation*}

As centroids move closer, their utility values decrease and become zero if they are identical, as they can perfectly substitute for each other in quantizing data points. However, removing such neighboring centroids because of low utility can drastically increase error, especially if the next nearest centroid is far away (see Figure~\ref{fig:utilex}). This often occurs in datasets with isolated smaller clusters, where data points from these close centroids are quantized by a distant centroid, leading to substantial error. To counter this, we introduce a "freezing" mechanism to prevent the concurrent removal of neighboring centroids.
\newcommand{\wii}{0.33}
\begin{figure}[t]
	\centering
	\subfloat[Two neighboring centroids with low utility values (red).]
	{\makebox[\wii\linewidth][c]
		{{\includegraphics[align=c,width=\wii\linewidth]{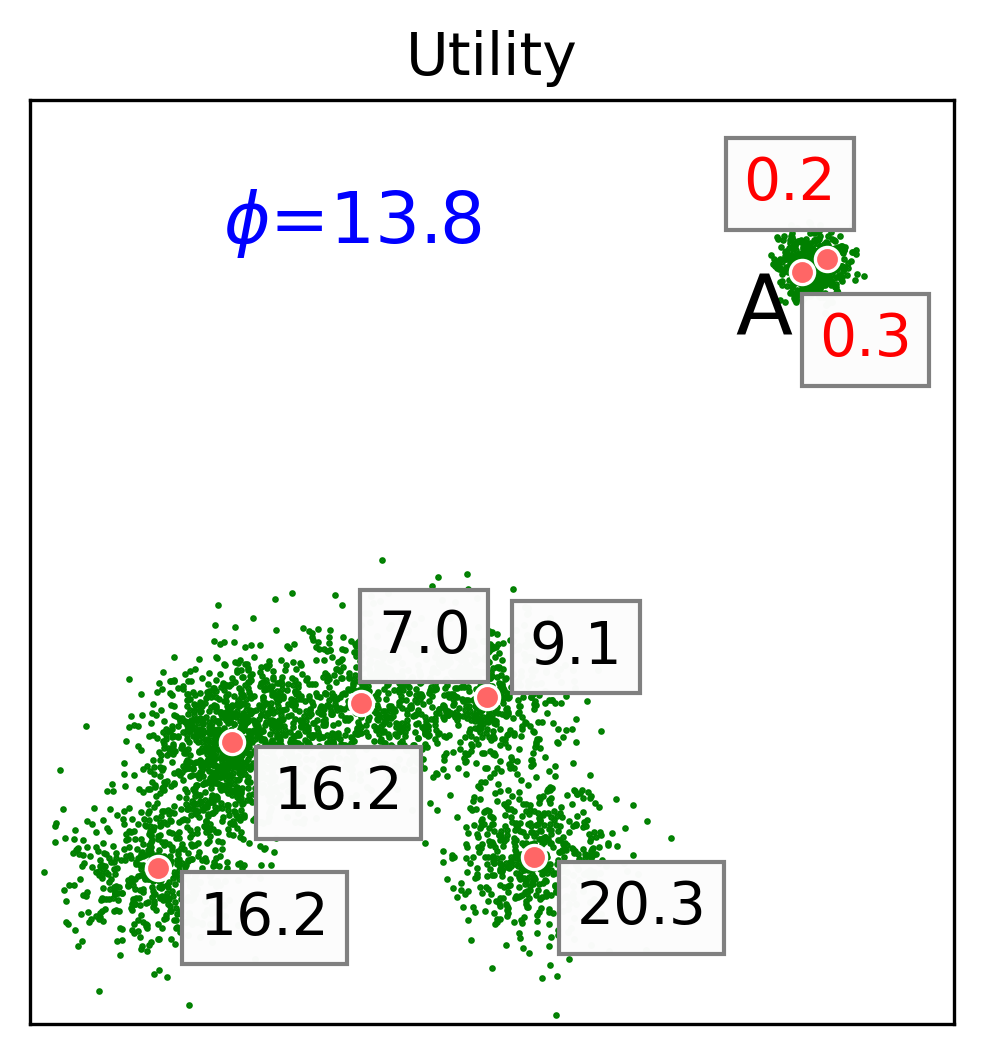} }}}
	~
	\subfloat[Removing one of them makes the other one very useful (red).]
	{\makebox[\wii\linewidth][c]
		{{\includegraphics[align=c,width=\wii\linewidth]{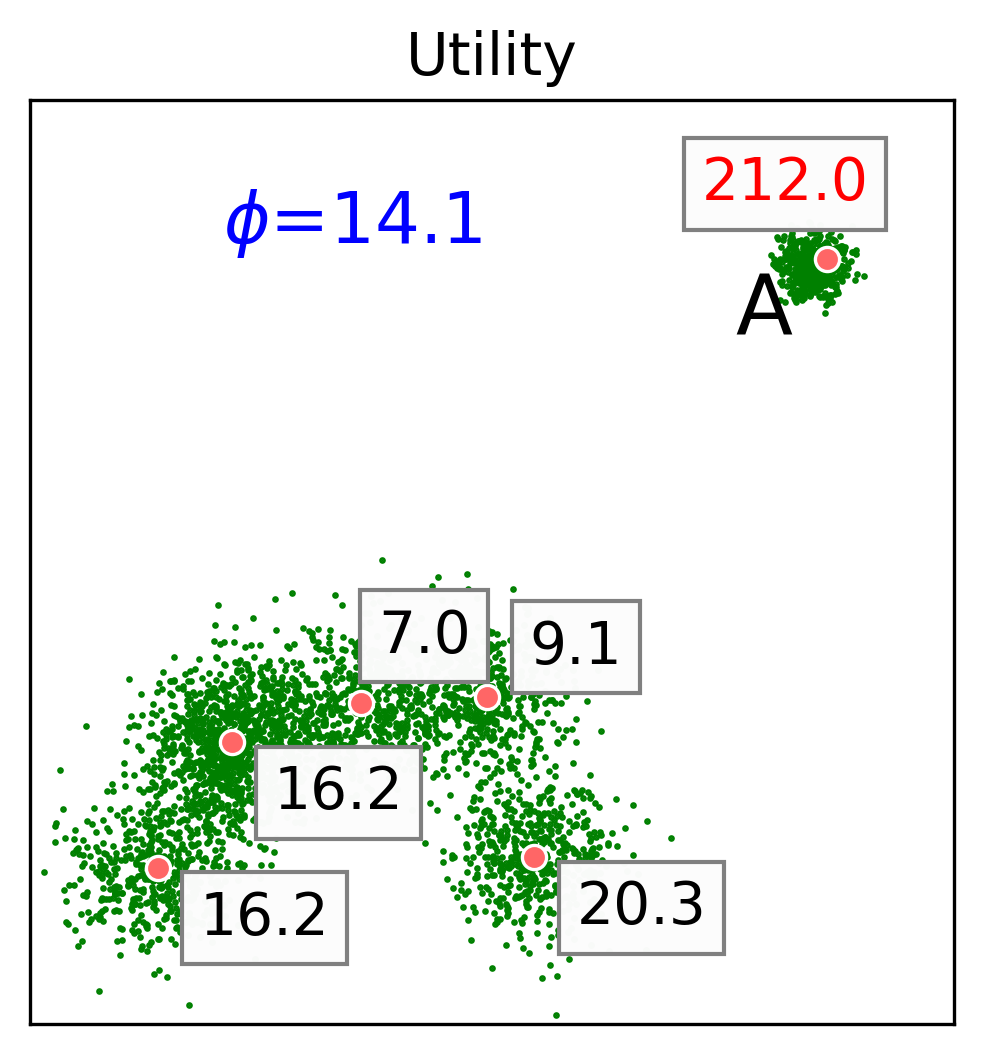} }}}
~
	\subfloat[Removing also the second one causes a huge overall error.]
	{\makebox[\wii\linewidth][c]
		{{\includegraphics[align=c,width=\wii\linewidth]{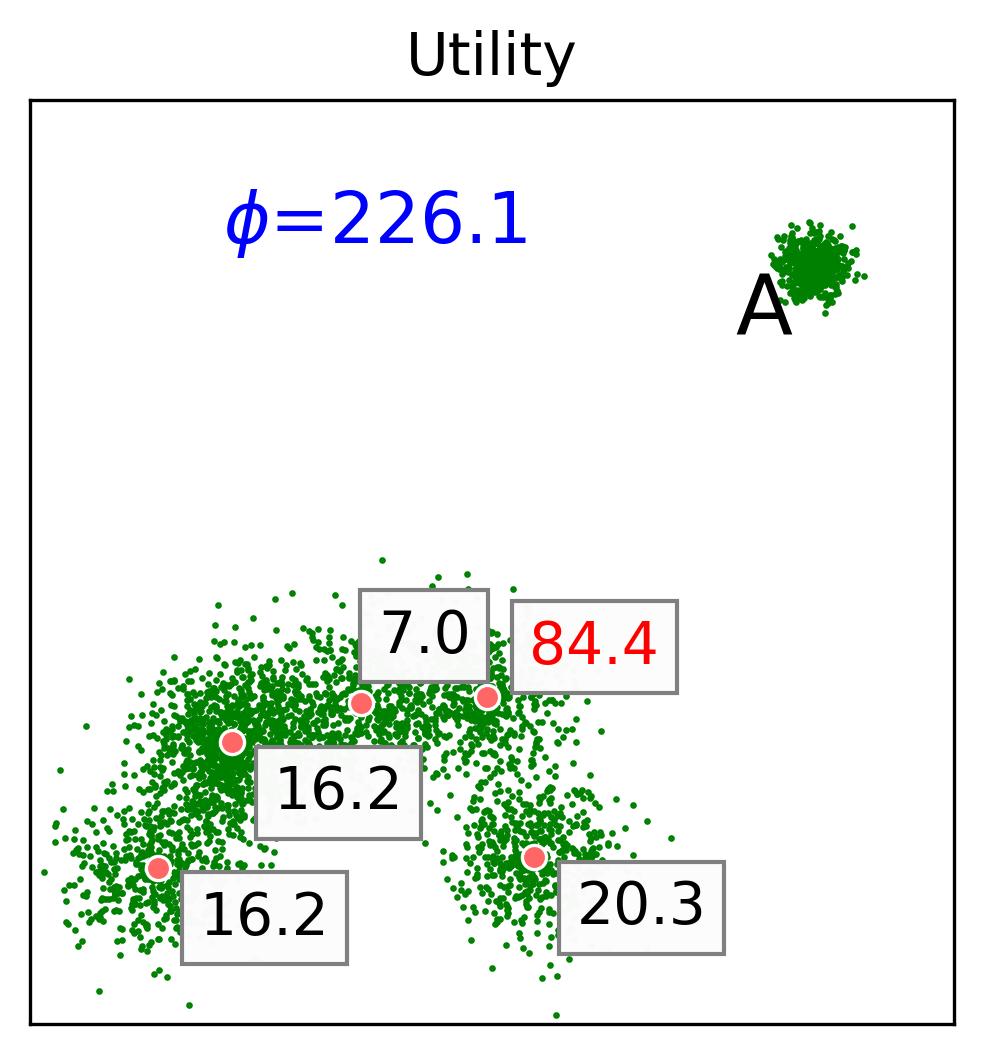} }}}
	\caption{The problem of misleading utility values of close neighbors. a) The two centroids in the small cluster A exhibit low utilities (red). b) Eliminating one of them marginally escalates the error $\phi$, while the remaining one sees its utility spike. c)~The simultaneous removal of the second centroid from A leads to an enormous total error (226.1), and the closest centroid to A becomes highly useful (84.4).
		\label{fig:utilex}}%
    \end{figure}
\subsection{Freezing The Nearest Neighbors}\label{sec:freezing}
	How can we avoid a significant error increase in the ``Breathe out'' step because of the removal
	of neighboring centroids? One possible solution would be to remove one centroid at
	a time, run the \gLa, recompute the utility, remove the next centroid, and so on. This strategy avoids large error increases, though it may come at the cost of high computational demands due to the many required runs of the \gLa . 

To enable the simultaneous removal of multiple centroids, we take the following approach:

\begin{enumerate}
    \item Initialize empty sets for "frozen" centroids ($\mathcal{F}$) and centroids to be removed ($\mathcal{D}^-$).
    \item Rank the centroids by increasing utility.
    \item Scan through the centroids; skip "frozen" ones. Add the first non-frozen centroid to $\mathcal{D}^-$.
    \item After selecting a centroid for removal, add its nearest neighbor to $\mathcal{F}$ (i.e., "freeze" it).
    \item Repeat steps 3 and 4 until $|\mathcal{D}^-|$ equals $m$.
\end{enumerate}

One can construct cases where the above procedure would deliver less than $m$ centroids to remove since too many have been ``frozen.'' To prevent this, we perform freezing (step 4) only as long as the following condition holds:
\begin{equation*}
|\mathcal{F}| + m < |\mc|.
\end{equation*} 
Together with this condition, the above strategy effectively prevents the problematic case of concurrently removing two closely neighboring centroids. 

	\subsection{Ensuring Termination}
	To define a termination criterion, we demand a decrease in error after each ``breathe out'' step (the error after a
	``breathe in'' step is irrelevant because of the enlarged number of centroids). Moreover, we empirically found that once the error stops sinking for a given value of $m$, additional breathing steps with reduced $m$-values  can further lower the error.	
	The above results in the simple approach to guarantee termination shown in Algorithm~\ref{alg:termination}:
    \begin{algorithm}
        \DontPrintSemicolon

        $m \assign m_0$.\tcc*[r]{initialize the breathing size}
        $\phi_{\text{best}} \assign \infty$.\tcc*[r]{initialize the error}
        $tol = $(small positive number, e.g., 0.0001) \tcc*[r]{tolerance for error decrease}
        \Repeat
		( /* breathing cycles */ )
		{
			$m = 0$
		}{
            Perform one breathing cycle with the current $m$.\;
            Compute current error $\phi$.\;    
            \eIf(\tcc*[f]{error improved sufficiently?}){$(\phi_{best}-\phi)/\phi_{best} > \mbox{\text{tol}}$}{
                $\phi_{\text{best}} \assign \phi$.\tcc*[r]{update the best error}
                } {
                $m \assign m - 1$.\tcc*[r]{decrement breathing size}
            }
        }
        \caption{Ensuring Termination}\label{alg:termination}
    \end{algorithm}
	For each value of $m$, breathing cycles are repeated as long as the error $\phi$ strictly decreases, which each time requires finding a previously unseen solution. Since both $m_0$ and the number of partitions of the data into $k$ Voronoi sets are finite and positive, termination occurs in finitely many steps.

	\subsection{The \texorpdfstring{\BKM}{Breathing K-Means}{} Algorithm in Pseudo-Code}
	The complete algorithm in pseudo-code is shown in Figure
	\ref*{alg:bkm}. 

	\begin{algorithm}[thp]
        \givenx
		$m\assign m_0$ (default: \mdefault) \tcc*[r]{number of centroids to add and remove} 
		$k\assign k_0$\tcc*[r]{the $k$ in \km}
		\SetKw{ini}{Seeding:}
		\SetKw{cont}{continue}
		\SetKw{break}{break}
		\DontPrintSemicolon
		$\mc \assign$ (result of \gkmp{} without repetition)\tcc*{seeding}
		$\mbox{tol}\assign\mbox{tol}_0$ (default: 0.0001) \tcc*[r]{tolerance to declare convergence} 		
		$\phi_{best} \assign \phi(\mc,\mx)$ \tcc*[r]{store best error so far}
		$\mc_{best} \assign C$  \tcc*[r]{store best codebook so far}
		
		\Repeat
		( /* breathing cycles */ )
		{
			$m = 0$
		}{
            {\centering \bfseries breathe in\par}

			(Compute error $\phi(c)$ for each $c \in \mc$)
			\tcc*[r]{see Eq.(\ref{eqn:Err})}
			$c_1,\,c_2,\, \ldots,\,c_{m},\,\ldots,\,c_k = \mbox{partial\_sort\_by\_error}(\mc,\mbox{``descending''},m)$\;
			\tcc*[r]{first $m$ centroids sorted}
			
			$\mathcal{M} \assign \{c_1,\,c_2,\, \ldots,\,c_{m}\}$ 
\tcc*[r]{subset of $m$ largest-error centroids}
			$\mathcal{D}^+ = \{c+v|c\in\mathcal{M}\} \;\mbox{with offset vectors $v$   according to Eq.~\eqref{eqn:offset}}$\;

			$\mc \assign \mc \cup \mathcal{D}^+$ 
			\tcc*[r]{insert $m$ additional centroids ("breathe in")}
			$\mc \assign $\gla($\mc,\mx$)
			\tcc*[r]{run the generalized Lloyd algorithm}
            {\centering \bfseries breathe out\par}

			(Compute utility $U(c)$ for each $c \in \mc$.)
			\tcc*[r]{see Eq.(\ref{eqn:util})}
			$c_1, c_2,\, \ldots,\,c_{k+m} = \mbox{sort\_by\_utility}(\mc,\mbox{``ascending''})$
			\tcc*[r]{sorted sequence}			
$\mathcal{D}^-\assign\varnothing$
			\tcc*[r]{initialize set of to-be-deleted centroids}
			$\mathcal{F}\assign\varnothing$
\tcc*[r]{initialize set of frozen centroids}
			\ForAll
			{
				$c \;\mbox{in} \;(c_1, c_2,\, \ldots,\,c_{k+m})$
			}{
				\If (\tcc*[f]{only remove un-frozen centroids}){$c \notin \mathcal{F}$}{

					$\mathcal{D}^-\assign\mathcal{D}^- \cup \{ c\} $
					\tcc*[r]{add centroid to to-be-deleted set}

				\If(\tcc*[f]{not yet too many centroids frozen}){$|\mathcal{F}| + m < |\mc|$}{		
	        $\hat{c} \assign \argmin_{x\in \mc\setminus\{c\}} \|	c-x\|$	
	                    \tcc*[r]{find nearest neighbor $\hat{c}$ of $c$}	                       
						$\mathcal{F}\assign\mathcal{F} \cup \{ \hat{c}\} $ 
						\tcc*[r]{freeze nearest neighbor $\hat{c}$}
					}
					
					\If(\tcc*[f]{found $m$ centroids to delete}){$|\mathcal{D}^-| = m$}{
						\break
					}
				}

			} 
			$\mc\assign\mc\setminus \mathcal{D}^-$
			\tcc*[r]{delete $m$ centroids  ("breathe out")}
			$\mc \assign $\gla($\mc,\mx$)
			\tcc*[r]{run the generalized Lloyd algorithm}
            {\centering \bfseries possibly reduce ``breathing depth''\par}
		
			\eIf{$(\phi_{best}-\phi(\mc,\mx))/\phi_{best} > \mbox{\text{tol}}$}{
				$\phi_{best} \assign \phi(\mc,\mx)$
				\tcc*[r]{improvement:~update best error} 
				$\mc_{best} \assign C$
				\tcc*[r]{update best codebook} 
			}{
				$m\assign m-1$
				\tcc*[r]{no improvement:~reduce "breathing depth"} 
			}
		}
		\Return{$\mc_{best}$}\;
		\caption{The \BKM{} Algorithm}
		
		\label{alg:bkm}	
	\end{algorithm}

\section{Related Work}\label{sec:related}
The literature on algorithms for the \km{} problem is vast and can not be fully surveyed here. In the following,  
we describe two relevant groups of approaches. The first group contains methods for finding a good seeding of the centroids before finally running the \gLa. The second group employs the \gLa{} also in intermediate phases or not at all.
\subsection{Seeding Methods}\label{sec:initialization}
\label{sec:ini}	
	Many methods proposed in the literature focus on finding a seeding used as a starting configuration for the \gLa . 
    Here several relevant examples are described in the order they were historically developed.

\subsubsection{Forgy's Method}
	\cite{Forgy65} randomly assigns each data point to a cluster and then calculates the centroids as the means of these clusters. Consequently, all centroids are typically very close together near the mean of the whole data set, and one can expect a large number of Lloyd iterations before convergence.
 
    \subsubsection{MacQueens First Method}\label{sec:macqueen1}
    In his first method, \cite{MacQueen1967} proposed using the first $k$ elements of the data set $\mx$ as initial centroids. A drawback of this method is that it may initialize all centroids to similar positions in the case of ordered data.

\subsubsection{MacQueens's Second Method}\label{sec:macqueen2}
In his second (and more popular) method, \cite{MacQueen1967}  proposed to pick random elements from the data set $\mx$. This avoids the possible problem of ordered data which his first method has. A drawback of this method is that it may initialize many centroids to similar positions, e.g., if the data set contains a large high-density cluster of data points and a smaller number of spaced-out data points.

\subsubsection{Maximin}	\label{sec:maximin}
	In the Maximin method \citep{Gonzales1985}, the first centroid $c_1$ is chosen arbitrarily. The $i$-th $(i \in {2, 3, . . . , k})$ centroid $c_i$
	is chosen to have the largest minimum distance to all previously selected centroids, i.e., $c_1, c_2, \hdots c_{i-1}$. The method can be seen as a deterministic ancestor of \kmp{} (see sections \ref{sec:kmp} and \ref{sec:kmp2}) and avoids positioning centroids close to each other even if the data contains high-density clusters.
\subsubsection{Method of Bradley and Fayyad}
\newcommand{\bfcm}{M}
\cite{Bradley1998} proposed a method to efficiently produce an initial codebook for large data sets. Initially, $J$ small random sub-samples $S_i, i \in \{1,\ldots,J\}$ are drawn from the original data set $\mc$,  and the \gLa{} is performed on each of the sub-samples $S_i$. Thereafter, the $J$ solutions $M_i, i \in \{1,\ldots,J\}$, are merged to a data set $M$ of size $J\times K$ on which the \gLa{} is run $J$ times with the solutions $M_i$ from the first step as seedings.  From all obtained solutions in the second step, the one with the smallest SSE when encoding $M$ is chosen.
\subsubsection{\KMpr}\label{sec:kmp2}
The \kmp{} algorithm \citep{Arthur2007} is described in detail in Section \ref{sec:kmp} and can be interpreted as a randomized version of the Maximin method (see Section \ref{sec:maximin})  since it uses a point's minimum distance to all previous centroids to set the probability of choosing this point as the next centroid.			
\subsubsection{Greedy \KMpr}			
	\Gkmp{} \citep{Arthur2007} differs from \kmp{} by drawing several new centroid candidates in each step and selecting the one that maximally reduces the overall error (see Section \ref{sec:gkmp}). \Gkmp{} is the default \km{} method for the \skl{} package \citep{scikit}.

\subsubsection{Better \KMpr}\label{sec:betterkmp}
The ``better'' \kmp{} variant \citep{Lattanzi2019} extends the \kmp{} initialization by continuing to select centroid candidates beyond $k$ and possibly replacing existing centroids if there is an improvement (see Algorithm~\ref{alg:betterkmp}). 

\SetKwFor{RepTimes}{repeat}{times}{end}
\begin{algorithm}
    \DontPrintSemicolon
    \givenx
    Initialize codebook with \kmp: $\mc \assign \{x_1,\dots,x_k\}, x_i \in \mx$\:

    \RepTimes{$Z$}{  
        $\bullet$ Select a new centroid candidate $q$, choosing $x \in \mx$ with
        probability $p(x)=\frac{D(x)^2}{\sum_{x \in \mx}D(x)^2}$ whereby $D(x)$ denotes the distance from a data point $x$ to the nearest centroid in $\mc$.
        $D(x) \assign \min_{c \in C} \|x-c\|$\;
        $\bullet$ Compute the minimal error $\phi_{min}$ resulting from replacing one of the centroids in $\mc$ with $q$:
            $\phi_{min} \assign \min_{i \in \{1,\dots,k\}} \phi(\mc\setminus \{c_i\} \cup \{q\} ))$  
        \;
        $\bullet$ \If{
            $\phi_{min} < \phi(\mc,\mx)$
            }{
                Perform the replacement resulting in the minimal error $\phi_{min}$
            }
    }
    return $\mc_{best}$
    \caption{Better K-Means++}\label{alg:betterkmp}
    \end{algorithm}

\cite{Lattanzi2019} proved the following theorem guaranteeing that with a sufficiently large computational budget (parameter Z), the expected cost of the solution produced by \bkmp{} will be close to the optimal cost (within a constant factor).

\begin{theorem} 
    Let $P \subseteq \mathbb{R}^{d}$
be a set of points and C be the
output of Algorithm~1 with $Z \ge 100000k \log \log k$ then we
have $E[cost(P, C)] \in O(cost(P, C^{*}))$, 
where $C^{*}$ is the set of optimum centers. The algorithm's running time is
$O(dnk^2 \log \log k)$.
\end{theorem}
\subsection{Integrated Methods}
The approaches described here 
have in common that they cannot be described as seeding methods for the \gLa. Rather, they perform various operations on the codebook (e.g., splitting, merging, adding, removing, or replacing centroids), and most of them alternate this with Lloyd iterations.

\subsubsection{MacQueen's \KM{}}\label{sec:MacQueen}
\cite{MacQueen1967} proposed an algorithm he called  \km{} (thereby coining the term \km) described as follows (excerpt from the article):
\begin{quote}
Informally, the \km{} procedure consists of simply starting with $k$ groups, each consisting of a single random point, and then adding each new point to the group whose mean the new point is nearest. After a point is added to a group, the mean of that group is adjusted to take the new point into account. Thus at each stage, the $k$ means are, in fact, the means of the groups they represent (hence the term \km{}).
\end{quote}

 This highly efficient algorithm recalculates the mean (centroid) by shifting towards the new point by $\frac{1}{n}$  of total distance upon adding the n-th point to a group. While the means situate at the gravity center of all nearest points when added, a full data sweep does not always ensure the centroid condition, implying each mean is not necessarily at the gravity center of its Voronoi set. As the centroid condition is key for optimality, additional Lloyd iterations often enhance MacQueen's \km{} solutions, even if only towards a local optimum.

\subsubsection{The Hartigan-Wong Algorithm}\label{sec:hawo}
The \hawo{} Algorithm \citep{Hartigan1979} skips Lloyd iterations. It starts by randomly choosing $k$ centroids, forming initial clusters using MacQueen’s Second Method (Section \ref{sec:macqueen2}). It then reassigns a random data point $x$ from its cluster $S$ to another cluster $T$ if it reduces the sum of intra-cluster variances of $S$ and $T$, choosing $T$ to maximize variance reduction. Termination occurs when no reassignment reduces overall variance (Algorithm~\ref{alg:hartigan}).

Hartigan and Wong's implementation introduces a \emph{Quick Transfer} phase, where $T$ is the centroid second-nearest to $x$, reducing computation. This phase iterates until no improvement occurs. The \emph{Optimal Transfer} phase, involving a complete search among all clusters (Algorithm 6), alternates with the Quick Transfer phase. Termination occurs when the Optimal Transfer phase finds no improvement.

This algorithm is the default \km{} algorithm in the \texttt{stats} package of R \citep{RCore2019}. While \cite{TelgarskyVattani2010} reported improvements over "online" \km{}, we generally found better results with \vkmp{} and \gkmp{} than \hawo{} (see Table \ref{tab:msee-high-D}).

\begin{algorithm}
    \DontPrintSemicolon
    \givenx
    Select $k$ centroids $c_1, c_2, \ldots, c_k$ randomly from $\mx$.\;
    Form  $k$ clusters $C_1, C_2, \ldots, C_k$ by assigníng each $x\in\mx$ to its nearest centroid.\;
    Recompute centroids $c_1, c_2, \ldots, c_k$ as means of their associated clusters.\;
    \Repeat{no error improvement for a complete sweep through all data points $x \in \mx$}{
        Select a data point $x \in \mx$.\;
        Let $S \in \{C_1, C_2, \ldots, C_k\}$ be the cluster to which $x$ is currently assigned.\;
        \ForAll{$T \in \{C_1, C_2, \ldots, C_k\}, T \ne S$}{
            Compute the error improvement $\Phi(x;S;T)$ of re-assigning $x$ from $S$ to $T$ taking into account the resulting centroid updates of $S$ and $T$.\;
            $\Phi(x;S;T) = \frac{|S|}{|S|-1} \|\mu(S)-x\|^2 - \frac{|T|}{|T|+1} \|\mu(T)-x\|^2$
        }
        \If{$\exists T \in \{C_1, C_2, \ldots, C_k\}, T \ne S| \Phi(x;S;T) > 0$}{
            Re-assign $x$ to cluster $T$ with $T = \argmax_T \Phi(x;S;T)$.\;
        }
    }
    \caption{Hartigan and Wong}\label{alg:hartigan}
\end{algorithm}

\subsubsection{LBG with Binary Splitting}				

	When the \gLa{} (a.k.a.~LBG) was proposed by \cite{Linde1980}, the authors also discussed a method to produce a series of increasingly large codebooks. In particular, given a codebook consisting of $m$ centroids, one can produce a codebook consisting of twice as many centroids by ``splitting'' each centroid, adding small offsets to enforce distinct values, and applying the \gLa{} to the resulting enlarged codebook. If one starts with a codebook of size one and performs $p$ splitting steps, the resulting codebook has the size $k=2^{p}$. 
    In each splitting step, all existing centroids are split. Thus, this method does not consider which centroids are
most suited for splitting to reduce the overall error. This can limit the quality of the results compared to approaches splitting based on error reduction. 
\subsubsection{LBG-U} \label{sec:lbgu}
\cite{Fritzke1997} proposed the \emph{LBG-U} algorithm to improve the \texttt{generalized Lloyd}{} \linebreak\texttt{algorithm} by non-local movements of centroids. Central to this approach is the concept of \emph{Utility} (thus the ``U'' in the name) initially defined in that work and also used for \bkm{} (see Equation~\ref{eqn:util}).  The core mechanism of LBG-U is to repeatedly move the least useful centroid to the centroid with maximum error and perform the \gLa{} after each such move. LBG-U delivered better solutions than the \gLa{} (\kmp{} was not yet invented) at the price of additional compute time. Lacking the idea of nearest neighbor freezing introduced in the current article, LBG-U could only insert and delete one centroid at a time which led to larger computational effort and smaller improvements than \bkm.

\subsubsection{Splitting}\label{sec:split}
This algorithm \citep{Franti1997} starts with a codebook of size one and iteratively enlarges the codebook by a splitting procedure until it reaches size $k$. Different approaches for selecting the cluster to be split (largest variance, largest width, largest skewness) and for technically performing the splitting (fixed offset vector, random choice of new centroids, mutually furthest data points, local PCA) are discussed. Also the question of how to refine the partition of the data set and the centroids after the splitting is addressed.

\subsubsection{Tabu Search}
\cite{Franti1998} proposed an algorithm adapted from a previous clustering method by \cite{AlSultan1995}. Tabu search generates new solution candidates through random operations, allowing for potentially worse solutions and possible cyclic behavior. To prevent non-termination, the algorithm employs a \emph{tabu list}—a record of previously visited solutions that helps avoid considering them multiple times.
For larger data sets, the exact recurrence of \km{} solutions is rare. To exclude \emph{similar}  solutions as well, the authors propose checking candidates for a minimum distance from all tabu list elements using a suitable distance measure. Two methods for the randomized generation of new solutions are considered: (1) randomly assigning a fraction of the data set to different (but nearby) clusters and (2) adding noise to existing cluster centers.

\subsubsection{Iterative Splitting and Merging}\label{sec:splitmerge}
\cite{Kaukoranta1998} describe an iterative splitting and merging algorithm for vector quantization codebook generation. 
Repeatedly the following steps are performed: (1) a cluster is selected which is split. (2) Two clusters are selected which are merged. (3) Some Lloyd iterations are performed to refine the codebook.
For (1) a local optimization strategy is applied where each cluster is tentatively split, and the one decreasing the
distortion most is chosen. For (2) the pairwise nearest neighbor (PNN) approach as described by \cite{Equitz1989} is employed. This approach determines the neighboring pair of clusters, $mc_i$ and $\mc_k$, which least increases the overall error when merged. For (3) the authors suggest performing a fixed small number (e.g., 2) of Lloyd iterations.

This approach is somewhat similar to \bkm{} with breathing depth $m=1$. 
\Bkm, however, avoids the effort to split each cluster by splitting only the cluster with the largest quantization error. Moreover, \bkm{} avoids the computation of many possible merges by always deleting the centroid with the smallest utility. 

\subsubsection{Bisecting K-Means}				

The \emph{bisecting \km}{} \citep{steinbach2000} has large similarities to the splitting method proposed by \cite{Franti1997}. It starts with one single cluster and iteratively ``bisects'' (i.e., ``splits'') one of the present clusters into two by performing \km{} with $k=2$ on the selected cluster. This bisection step is repeated several times (say $m$ times) with different random initializations before choosing the bisection with the lowest error. This is iterated until a predefined number of clusters is reached 
or the overall error falls below a threshold.
Optionally, the \gLa{} can be applied to the resulting codebook after each bisecting step.
If this optimization is \emph{not} done, the algorithm produces a hierarchical clustering (obtained by considering all intermediate codebooks).
To select the cluster to be split, \cite{steinbach2000} propose to use either the size of the cluster or the SSE of the cluster as a criterion. In the latter case, there is a similarity to the error-based insertion proposed by \cite{Fritzke93, Fritzke94c}.

\subsubsection{Genetic Algorithm with Deterministic Cross-Over}\label{sec:ga}
Genetic algorithms typically make use of a condensed ``genetic'' representation of the candidate solutions. They do attempt to simulate natural evolution by employing concepts such as selection (survival of the fittest), cross-over (recombination of several different genetic representations), and mutation (random modifications of genetic representations). Here we consider the genetic algorithm described by \cite{Franti2000}, which is specifically adapted to the problem of vector quantization.

The top-level algorithm is shown in Algorithm~\ref{alg:ga} and does not show any problem-specific properties apart from the use of the \gLa{} for fine-tuning.  

\begin{algorithm}
    Generate S initial solutions.\;
    Sort the solutions by error.\;
    \RepTimes(/* T = number of generations */){T}{
        \RepTimes(/* S = number of individuals per generation */){S}{
            Select two solutions for cross-over.\;
            Generate a new solution by crossing the selected solutions.\;
            Optionally mutate the new solution.\;
            Fine-tune the new solution by \gLa.\;
        }
        Sort the solutions by error.\;
        Store the best solution.\;
    }
    \Return{best solution}
    \caption{Genetic Algorithm}\label{alg:ga}
\end{algorithm}

The problem-specific properties proposed by \cite{Franti2000} are the representation of a solution and the cross-over operation. Each solution is represented as a pair $(C,P)$ where $C$ is a codebook, and $P$ is a corresponding partition of the data set. Maintaining both types of information makes it possible to perform the cross-over operation of two solutions $(C^1,P^1)$ and $(C^2,P^2)$ in a very effective and efficient way:
\begin{itemize}
    \item A new codebook $C^{new}$ is created as the union of $C^1$ and $C^2$: \newline $C^{new} = C^1 \cup C^2$.
    \item The partitions $P^1$ and  $P^2$ are combined to form a new partition $P^{new}$ by mapping each data point to its cluster from $P^1$ or $P^2$, depending on which corresponding centroid is closer.
    \item $C^{new}$ is updated to contain the centroids of the clusters in $P^{new}$.
    \item Empty clusters are removed from $P^{new}$.
    \item The number of clusters in $P^{new}$ is reduced to the desired number $k$ of clusters by using the pairwise-nearest-neighbor (PNN) approach on ($C^{new}$,$P^{new}$). PNN is performed on the $2\times k$ (fewer if empty clusters were removed) centroids instead of the full data set. The partition is updated accordingly by combining merged clusters.
\end{itemize}
Mutation was considered but not performed in \cite{Franti2000} because of the concentration on efficiency.
\subsubsection{Global K-Means}				
``Global \km'' \citep{Likas2003} is a deterministic method that finds an approximate solution for a given \km{} problem ($\mx,k$) by starting with the trivial solution for a codebook size of one. This solution is used to find a solution for codebook size two by combining the size-one solution sequentially with each element of the data set $\mx$ and running the \gLa{} starting from there. The best solution found is taken as the solution for size two. This is iterated for all codebook sizes until a solution for codebook size $k$ is found. The algorithm requires $k \times n$ runs of the \gLa{} leading to very high computation demand for data sets of non-trivial size. 

\subsubsection{Iterative Shrinking}

\cite{Franti2006}  describe an iterative shrinking algorithm for vector quantization codebook generation. The method starts by assigning each data vector to its own cluster. This huge codebook is then stepwise reduced by deleting the single centroid leading to the smallest error increase. The major difference between the ``shrinking'' described in this algorithm and the ``merging'' described in \cite{Kaukoranta1998} (see section \ref{sec:splitmerge}) is the following: During ``shrinking," a centroid $c_i$ is removed, and each associated data point $x \in \mc_i$ is assigned to the respective nearest other centroid $c \in \mc \setminus \{c_i\}$, whereas during ``merging," two neighboring clusters, $\mc_i$ and $\mc_k$ are combined into a new cluster $\mc_m$, i.e., all affected data points end up in the same cluster $\mc_m$ (before any further optimizations, e.g.,  Lloyd iterations, are done.). 

\subsubsection{Random Swap Clustering}\label{sec:rs}

\Rsc{} \citep{Franti2018} is based on the idea of repeatedly replacing a randomly chosen centroid $c \in \mc$ with a randomly chosen data vector $x\in \mx$. This operation, also called ``swap," is followed by a small number of Lloyd iterations. If the resulting error is lower than before the swap, the swap is ``accepted," and the algorithm continues. If the resulting error is higher than before the swap, the algorithm continues from the codebook state before the swap (basically ignoring the
 swap). The algorithm is shown in Algorithm~\ref{alg:swap}.  

 This number of required Lloyd operations after a swap operation can be reduced by locally repartitioning the data points associated with the deleted centroid and by specifically determining which data points will be assigned to the new centroid. This is merely an efficiency measure and is denoted as optional by \cite{Franti2018} if two or more Lloyd iterations are performed.
\begin{algorithm}
    \DontPrintSemicolon
    \givenx
    \randomc
    \RepTimes{$Z$}{ 
        Randomly select a centroid $c,\,c \in \mc$.\;
        Randomly select a data vector $x \in \mx$.\;
        $\mc_{new}=\mc \cup \{x\} \setminus \{c\}$\tcc*[r]{replace $c$ with $x$}
        Optional: Locally repartition data points.\tcc*{see text}
        Perform a few Lloyd iterations on $\mc_{new}$.\;
        \If {$\phi(\mc_{new},\mx) < \phi(\mc,\mx)$\tcc*[r]{is the new codebook better?}}{$\mc \assign \mc_{new}$ \tcc*[r]{Accept the new codebook.}}
    }
    \Return{$\mc$}
    \caption{Random Swap Clustering}\label{alg:swap}
    \end{algorithm}

\subsubsection{Improvement by Multiple Runs}\label{sec:multirun}	
	
	A general method to improve results for any randomized algorithm is selecting the best result from multiple repeated runs \citep{Franti2019b}. The improvements obtained depend on the variance of the results produced by the algorithm at hand. The required amount of computation is proportional to the number of repetitions.

\section{Algorithms Selected for Comparison}\label{sec:algorithms}
    The following algorithms were selected as contenders for \bkm{} (a brief reasoning is given for the inclusion of each approach):

    \begin{description}[style=nextline]
    
    \item [\gkmp] (see Section \ref{sec:gkmp}). This method was selected as the baseline algorithm for all other methods since it probably is the most widely-used algorithm for \km{} because of being the default \km{} method in the popular \skl{} package \citep{scikit}.  The implementation is the class \texttt{KMeans} of \skl.
    Sources:  \url{https://scikit-learn.org/stable/modules/generated/sklearn.cluster.KMeans.html}
    
    \item [\vkmp] (see Section  \ref{sec:kmp}). The original (non-greedy) variant of \kmp. For this method, an $\mathcal{O}(\log{}k)$ upper bound was proven. The implementation stems from the \skl{} package \citep{scikit}, in particular from the function \texttt{kmeans\_plusplus} with default parameters, except for setting \texttt{n\_local\_trials=1}. 
    Sources:  \url{https://scikit-learn.org/stable/modules/generated/sklearn.cluster.kmeans_plusplus.html}
    
    \item [\bkmp] (see Section \ref{sec:betterkmp}). A modification of \kmp{} that was selected after being independently suggested by two reviewers. The algorithm was implemented by the author in Python since no open-source (or other) implementation could be found. The parameter $Z$ for the number of additional centroid selections was set to $25$, following the experiments described in the original paper of \cite{Lattanzi2019}.
    
    \item [\hawo] (see Section~\ref{sec:hawo}). This method was selected since it is the default \km{} algorithm in the \texttt{stats} package of the R programming language \citep{RCore2019}. The approach is special since it does not rely on the \GLA. Parameter settings: \texttt{iter.max=500, nstart=1, algorithm="Hartigan-Wong"}. Implementation: Fortran code from the \texttt{stats} package of R.
    Sources: \url{https://cloud.r-project.org}
    
    \item [\ga] (see Section \ref{sec:ga}). We included this approach following a reviewer's suggestion and promising results in a pre-study. The implementation used is part of a C-Package published on github by the authors of the original publication \citep{Franti2000}. The default parameters defined by the package were used in the experiments.
    Sources: \url{https://github.com/uef-machine-learning/CBModules}
    
    \item [\rs] (see Section~\ref{sec:rs}). We included this approach following a reviewer's suggestion and promising results in a pre-study. The implementation used is part of a C-Package published on github by the authors of the original publication \citep{Franti2018}. The default parameters defined by the package were used in the experiments.
    Sources: \url{https://github.com/uef-machine-learning/CBModules}

    \end{description}

\section{Empirical Results}\label{sec:empirical}
\setlength{\fboxsep}{3pt}
We first list the algorithms selected for comparison with \bkm. Then, the \km{} problems used for the experiments are described. Finally, the empirical results are presented.

\subsection{\KM{} Problems Investigated}\label[]{sec:problems}
We used four groups of two-dimensional problems (9 problems per group) and one group of high-dimensional problems (15 members) for a total of 51 \km{} problems. These groups were selected to showcase the algorithms' strengths and weaknesses.

The problems included varied point densities, reflecting the diversity of real datasets, unlike the mixtures of identically shaped Gaussians found in many textbook examples. Of the 51 problems, 24 are from the literature, and 27 are self-generated. The following subsections describe the problems in detail.

\subsubsection{Problems with known optimum}
Each problem in this group is constructed with a known optimal solution. The datasets consist of $g$ identical, well-separated \emph{macro-blocks}, each comprising $b$ adjacent quadratic base blocks of data points, where $b$ can be 1, 3, or 4. The number of centroids $k$ is set to $k = g \times b$. The optimal solution places one centroid at the center of each base block, as shown in Figure \ref{fig:prob-opt} with optimal solutions in red.

The optimality is justified because the macro-blocks are identical and well-separated, ensuring an optimal solution distributes centroids evenly among them. Thus, the overall solution's optimality equates to that of an individual macro-block.

For $b = 1$, the optimal solution is the centroid of the single block, thus satisfying the centroid condition. For $b = 3$ or $b = 4$, it is assumed that the optimal partial solution involves placing one centroid in each base block's center. Hence, the optimal solution for the entire problem is to center one centroid in each base block across all macro-blocks.

%
%
\newcommand{\capProbLit}{...}
\newcommand{\capProbOptDetail}{...}
\newcommand{\capProbOpt}{...}
\newcommand{\capProbLitmod}{...}
\newcommand{\capProbEspiral}{...}
\newcommand{\capProbHighD}{...}

\renewcommand{\capProbOpt}{Problems with known optimum. The data points are shown in green, and the optimal centroids are in red.}
\newcommand{\fiwid}{1.0}
\renewcommand{\fiwid}{0.76}

\input{img/prob-opt.tex}

\subsubsection{Literature Problems}\label{sec:prob-lit}
The problems in this group are based on data sets from the literature. The $k$-values were freely chosen, resulting in the problems listed in Table \ref{tab:datalit} and displayed in Figure~\ref{fig:prob-lit}. 
\begin{table}
    \begin{center}
    \begin{tabular}{llrr}
    \toprule
    data set&  origin &$n$ & $k$\\ 
    \midrule
    ``Aggregation''&  \cite{Gionis2007} &788& 200 \\
    ``Compound``&  \cite{Zahn1971} & 399 & 50 \\
    ``D31''&  \cite{Veenman2002} &3100& 100 \\
    ``Flame''&  \cite{Fu2007} &240& 80 \\
    ``Jain''&  \cite{Jain2005} &373& 30 \\
    ``R15''&  \cite{Veenman2002} &600& 30 \\
    ``S2''&  \cite{Franti2006} &5000& 100 \\
    ``Spiral''&  \cite{Chang2008} &312& 80 \\
    ``Pathbased''&  \cite{Chang2008} &312& 80 \\
    \bottomrule
    \end{tabular}
    \end{center}
        \caption[short caption]{Two-dimensional data sets from the literature with chosen $k$-values. All data sets were obtained from \url{http://cs.joensuu.fi/sipu/datasets/}.}
        \label{tab:datalit}
    \end{table}

\renewcommand{\capProbLit}{\Km{} problems based on two-dimensional data sets from the literature (see Table \ref{tab:datalit}).}
\input{img/prob-lit.tex}

\subsubsection{Modified Literature Problems.}\label{sec:mod-prob-lit}
The problems in this group were generated from the ``literature problems'' in the previous section as follows.
\begin{itemize}
\item Randomly select 200 data points from the literature problem.
\item Add a very dense cluster of 4000 data points below the area occupied by the selected 200 data points.
\end{itemize}
The purpose of this modification is to test the algorithms' ability to deal with data sets having a large variation in density. 

\renewcommand{\capProbLitmod}{Modified literature problems. The data sets were constructed from the problems shown in Figure~\ref{fig:prob-lit} by taking a random subset of size 200 and adding a high-density cluster consisting of 4000 data points below the centroid of the subset. The $k$-values remain the same as in Table \ref{tab:datalit}.}

\input{img/prob-litmod.tex}
\subsubsection{``Evil Spiral'' Problems}
This group of problems (see Figure~\ref{fig:prob-espiral}) was designed to check the effect of letting an increasing fraction of the data originate from high-density clusters. One common part of all data sets consists of 500 data points located in 25 Gaussian clusters of 20 points, each positioned on a spiral. The other part of the data sets consists of one or more high-density Gaussian clusters arranged on a second spiral intertwined with the first one. The number of high-density clusters, each consisting of 500 data points, varies from one to 25 in steps of three.
The value of $k$ is always set to 100.

\renewcommand{\capProbEspiral}{``Evil Spiral'' problems. Each data set contains the same 25 wide Gaussian clusters (20 points each) arranged in a spiral (having $20\times25=500$ points) and an increasing number ($1,4,\ldots,25$) of very dense clusters arranged on a spiral intertwined with the first spiral. Each of the dense clusters consists of 500 points, i.e., as many as the complete first spiral. For all problems, $k=100$ is used.}

\input{img/prob-espiral.tex}
\subsubsection{High-Dimensional Problems}\label{sec:prob-highD}

The problems in this group are based on three data sets (see Figure~\ref{fig:prob-highD}) used in the original paper on \kmp{} \citep{Arthur2007} where the $k$-values were chosen from $k \in \{10,\,25,\,50\}$. We added two larger $k$-values resulting in $k \in \{10,\,25,\,50,\,100,\,200\}$. The data sets are described below. \newline

\textbf{Norm25:}
This data set consists of $n=10000$ vectors of dimension $d=15$. The original data used by \cite{Arthur2007} is not publicly available anymore, but their paper contains the following description (and also states that the number of data points is 10000):
 \begin{quote}``The first data set, \emph{Norm25}, is synthetic. 25 “true” centers were drawn uniformly at random
from a 15-dimensional hypercube of side length 500.
Then points from Gaussian distributions of variance 1 around each true center were added, resulting in 
25 well-separated Gaussians with the true centers providing a close approximation to the optimal
clustering.''
 \end{quote}

 Using this information, we generated a new data set, \emph{Norm25}, with statistical properties similar to those of the original one.\newline

 \textbf{Cloud:}
 This data set consists of $n=1024$ vectors of dimension $d=10$. It is the \emph{Cloud} data set from the UCI Machine Learning Repository \citep{Dua:2019} and is available at \url{https://archive.ics.uci.edu/ml/datasets/Cloud}. The data was derived from two $512 \times 512$-pixel satellite images of clouds (one image in the visible spectrum and one in the infrared spectrum) taken with an AVHRR (Advanced Very High-Resolution Radiometer) sensor. The images were divided into 1024 super-pixels of size $16\times 16$, and from each pair of super-pixels, ten numerical features were extracted to form the final data set.\newline

 \textbf{Spam:}
 This data set consists of $n=4601$ vectors of dimension $d=58$. It is the \emph{Spam} data set from the UCI Machine Learning Repository \citep{Dua:2019} and is available at \url{https://archive.ics.uci.edu/ml/datasets/Spambase}. According to the data set description, the data was generated from spam and non-spam emails. Most of the features (48 of 58) are word frequencies from different words. Other features measure the occurrence frequencies of certain characters or capital letters.

\renewcommand{\capProbHighD}{Three high-dimensional data sets which were also used by \cite{Arthur2007}. Descriptions are in the text. The $k$-values used with each data set were 10, 25, 50, 100, and 200, resulting in 15 different problems. The figures display projections of the high-dimensional data onto two selected axes.}

\begin{figure}
	\centering
	\includegraphics[width=\fiwid\linewidth]{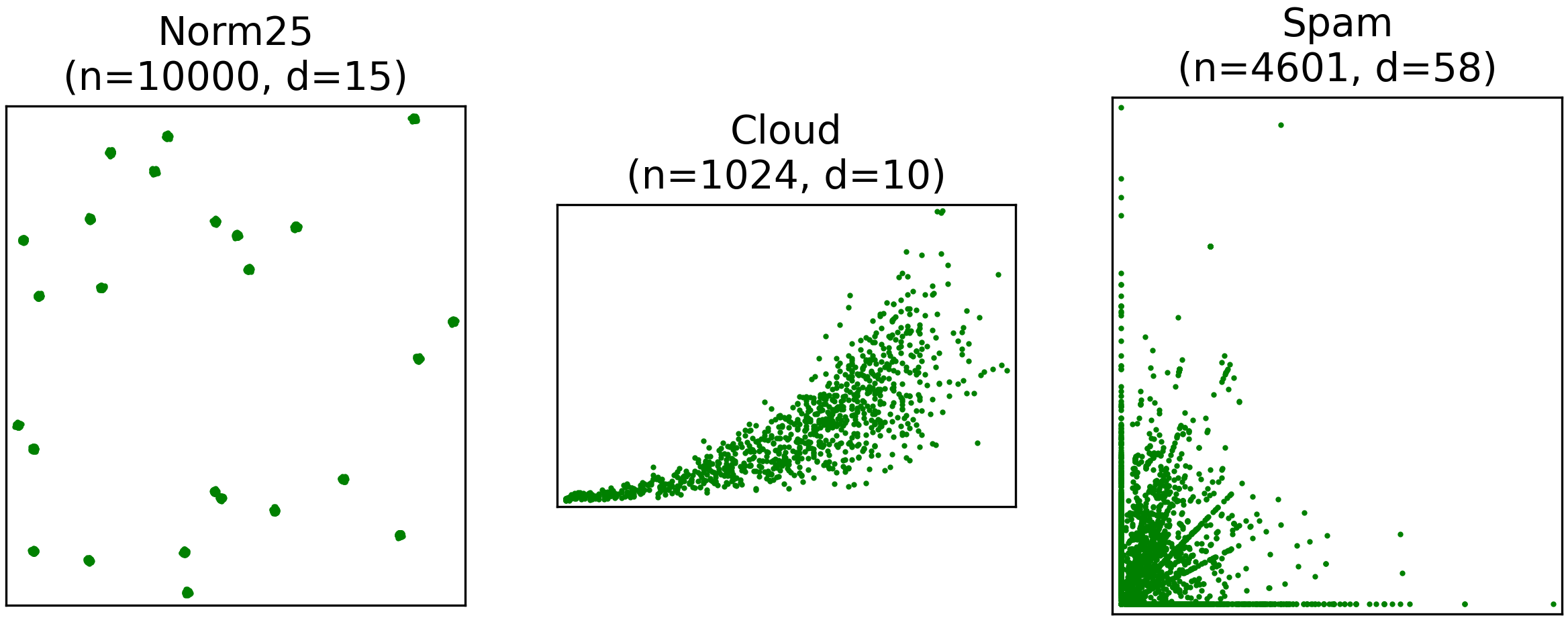}
	\caption[short caption]{\capProbHighD}
	\label{fig:prob-highD}
\end{figure}

\subsection{Solution Quality}\label{sec:results-solqual}
\setlength{\tabcolsep}{3pt}
The primary findings on solution quality are summarized below, with detailed results in Appendix \ref{sec:appendixA}. Table \ref{tab:mse-top} presents the core results, with each row representing a problem group and each column an algorithm. The values denote the mean relative MSE improvement over the baseline \gkmp{} algorithm, marked as 0.0\%. Positive values indicate improvements, while negative values denote poorer results.

\newcommand{\tabcom}{}
\renewcommand{\tabcom}{}

\begin{figure}[t]
    \centering
    \includegraphics[width=0.8\linewidth]{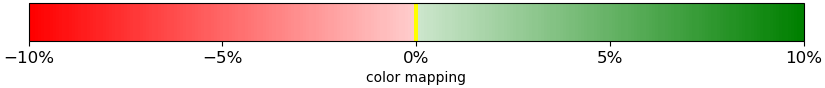}
    \caption{Color mapping used for the ``SSE improvements'' tables.}
    \label{fig:colorbar}
\end{figure}

\renewcommand{\tabcom}{Both \bkmp{} and \bkm{} consistently found improvements for all problem groups. With the exception of the ``Literature'' problems (where \rs{} performed best), \bkm{} always found the largest improvements. The values have been colorized according to the mapping shown in Figure~\ref{fig:colorbar}. The best results in each row are boxed.}
\input{coltables/t-summary-mse.tex}

\renewcommand{\tabcom}{The values for \bkm{} are tiny, indicating a quite homogeneous quality of the solutions.}
\input{coltables/t-summary-msestd.tex}

The following observations can be made regarding solution quality:

\begin{itemize} 
    
    \item Only  \bkm{} and \bkmp{} consistently beat the baseline algorithm for all problem groups. Thereby, the improvements found by \bkm{} were much larger than those found by \bkmp{}.
    \item \Rs{} found the best solutions for the ``Literature'' problems.
    \item \Vkmp{} (the original \kmp{} variant with the $\mathcal{O}(\log{}k)$ upper bound) found significantly worse solutions than the baseline algorithm.
    \item \hawo{} (the default \km{} algorithm in the \texttt{stats} package of the R programming language) always produced the worst results of all methods (far below the baseline algorithm).
\end{itemize}

In Table \ref{tab:msestd-top}, the average standard deviation corresponding to Table \ref{tab:mse-top} is shown. A noticeable pattern is that the standard deviations for \bkm{} are tiny, indicating a quite homogeneous quality of the solutions. Overall there seems to exist a negative correlation between SSE improvement and standard deviation: the higher the SSE improvement, the smaller the standard deviation, and vice versa.

\subsection{CPU Time Usage}\label{sec:results-cputime}
Here we present the primary findings on CPU time usage from our experiments which were performed on a Linux PC (AMD FX\textsuperscript{\texttrademark}-8300 eight-core Processor, 16GB RAM) running Ubuntu 22.04 LTS. Detailed, problem-specific results are provided in Appendix \ref{sec:appendixB}.

To reduce bias in the experimental results, we selected the most widely used open-source implementation of each algorithm. For \bkmp, no implementation was available, so we implemented it in Python based on \skl{} with reasonable effort to obtain an efficient implementation. Overall, this approach resulted in three different programming languages being used: Python, Fortran, and C. For algorithms with different implementation languages, the CPU time usage is not directly comparable. With this caveat, the observed CPU time usage is reported here to provide some efficiency indication.

Table \ref{tab:cpu-top} contains one row per problem group. It shows in the second column the mean CPU time in seconds used by the baseline algorithm, \gkmp, and in the following columns, the percentual CPU time usage relative to the baseline algorithm for all investigated approaches.

\renewcommand{\tabcom}{The column {``t(greedy km++)"} shows the mean CPU time used by the baseline algorithm. Green (red) background coloring indicates faster (slower) execution than the baseline algorithm}
\input{coltables/t-summary-cpu.tex} 

The algorithms implemented in Python (\gkmp, \vkmp, \bkmp, \bkm) are all based on the \skl{} library, which makes a comparison among them relatively meaningful.
The following observations can be made for CPU time usage of these
algorithms:

\begin{itemize}
    \item The fastest of these algorithms was \vkmp. It was about 23\% faster than the baseline algorithm, \gkmp{}, which, however,  had a much better solution quality.
    \item \Bkmp{} required about 9.5 times as much CPU time as the baseline algorithm. 
    \item \Bkm{} required about 5.6 times as much CPU time as the baseline algorithm).
\end{itemize}    

The only algorithm implemented in Fortran, \hawo, was the fastest overall (about 40\% faster than \gkmp). It is called from within R, but the implementation language is Fortran.

Among the two algorithms implemented in C, \ga{} was on average over 30 times faster than \rs, which was by far the most compute-heavy algorithm of all we investigated.

\subsection{Effects of Multiple Runs and Running \BKM{} Only Once}\label{sec:results-multirun}

\emph{Multiple runs} is a technique to enhance algorithmic outcomes by running it several times and choosing the best result (Section \ref{sec:multirun}). Table  \ref{tab:mse_bo10-top} shows the average improvements from ten runs of all algorithms, over the baseline.
Using the same data as Table \ref{tab:mse-top} (100 runs per problem-algorithm combo), results are grouped into ten clusters of ten, selecting the best from each. Percentage SSE difference from the baseline is calculated and averaged. Positive values in the \gkmp{} column represent the improvement from ten runs.

\renewcommand{\tabcom}{The performed experimental results for each combination of problem and algorithm have been partitioned into groups of size ten, and the best result from each group was selected. 
}
\input{coltables/t-summary-mse_bo10.tex}

Table \ref{tab:mse_diff-top} shows the SSE difference between single run and best-of-ten runs. The minimal improvements from multiple runs of \bkm{} suggest that it can be effectively compared to other algorithms' best-of-ten results, even when executed only once. Table \ref{tab:wildcard-top} confirms that \bkm{} is overall superior, averaging across all problem groups.

\renewcommand{\tabcom}{This table can be computed as the difference of Table \ref{tab:mse_bo10-top} and Table \ref{tab:mse-top}. It shows the improvement obtained by selecting the best of ten runs instead of just a single run. The tiny improvements for \bkm{} suggest that performing multiple runs is optional here to obtain good results.}   

\renewcommand{\tabcom}{This table compares the relative SSE difference between a single run and the best of ten runs, computed from Tables \ref{tab:mse_bo10-top} and \ref{tab:mse-top}. 
The minor improvements for \bkm{} indicate that multiple runs are not required to achieve desirable results.}
\input{coltables/t-summary-mse_diff.tex}

\renewcommand{\tabcom}{A single run of  \bkm{} generally outperforms the best of ten runs for the analyzed competing algorithms, barring few exceptions like \rs{} and \ga. The figures represent the mean SSE improvements over the baseline algorithm from best-of-ten runs for competitors (Table \ref{tab:mse_bo10-top}) and a single run for \bkm{} (last column of Table \ref{tab:mse-top}). \Bkm{} (1 run) was the best overall, topping three individual problem groups, and ranking second and third in the remaining two.}
\input{coltables/t-summary-wildcard.tex}

CPU time for this scenario (one run of \bkm, ten runs for others) is computed by multiplying Table  \ref{tab:cpu-top}'s first column by ten and dividing the \bkm{} percentage by ten (see Table \ref{tab:cpu-one-top}). \Bkm{} is the quickest, requiring just $55.6\%$ of the CPU time needed for ten runs of \gkmp.

\renewcommand{\tabcom}{ when all algorithms including \gkmp{} are run ten times, but \bkm{} is run only once. In this case \bkm{} is the fastest algorithm, requiring only $55.6\%$ of the CPU time needed for ten runs of \gkmp.}
\input{coltables/t-summary-cpu-one.tex}

\subsection{Effect of Varying the Breathing Depth Parameter $m$}

The breathing depth parameter $m$ (default value: 5) serves as a means of balancing solution quality and computational resource demands. Higher values of $m$ typically result in improved solutions, albeit at the expense of greater computation time, and vice versa. Figure \ref{fig:errhisto} illustrates the average error improvement relative to the baseline algorithm across all test problems for varying values of $m$. Concurrently, Figure \ref{fig:cpuhisto} presents the corresponding computation time relative to the baseline algorithm. For instance, replacing the default $m = 5$ with $m = 25$ improved the average solution quality by 0.9\% for our test problems, but quadrupled the CPU time required.

\renewcommand{\tabcom}{Average error improvement across all test problems of \bkm{} over \gkmp, for varying values of the ``breathing depth'' parameter $m$. For the default value of $m=5$, the average error improvement is 8.1\% (see the lower-right value in Table \ref{tab:mse-top}).}
\begin{figure}
	\centering
	\includegraphics[width=0.70\linewidth]{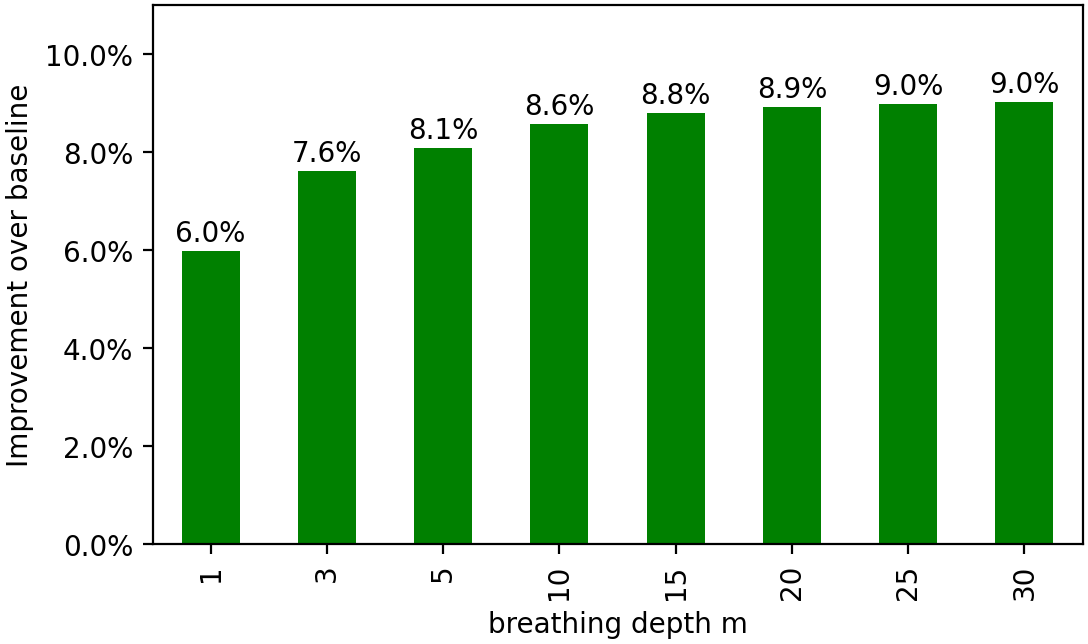}
	\caption[short caption]{\tabcom}
	\label{fig:errhisto}
\end{figure}

\renewcommand{\tabcom}{Average computation time across all test problems of \bkm{} relative to \gkmp, for varying values of the ``breathing depth'' parameter 
$m$. For the default value of $m=5$, the relative size of the CPU time is 555.6\% (see the lower-right value in Table \ref{tab:cpu-top}).}
\begin{figure}
	\centering
	\includegraphics[width=0.70\linewidth]{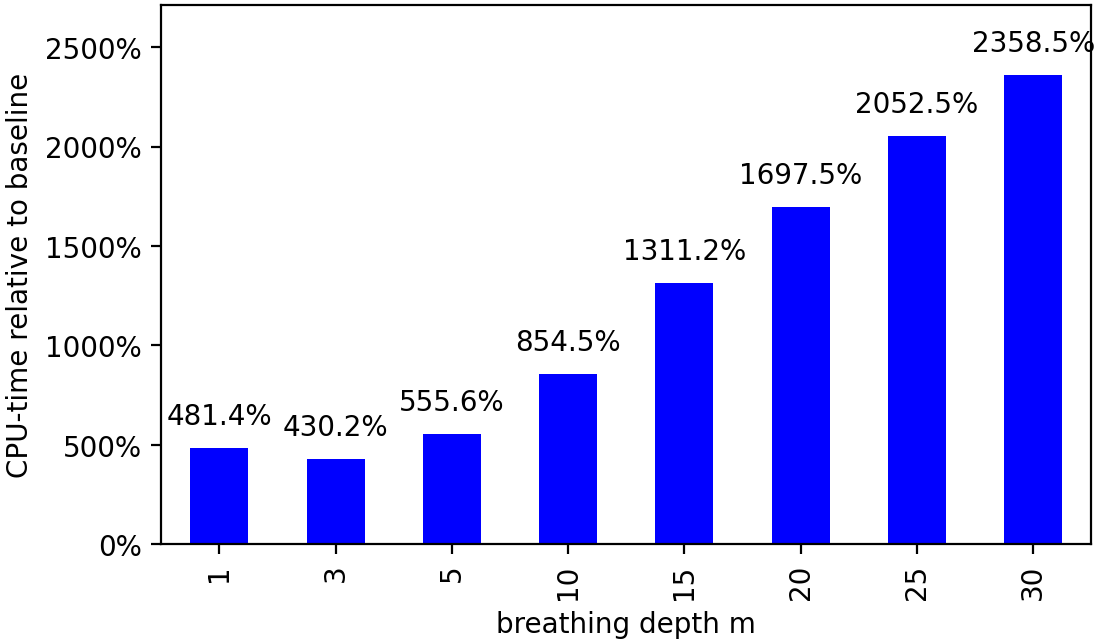}
	\caption[short caption]{\tabcom}
	\label{fig:cpuhisto}
\end{figure}

\section{Conclusion}

We introduced the novel \bkm{} algorithm which dynamically changes the size $k$ of the codebook to improve solutions found by the \GLA{}. We empirically compared \bkm{} (initialized by \gkmp) to the baseline \gkmp{} (followed by the \GLA) and five other algorithms across diverse test problems. Our approach consistently outperformed all other methods in terms of solution quality, with only a few exceptions where it slightly lagged behind \rs{} or \ga. It also was the only approach able to find near-optimal solutions across all problems in the "Known Optimum" problem group.

While \rs{} and \ga{} did outperform the baseline for most problems, they were dramatically inferior to the baseline in several cases, making them less suitable for unknown data.
The comparison between \gkmp{} and \vkmp{} underlined the improved solution quality offered by the former, validating its use as the present default \km{} algorithm in \skl.

\hawo{}, the default algorithm in R's stats package, consistently underperformed, suggesting its use should be limited to situations where low computational cost is a priority.

Notably, \bkm{} maintained its superior performance even when other algorithms were run ten times and it was only run once. In this scenario, it continued to deliver significantly better solutions than \gkmp{} while being nearly twice as fast.

Based on these findings, we recommend using \bkm{} over \gkmp{} for improved solution quality and speed.

\bibliography{bkmshort}
\FloatBarrier
\newpage
\setcounter{table}{0}
\renewcommand*\thetable{\Alph{section}.\arabic{table}}
\renewcommand{\theHsection}{A\arabic{section}}
\appendix
\pagenumbering{roman}

\section{Solution Quality Details}\label{sec:appendixA}

This section presents tables depicting solution quality per problem for all studied algorithms. The average SSE ($\Phi$) for the baseline algorithm, \gkmp, is shown with a grey background. The percentages for non-baseline algorithms represent mean relative SSE improvement over the baseline, with negative values indicating worse performance. The color scheme corresponds to Figure~\ref{fig:colorbar}. The best results in each row are boxed, including ties.

\renewcommand{\tabcom}{\Bkm{} dominates, closely followed by \rs{} and less closely by \ga{} and \bkmp. \Vkmp{} and \hawo{} are clearly inferior.}
\input{coltables/t-msee-opt.tex}

\renewcommand{\tabcom}{Green background marks cases where the optimum has been approached up to 0.001\% tolerance. Only \bkm{} was able to consistently find such near-optimal solutions. Shades of blue indicate varying deviations from the optimum.}
\input{coltables/t-mse-realopt.tex}

\renewcommand{\tabcom}{For all problems in this group, \rs{} found the best solutions.}
\input{coltables/t-msee-fraenti.tex}

\renewcommand{\tabcom}{For all problems, \bkm{} found the best solutions. Also, \bkmp{} was able to improve upon the baseline algorithm in all cases but by a considerably smaller margin. The other approaches produced worse solutions than the baseline algorithm for some or all problems.}
\input{coltables/t-msee-fr-mod.tex}

\renewcommand{\tabcom}{\Bkm{} is best apart from the first two problems. 
}
\input{coltables/t-msee-espiralX.tex}

\renewcommand{\tabcom}{The problem with the \emph{Norm25} data set and $k = 25$ seems to be so simple that all algorithms, except \hawo, found the same solution (one centroid per cluster).
The dimensionality of the data sets \emph{Norm25}, \emph{Cloud}, and \emph{Spam} is 15, 10, and 58, respectively.
}
\input{coltables/t-msee-high-D.tex}
\FloatBarrier
\setcounter{table}{0}
\section{CPU Time Details}\label{sec:appendixB}

In this section, tables for each problem group illustrate the CPU time usage per problem for all investigated algorithms. The average CPU time in seconds for the baseline algorithm, \gkmp, is displayed with a blue background. Percentage values for non-baseline algorithms represent the \emph{CPU time relative to the baseline algorithm} for each specific \km{} problem. Values below 100\% (faster than the baseline) are on green backgrounds, while values above 100\% (slower than the baseline) are on red backgrounds.

\input{coltables/t-cpu-opt.tex}

\input{coltables/t-cpu-fraenti.tex}

\input{coltables/t-cpu-fr-mod.tex}

\input{coltables/t-cpu-espiralX.tex}

\renewcommand{\tabcom}{The dimensionality of the data sets \emph{Norm25}, \emph{Cloud}, and \emph{Spam} is 15, 10, and 58, respectively.}
\input{coltables/t-cpu-high-D.tex}

\end{document}

%% file: img/prob-opt.tex
\begin{figure}
	\centering
	\includegraphics[width=\fiwid\linewidth]{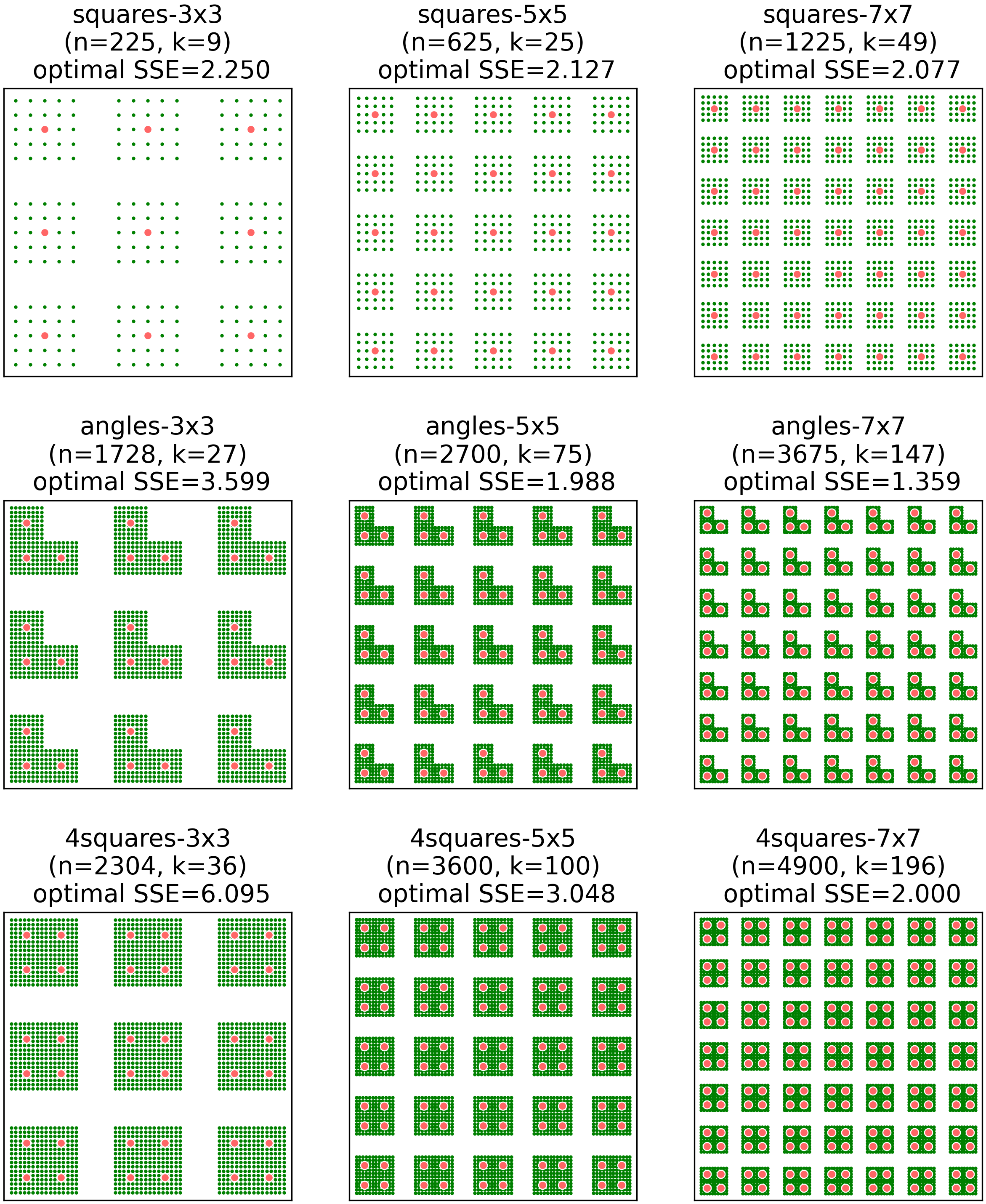}
	\caption[short caption]{\capProbOpt}
	\label{fig:prob-opt}
\end{figure}

%% file: img/prob-lit.tex
\begin{figure}
	\centering
	\includegraphics[width=\fiwid\linewidth]{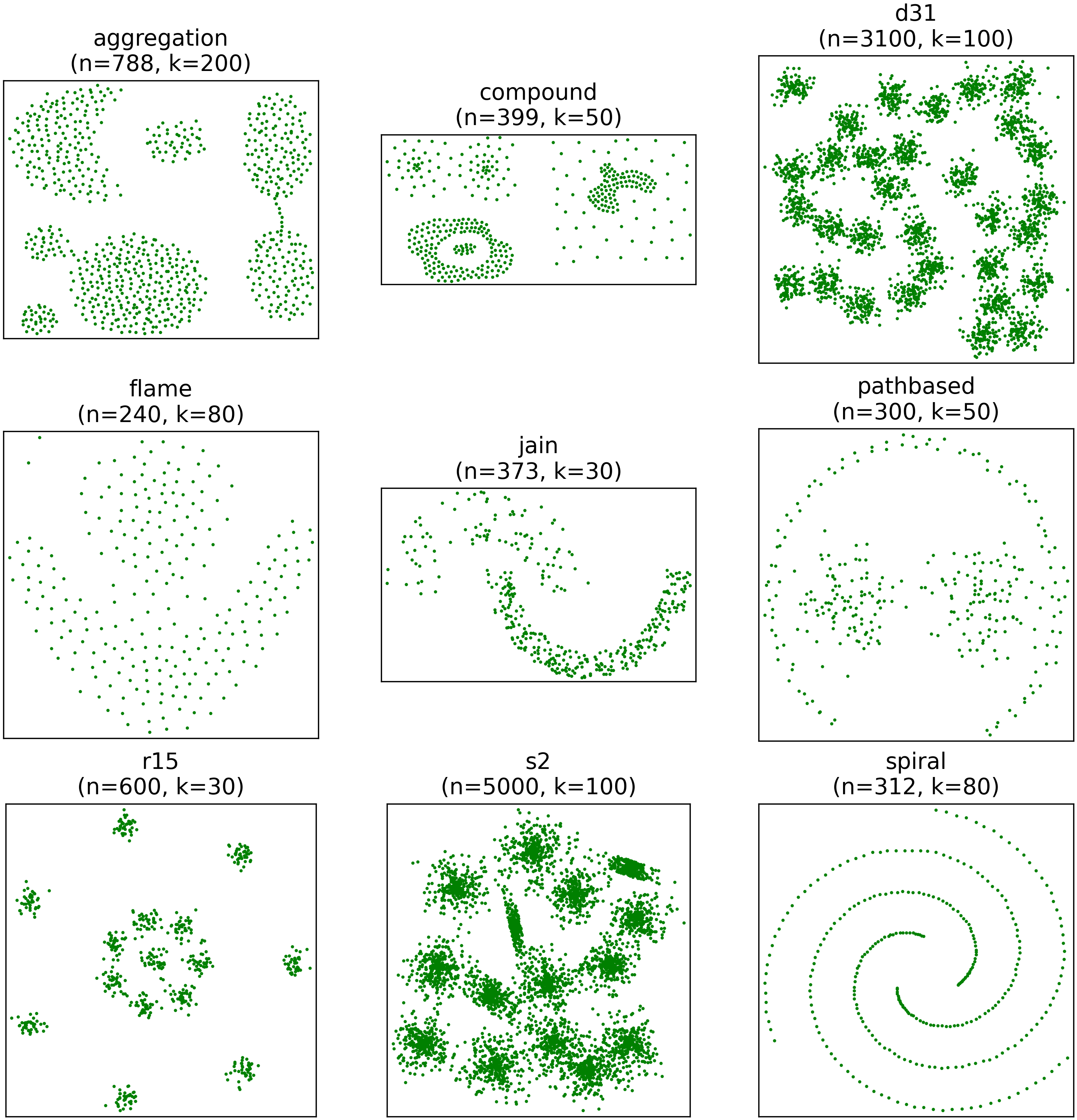}
	\caption[short caption]{\capProbLit}
	\label{fig:prob-lit}
\end{figure}

%% file: img/prob-litmod.tex
\begin{figure}
	\centering
	\includegraphics[width=\fiwid\linewidth]{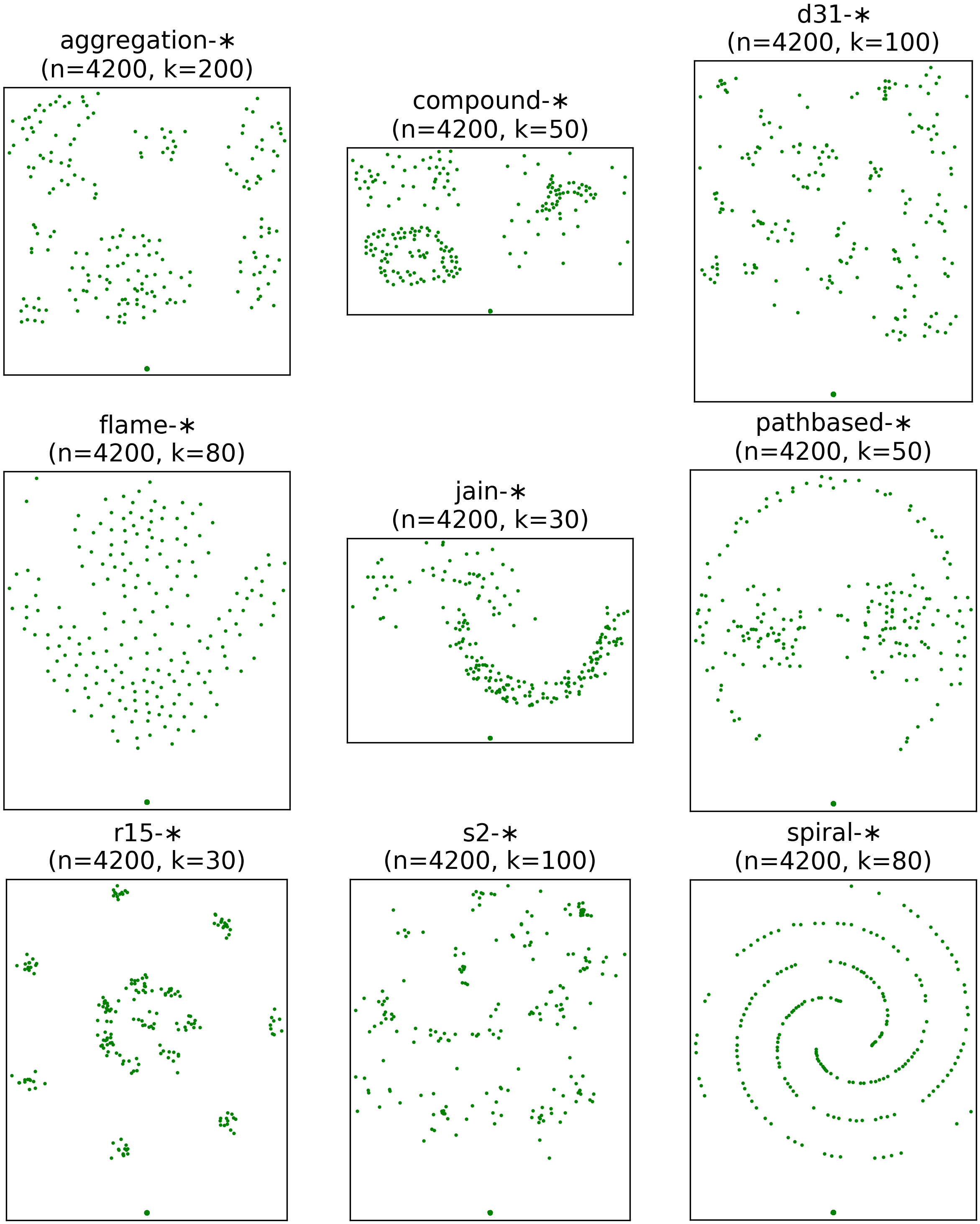}
	\caption[short caption]{\capProbLitmod}
	\label{fig:prob-litmod}
\end{figure}

%% file: img/prob-espiral.tex
\begin{figure}
	\centering
	\includegraphics[width=\fiwid\linewidth]{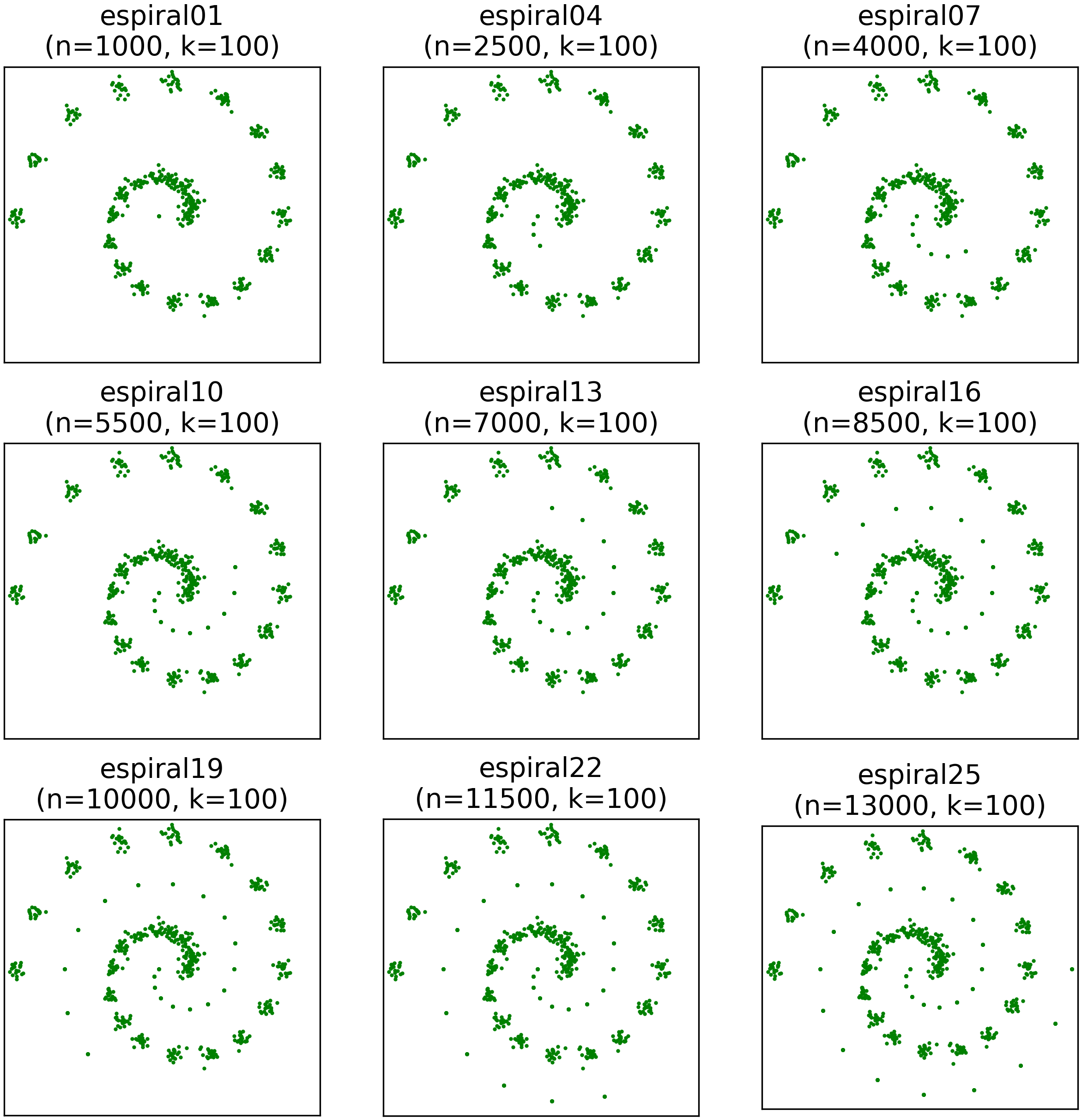}
	\caption[short caption]{\capProbEspiral}
	\label{fig:prob-espiral}
\end{figure}

%% file: coltables/t-summary-mse.tex
\begin{table}[t]
    \small
    \begin{center}
    \input{coltables/mse-top.tex}
    \caption{SSE improvements relative to \gkmp. \tabcom}
    \label{tab:mse-top}
    \end{center}
\end{table}
\renewcommand{\tabcom}{}

%% file: coltables/mse-top.tex
\begin{tabular}{lrrrrrrr}
\toprule
problem group & \thead{greedy \\ km++} & \thead{vanilla \\ km++} & \thead{better \\ km++} & \thead{Hartigan- \\ Wong} & \thead{genetic \\ algorithm} & \thead{random \\ swap} & \thead{breathing \\ k-means} \\
\midrule
Known Optimum & {\cellcolor[HTML]{FFFF00}} \color[HTML]{000000} 0.0\% & {\cellcolor[HTML]{FF0000}} \color[HTML]{F1F1F1} -15.5\% & {\cellcolor[HTML]{86C286}} \color[HTML]{000000} 5.2\% & {\cellcolor[HTML]{FF0000}} \color[HTML]{F1F1F1} -30.4\% & {\cellcolor[HTML]{50A850}} \color[HTML]{F1F1F1} 9.1\% & {\cellcolor[HTML]{4AA44A}} \color[HTML]{F1F1F1} 9.6\% &\framebox[35pt]{ ~\hfill {\cellcolor[HTML]{48A348}} \color[HTML]{F1F1F1} 9.7\% }\\
Literature & {\cellcolor[HTML]{FFFF00}} \color[HTML]{000000} 0.0\% & {\cellcolor[HTML]{FF7272}} \color[HTML]{F1F1F1} -6.5\% & {\cellcolor[HTML]{B4DAB4}} \color[HTML]{000000} 1.8\% & {\cellcolor[HTML]{FF4E4E}} \color[HTML]{F1F1F1} -9.2\% & {\cellcolor[HTML]{60B060}} \color[HTML]{F1F1F1} 7.9\% &\framebox[35pt]{ ~\hfill {\cellcolor[HTML]{4AA44A}} \color[HTML]{F1F1F1} 9.6\% }& {\cellcolor[HTML]{66B266}} \color[HTML]{F1F1F1} 7.5\% \\
Modified Literature & {\cellcolor[HTML]{FFFF00}} \color[HTML]{000000} 0.0\% & {\cellcolor[HTML]{FF0000}} \color[HTML]{F1F1F1} -19.6\% & {\cellcolor[HTML]{A8D3A8}} \color[HTML]{000000} 2.8\% & {\cellcolor[HTML]{FF0000}} \color[HTML]{F1F1F1} -16785.0\% & {\cellcolor[HTML]{FF0000}} \color[HTML]{F1F1F1} -141.8\% & {\cellcolor[HTML]{FF0000}} \color[HTML]{F1F1F1} -221.6\% &\framebox[35pt]{ ~\hfill {\cellcolor[HTML]{44A244}} \color[HTML]{F1F1F1} 10.0\% }\\
Evil Spiral & {\cellcolor[HTML]{FFFF00}} \color[HTML]{000000} 0.0\% & {\cellcolor[HTML]{FF2A2A}} \color[HTML]{F1F1F1} -11.8\% & {\cellcolor[HTML]{A8D3A8}} \color[HTML]{000000} 2.8\% & {\cellcolor[HTML]{FF0000}} \color[HTML]{F1F1F1} -1910.9\% & {\cellcolor[HTML]{BCDEBC}} \color[HTML]{000000} 1.2\% & {\cellcolor[HTML]{74BA74}} \color[HTML]{F1F1F1} 6.4\% &\framebox[35pt]{ ~\hfill {\cellcolor[HTML]{50A850}} \color[HTML]{F1F1F1} 9.1\% }\\
High-Dimensional & {\cellcolor[HTML]{FFFF00}} \color[HTML]{000000} 0.0\% & {\cellcolor[HTML]{FF7474}} \color[HTML]{F1F1F1} -6.5\% & {\cellcolor[HTML]{B8DCB8}} \color[HTML]{000000} 1.5\% & {\cellcolor[HTML]{FF0000}} \color[HTML]{F1F1F1} -23642.4\% & {\cellcolor[HTML]{FF0000}} \color[HTML]{F1F1F1} -335.2\% & {\cellcolor[HTML]{FF0000}} \color[HTML]{F1F1F1} -72.2\% &\framebox[35pt]{ ~\hfill {\cellcolor[HTML]{92C892}} \color[HTML]{000000} 4.3\% }\\
\midrule
Mean & {\cellcolor[HTML]{FFFF00}} \color[HTML]{000000} 0.0\% & {\cellcolor[HTML]{FF2828}} \color[HTML]{F1F1F1} -12.0\% & {\cellcolor[HTML]{A6D2A6}} \color[HTML]{000000} 2.8\% & {\cellcolor[HTML]{FF0000}} \color[HTML]{F1F1F1} -8475.6\% & {\cellcolor[HTML]{FF0000}} \color[HTML]{F1F1F1} -91.8\% & {\cellcolor[HTML]{FF0000}} \color[HTML]{F1F1F1} -53.6\% &\framebox[35pt]{ ~\hfill {\cellcolor[HTML]{5EAE5E}} \color[HTML]{F1F1F1} 8.1\% }\\
\toprule
\end{tabular}

%% file: coltables/t-summary-msestd.tex
\begin{table}[t]
    \small
    \begin{center}
    \input{coltables/msestd-top.tex}
    \caption{Standard deviations corresponding to the SSE improvements from Table \ref{tab:mse-top}. \tabcom}
    \label{tab:msestd-top}
    \end{center}
\end{table}
\renewcommand{\tabcom}{}

%% file: coltables/msestd-top.tex
\begin{tabular}{lrrrrrrr}
\toprule
problem group & \thead{greedy \\ km++} & \thead{vanilla \\ km++} & \thead{better \\ km++} & \thead{Hartigan- \\ Wong} & \thead{genetic \\ algorithm} & \thead{random \\ swap} & \thead{breathing \\ k-means} \\
\midrule
Known Optimum & 5.7\% & 9.9\% & 2.5\% & 13.6\% & 0.2\% & 0.1\% & 0.0\% \\
Literature & 2.1\% & 3.5\% & 1.8\% & 7.9\% & 0.6\% & 0.4\% & 0.9\% \\
Modified Literature & 3.5\% & 8.9\% & 2.8\% & 5606.9\% & 119.9\% & 38.3\% & 1.3\% \\
Evil Spiral & 2.3\% & 4.6\% & 1.8\% & 828.5\% & 14.5\% & 1.2\% & 0.8\% \\
High-Dimensional & 2.1\% & 5.4\% & 1.3\% & 9395.0\% & 532.2\% & 248.2\% & 0.5\% \\
\midrule
Mean & 3.1\% & 6.5\% & 2.0\% & 3170.4\% & 133.5\% & 57.6\% & 0.7\% \\
\toprule
\end{tabular}

%% file: coltables/t-summary-cpu.tex
\begin{table}[t]
    \small
    \begin{center}
    \input{coltables/cpu-top.tex}
    \caption{CPU time relative to \gkmp. \tabcom}
    \label{tab:cpu-top}
    \end{center}
\end{table}
\renewcommand{\tabcom}{}

%% file: coltables/cpu-top.tex
\begin{tabular}{lrrrrrrr}
\toprule
problem group & t(\thead{t(greedy \\ km++)\\Python}) & \thead{vanilla \\ km++\\Python} & \thead{better \\ km++\\ \textsf{Python}} & \thead{Hartigan- \\ Wong\\R/Fortran} & \thead{genetic \\ algorithm\\C} & \thead{random \\ swap\\C} & \thead{breathing \\ k-means\\Python} \\
\midrule
Known Optimum & {\cellcolor{blue}} \color[HTML]{FFFFFF} 0.07s & {\cellcolor[HTML]{D6EBD6}} \color[HTML]{000000} 83.8\% & {\cellcolor[HTML]{FFDDDD}} \color[HTML]{000000} 754.8\% & {\cellcolor[HTML]{2E972E}} \color[HTML]{FFFFFF} 18.2\% & {\cellcolor[HTML]{FFF6F6}} \color[HTML]{000000} 289.4\% & {\cellcolor[HTML]{FF2323}} \color[HTML]{FFFFFF} 4327.4\% & {\cellcolor[HTML]{FFF1F1}} \color[HTML]{000000} 382.4\% \\
Literature & {\cellcolor{blue}} \color[HTML]{FFFFFF} 0.06s & {\cellcolor[HTML]{D5EAD5}} \color[HTML]{000000} 83.3\% & {\cellcolor[HTML]{FFE7E7}} \color[HTML]{000000} 568.2\% & {\cellcolor[HTML]{219021}} \color[HTML]{FFFFFF} 13.1\% & {\cellcolor[HTML]{FFF8F8}} \color[HTML]{000000} 237.1\% & {\cellcolor[HTML]{FF8585}} \color[HTML]{000000} 2439.9\% & {\cellcolor[HTML]{FFEDED}} \color[HTML]{000000} 454.7\% \\
Modified Literature & {\cellcolor{blue}} \color[HTML]{FFFFFF} 0.07s & {\cellcolor[HTML]{ABD5AB}} \color[HTML]{000000} 67.0\% & {\cellcolor[HTML]{FFCBCB}} \color[HTML]{000000} 1104.1\% & {\cellcolor[HTML]{71B871}} \color[HTML]{000000} 44.5\% & {\cellcolor[HTML]{FFE7E7}} \color[HTML]{000000} 575.9\% & {\cellcolor[HTML]{FF0000}} \color[HTML]{FFFFFF} 36358.1\% & {\cellcolor[HTML]{FFE9E9}} \color[HTML]{000000} 523.2\% \\
Evil Spiral & {\cellcolor{blue}} \color[HTML]{FFFFFF} 0.10s & {\cellcolor[HTML]{A9D4A9}} \color[HTML]{000000} 66.1\% & {\cellcolor[HTML]{FFBCBC}} \color[HTML]{000000} 1398.4\% & {\cellcolor[HTML]{43A143}} \color[HTML]{FFFFFF} 26.3\% & {\cellcolor[HTML]{FFEDED}} \color[HTML]{000000} 454.5\% & {\cellcolor[HTML]{FF3B3B}} \color[HTML]{FFFFFF} 3857.2\% & {\cellcolor[HTML]{FFE5E5}} \color[HTML]{000000} 605.3\% \\
High-Dimensional & {\cellcolor{blue}} \color[HTML]{FFFFFF} 0.10s & {\cellcolor[HTML]{DAEDDA}} \color[HTML]{000000} 85.4\% & {\cellcolor[HTML]{FFD6D6}} \color[HTML]{000000} 903.1\% & {\cellcolor[HTML]{FFFBFB}} \color[HTML]{000000} 195.6\% & {\cellcolor[HTML]{FFB3B3}} \color[HTML]{000000} 1571.5\% & {\cellcolor[HTML]{FF0000}} \color[HTML]{FFFFFF} 69674.6\% & {\cellcolor[HTML]{FFDADA}} \color[HTML]{000000} 812.7\% \\
\midrule
Mean & {\cellcolor{blue}} \color[HTML]{FFFFFF} 0.08s & {\cellcolor[HTML]{C5E2C5}} \color[HTML]{000000} 77.1\% & {\cellcolor[HTML]{FFD3D3}} \color[HTML]{000000} 945.7\% & {\cellcolor[HTML]{98CC98}} \color[HTML]{000000} 59.5\% & {\cellcolor[HTML]{FFE4E4}} \color[HTML]{000000} 625.7\% & {\cellcolor[HTML]{FF0000}} \color[HTML]{FFFFFF} 23331.4\% & {\cellcolor[HTML]{FFE8E8}} \color[HTML]{000000} 555.6\% \\
\toprule
\end{tabular}

%% file: coltables/t-summary-mse_bo10.tex
\begin{table}[t]
    \small
    \begin{center}
    \input{coltables/mse_bo10-top.tex}
    \caption{Best of 10 runs: SSE improvements relative to \gkmp. \tabcom}
    \label{tab:mse_bo10-top}
    \end{center}
\end{table}
\renewcommand{\tabcom}{}

%% file: coltables/mse_bo10-top.tex
\begin{tabular}{lrrrrrrr}
\toprule
problem group & \thead{greedy \\ km++ \\ (10 runs)} & \thead{vanilla \\ km++\\ (10 runs)} & \thead{better \\ km++\\ (10 runs)} & \thead{Hartigan- \\ Wong\\ (10 runs)} & \thead{genetic \\ algorithm\\ (10 runs)} & \thead{random \\ swap\\ (10 runs)} & \thead{breathing \\ k-means\\ (10 runs)} \\
\midrule
Known Optimum & {\cellcolor[HTML]{80C080}} \color[HTML]{000000} 5.6\% & {\cellcolor[HTML]{FFA2A2}} \color[HTML]{000000} -3.1\% & {\cellcolor[HTML]{62B062}} \color[HTML]{F1F1F1} 7.8\% & {\cellcolor[HTML]{FF2020}} \color[HTML]{F1F1F1} -12.5\% & {\cellcolor[HTML]{4CA64C}} \color[HTML]{F1F1F1} 9.3\% & {\cellcolor[HTML]{48A348}} \color[HTML]{F1F1F1} 9.6\% &\framebox[35pt]{ ~\hfill {\cellcolor[HTML]{48A348}} \color[HTML]{F1F1F1} 9.7\% }\\
Literature & {\cellcolor[HTML]{A4D2A4}} \color[HTML]{000000} 3.0\% & {\cellcolor[HTML]{FFB8B8}} \color[HTML]{000000} -1.5\% & {\cellcolor[HTML]{90C890}} \color[HTML]{000000} 4.4\% & {\cellcolor[HTML]{C6E2C6}} \color[HTML]{000000} 0.6\% & {\cellcolor[HTML]{54AA54}} \color[HTML]{F1F1F1} 8.7\% &\framebox[35pt]{ ~\hfill {\cellcolor[HTML]{42A042}} \color[HTML]{F1F1F1} 10.1\% }& {\cellcolor[HTML]{54AA54}} \color[HTML]{F1F1F1} 8.8\% \\
Modified Literature & {\cellcolor[HTML]{88C388}} \color[HTML]{000000} 5.0\% & {\cellcolor[HTML]{FF6868}} \color[HTML]{F1F1F1} -7.2\% & {\cellcolor[HTML]{70B870}} \color[HTML]{F1F1F1} 6.9\% & {\cellcolor[HTML]{FF0000}} \color[HTML]{F1F1F1} -9395.2\% & {\cellcolor[HTML]{FF0000}} \color[HTML]{F1F1F1} -37.9\% & {\cellcolor[HTML]{FF0000}} \color[HTML]{F1F1F1} -167.1\% &\framebox[35pt]{ ~\hfill {\cellcolor[HTML]{2E962E}} \color[HTML]{F1F1F1} 11.6\% }\\
Evil Spiral & {\cellcolor[HTML]{A0D0A0}} \color[HTML]{000000} 3.4\% & {\cellcolor[HTML]{FF8888}} \color[HTML]{F1F1F1} -5.0\% & {\cellcolor[HTML]{84C284}} \color[HTML]{000000} 5.4\% & {\cellcolor[HTML]{FF0000}} \color[HTML]{F1F1F1} -967.2\% & {\cellcolor[HTML]{56AB56}} \color[HTML]{F1F1F1} 8.6\% & {\cellcolor[HTML]{5EAE5E}} \color[HTML]{F1F1F1} 8.1\% &\framebox[35pt]{ ~\hfill {\cellcolor[HTML]{42A042}} \color[HTML]{F1F1F1} 10.1\% }\\
High-Dimensional & {\cellcolor[HTML]{AAD4AA}} \color[HTML]{000000} 2.5\% & {\cellcolor[HTML]{FFC6C6}} \color[HTML]{000000} -0.7\% & {\cellcolor[HTML]{A4D2A4}} \color[HTML]{000000} 3.0\% & {\cellcolor[HTML]{FF0000}} \color[HTML]{F1F1F1} -11848.1\% & {\cellcolor[HTML]{C6E2C6}} \color[HTML]{000000} 0.5\% & {\cellcolor[HTML]{8CC68C}} \color[HTML]{000000} 4.7\% &\framebox[35pt]{ ~\hfill {\cellcolor[HTML]{8AC48A}} \color[HTML]{000000} 4.9\% }\\
\midrule
Mean & {\cellcolor[HTML]{96CB96}} \color[HTML]{000000} 3.9\% & {\cellcolor[HTML]{FF9E9E}} \color[HTML]{000000} -3.5\% & {\cellcolor[HTML]{82C082}} \color[HTML]{000000} 5.5\% & {\cellcolor[HTML]{FF0000}} \color[HTML]{F1F1F1} -4444.5\% & {\cellcolor[HTML]{FFB0B0}} \color[HTML]{000000} -2.1\% & {\cellcolor[HTML]{FF0000}} \color[HTML]{F1F1F1} -26.9\% &\framebox[35pt]{ ~\hfill {\cellcolor[HTML]{52A852}} \color[HTML]{F1F1F1} 9.0\% }\\
\toprule
\end{tabular}

%% file: coltables/t-summary-mse_diff.tex
\begin{table}[t]
    \small
    \begin{center}
    \input{coltables/mse_diff-top.tex}
    \caption{Relative SSE difference between single run and best of 10 runs. \tabcom}
    \label{tab:mse_diff-top}
    \end{center}
\end{table}
\renewcommand{\tabcom}{}

%% file: coltables/mse_diff-top.tex
\begin{tabular}{lrrrrrrr}
\toprule
problem group & \thead{greedy \\ km++} & \thead{vanilla \\ km++} & \thead{better \\ km++} & \thead{Hartigan- \\ Wong} & \thead{genetic \\ algorithm} & \thead{random \\ swap} & \thead{breathing \\ k-means} \\
\midrule
Known Optimum & 5.6\% & 12.4\% & 2.6\% & 17.9\% & 0.2\% & 0.1\% & 0.0\% \\
Literature & 3.0\% & 5.0\% & 2.6\% & 9.9\% & 0.9\% & 0.5\% & 1.2\% \\
Modified Literature & 5.0\% & 12.4\% & 4.1\% & 7389.8\% & 103.9\% & 54.5\% & 1.6\% \\
Evil Spiral & 3.4\% & 6.8\% & 2.6\% & 943.7\% & 7.5\% & 1.7\% & 1.0\% \\
High-Dimensional & 2.5\% & 5.8\% & 1.5\% & 11794.3\% & 335.7\% & 76.9\% & 0.7\% \\
\midrule
Mean & 3.9\% & 8.5\% & 2.7\% & 4031.1\% & 89.6\% & 26.7\% & 0.9\% \\
\toprule
\end{tabular}

%% file: coltables/t-summary-wildcard.tex
\begin{table}[t]
    \small
    \begin{center}
    \input{coltables/wildcard-top.tex}
    \caption{  \tabcom}
    \label{tab:wildcard-top}
    \end{center}
\end{table}
\renewcommand{\tabcom}{}

%% file: coltables/wildcard-top.tex
\begin{tabular}{lrrrrrrr}
\toprule
problem group & \thead{greedy \\ km++ \\ (10 runs)} & \thead{vanilla \\ km++ \\ (10 runs)} & \thead{better \\ km++ \\ (10 runs)} & \thead{Hartigan- \\ Wong \\ (10 runs)} & \thead{genetic \\ algorithm \\ (10 runs)} & \thead{random \\ swap \\ (10 runs)} & \thead{\color[HTML]{FF0000} \textbf{breathing} \\ \color[HTML]{FF0000} \textbf{k-means} \\ \color[HTML]{FF0000} \textbf{(1 run)}} \\
\midrule
Known Optimum & {\cellcolor[HTML]{80C080}} \color[HTML]{000000} 5.6\% & {\cellcolor[HTML]{FFA2A2}} \color[HTML]{000000} -3.1\% & {\cellcolor[HTML]{62B062}} \color[HTML]{F1F1F1} 7.8\% & {\cellcolor[HTML]{FF2020}} \color[HTML]{F1F1F1} -12.5\% & {\cellcolor[HTML]{4CA64C}} \color[HTML]{F1F1F1} 9.3\% & {\cellcolor[HTML]{48A348}} \color[HTML]{F1F1F1} 9.6\% &\framebox[35pt]{ ~\hfill {\cellcolor[HTML]{48A348}} \color[HTML]{F1F1F1} 9.7\% }\\
Literature & {\cellcolor[HTML]{A4D2A4}} \color[HTML]{000000} 3.0\% & {\cellcolor[HTML]{FFB8B8}} \color[HTML]{000000} -1.5\% & {\cellcolor[HTML]{90C890}} \color[HTML]{000000} 4.4\% & {\cellcolor[HTML]{C6E2C6}} \color[HTML]{000000} 0.6\% & {\cellcolor[HTML]{54AA54}} \color[HTML]{F1F1F1} 8.7\% &\framebox[35pt]{ ~\hfill {\cellcolor[HTML]{42A042}} \color[HTML]{F1F1F1} 10.1\% }& {\cellcolor[HTML]{66B266}} \color[HTML]{F1F1F1} 7.5\% \\
Modified Literature & {\cellcolor[HTML]{88C388}} \color[HTML]{000000} 5.0\% & {\cellcolor[HTML]{FF6868}} \color[HTML]{F1F1F1} -7.2\% & {\cellcolor[HTML]{70B870}} \color[HTML]{F1F1F1} 6.9\% & {\cellcolor[HTML]{FF0000}} \color[HTML]{F1F1F1} -9395.2\% & {\cellcolor[HTML]{FF0000}} \color[HTML]{F1F1F1} -37.9\% & {\cellcolor[HTML]{FF0000}} \color[HTML]{F1F1F1} -167.1\% &\framebox[35pt]{ ~\hfill {\cellcolor[HTML]{44A244}} \color[HTML]{F1F1F1} 10.0\% }\\
Evil Spiral & {\cellcolor[HTML]{A0D0A0}} \color[HTML]{000000} 3.4\% & {\cellcolor[HTML]{FF8888}} \color[HTML]{F1F1F1} -5.0\% & {\cellcolor[HTML]{84C284}} \color[HTML]{000000} 5.4\% & {\cellcolor[HTML]{FF0000}} \color[HTML]{F1F1F1} -967.2\% & {\cellcolor[HTML]{56AB56}} \color[HTML]{F1F1F1} 8.6\% & {\cellcolor[HTML]{5EAE5E}} \color[HTML]{F1F1F1} 8.1\% &\framebox[35pt]{ ~\hfill {\cellcolor[HTML]{50A850}} \color[HTML]{F1F1F1} 9.1\% }\\
High-Dimensional & {\cellcolor[HTML]{AAD4AA}} \color[HTML]{000000} 2.5\% & {\cellcolor[HTML]{FFC6C6}} \color[HTML]{000000} -0.7\% & {\cellcolor[HTML]{A4D2A4}} \color[HTML]{000000} 3.0\% & {\cellcolor[HTML]{FF0000}} \color[HTML]{F1F1F1} -11848.1\% & {\cellcolor[HTML]{C6E2C6}} \color[HTML]{000000} 0.5\% &\framebox[35pt]{ ~\hfill {\cellcolor[HTML]{8CC68C}} \color[HTML]{000000} 4.7\% }& {\cellcolor[HTML]{92C892}} \color[HTML]{000000} 4.3\% \\
\midrule
Mean & {\cellcolor[HTML]{96CB96}} \color[HTML]{000000} 3.9\% & {\cellcolor[HTML]{FF9E9E}} \color[HTML]{000000} -3.5\% & {\cellcolor[HTML]{82C082}} \color[HTML]{000000} 5.5\% & {\cellcolor[HTML]{FF0000}} \color[HTML]{F1F1F1} -4444.5\% & {\cellcolor[HTML]{FFB0B0}} \color[HTML]{000000} -2.1\% & {\cellcolor[HTML]{FF0000}} \color[HTML]{F1F1F1} -26.9\% &\framebox[35pt]{ ~\hfill {\cellcolor[HTML]{5EAE5E}} \color[HTML]{F1F1F1} 8.1\% }\\
\toprule
\end{tabular}

%% file: coltables/t-summary-cpu-one.tex
\begin{table}[t]
    \small
    \begin{center}
    \input{coltables/cpu-one-top.tex}
    \caption{CPU time relative to \gkmp\tabcom}
    \label{tab:cpu-one-top}
    \end{center}
\end{table}
\renewcommand{\tabcom}{}

%% file: coltables/cpu-one-top.tex
\begin{tabular}{lrrrrrrr}
\toprule
problem group & \thead{t(greedy \\ km++) \\ (10 runs)} & \thead{vanilla \\ km++ \\ (10 runs)} & \thead{better \\ km++ \\ (10 runs)} & \thead{Hartigan- \\ Wong \\ (10 runs)} & \thead{genetic \\ algorithm \\ (10 runs)} & \thead{random \\ swap \\ (10 runs)} & \thead{\color[HTML]{FF0000} \textbf{breathing} \\ \color[HTML]{FF0000} \textbf{k-means} \\ \color[HTML]{FF0000} \textbf{(1 run)}} \\
\midrule
Known Optimum & {\cellcolor{blue}} \color[HTML]{FFFFFF} 0.69s & {\cellcolor[HTML]{D6EBD6}} \color[HTML]{000000} 83.8\% & {\cellcolor[HTML]{FFDDDD}} \color[HTML]{000000} 754.8\% & {\cellcolor[HTML]{2E972E}} \color[HTML]{FFFFFF} 18.2\% & {\cellcolor[HTML]{FFF6F6}} \color[HTML]{000000} 289.4\% & {\cellcolor[HTML]{FF2323}} \color[HTML]{FFFFFF} 4327.4\% & {\cellcolor[HTML]{61B061}} \color[HTML]{000000} 38.2\% \\
Literature & {\cellcolor{blue}} \color[HTML]{FFFFFF} 0.57s & {\cellcolor[HTML]{D5EAD5}} \color[HTML]{000000} 83.3\% & {\cellcolor[HTML]{FFE7E7}} \color[HTML]{000000} 568.2\% & {\cellcolor[HTML]{219021}} \color[HTML]{FFFFFF} 13.1\% & {\cellcolor[HTML]{FFF8F8}} \color[HTML]{000000} 237.1\% & {\cellcolor[HTML]{FF8585}} \color[HTML]{000000} 2439.9\% & {\cellcolor[HTML]{74BA74}} \color[HTML]{000000} 45.5\% \\
Modified Literature & {\cellcolor{blue}} \color[HTML]{FFFFFF} 0.68s & {\cellcolor[HTML]{ABD5AB}} \color[HTML]{000000} 67.0\% & {\cellcolor[HTML]{FFCBCB}} \color[HTML]{000000} 1104.1\% & {\cellcolor[HTML]{71B871}} \color[HTML]{000000} 44.5\% & {\cellcolor[HTML]{FFE7E7}} \color[HTML]{000000} 575.9\% & {\cellcolor[HTML]{FF0000}} \color[HTML]{FFFFFF} 36358.1\% & {\cellcolor[HTML]{85C285}} \color[HTML]{000000} 52.3\% \\
Evil Spiral & {\cellcolor{blue}} \color[HTML]{FFFFFF} 0.98s & {\cellcolor[HTML]{A9D4A9}} \color[HTML]{000000} 66.1\% & {\cellcolor[HTML]{FFBCBC}} \color[HTML]{000000} 1398.4\% & {\cellcolor[HTML]{43A143}} \color[HTML]{FFFFFF} 26.3\% & {\cellcolor[HTML]{FFEDED}} \color[HTML]{000000} 454.5\% & {\cellcolor[HTML]{FF3B3B}} \color[HTML]{FFFFFF} 3857.2\% & {\cellcolor[HTML]{9ACD9A}} \color[HTML]{000000} 60.5\% \\
High-Dimensional & {\cellcolor{blue}} \color[HTML]{FFFFFF} 0.95s & {\cellcolor[HTML]{DAEDDA}} \color[HTML]{000000} 85.4\% & {\cellcolor[HTML]{FFD6D6}} \color[HTML]{000000} 903.1\% & {\cellcolor[HTML]{FFFBFB}} \color[HTML]{000000} 195.6\% & {\cellcolor[HTML]{FFB3B3}} \color[HTML]{000000} 1571.5\% & {\cellcolor[HTML]{FF0000}} \color[HTML]{FFFFFF} 69674.6\% & {\cellcolor[HTML]{D0E8D0}} \color[HTML]{000000} 81.3\% \\
\midrule
Mean & {\cellcolor{blue}} \color[HTML]{FFFFFF} 0.77s & {\cellcolor[HTML]{C5E2C5}} \color[HTML]{000000} 77.1\% & {\cellcolor[HTML]{FFD3D3}} \color[HTML]{000000} 945.7\% & {\cellcolor[HTML]{98CC98}} \color[HTML]{000000} 59.5\% & {\cellcolor[HTML]{FFE4E4}} \color[HTML]{000000} 625.7\% & {\cellcolor[HTML]{FF0000}} \color[HTML]{FFFFFF} 23331.4\% & {\cellcolor[HTML]{8EC78E}} \color[HTML]{000000} 55.6\% \\
\toprule
\end{tabular}

%% file: coltables/t-msee-opt.tex
\begin{table}[ht]
    \small
    \begin{center}
    \input{coltables/msee-opt.tex}
    \caption{Problems with known optimum: SSE improvements relative to \gkmp. \tabcom}
    \label{tab:msee-opt}
    \end{center}
\end{table}
\renewcommand{\tabcom}{}

%% file: coltables/msee-opt.tex
\begin{tabular}{lrrrrrrrrr}
\toprule
data set & n & k & \thead{$\Phi($greedy \\ km++$)$} & \thead{vanilla \\ km++} & \thead{better \\ km++} & \thead{Hartigan- \\ Wong} & \thead{genetic \\ algorithm} & \thead{random \\ swap} & \thead{breathing \\ k-means} \\
\midrule
squares-3x3 & 225 & 9 & {\cellcolor[HTML]{BBBBBB}} 2.37e+00 & {\cellcolor[HTML]{FF0000}} \color[HTML]{F1F1F1} -22.4\% &\framebox[35pt]{ ~\hfill {\cellcolor[HTML]{8AC48A}} \color[HTML]{000000} 4.9\% }& {\cellcolor[HTML]{FF0000}} \color[HTML]{F1F1F1} -27.2\% & {\cellcolor[HTML]{8AC48A}} \color[HTML]{000000} 4.9\% & {\cellcolor[HTML]{8AC48A}} \color[HTML]{000000} 4.9\% &\framebox[35pt]{ ~\hfill {\cellcolor[HTML]{8AC48A}} \color[HTML]{000000} 4.9\% }\\
squares-5x5 & 625 & 25 & {\cellcolor[HTML]{BBBBBB}} 2.40e+00 & {\cellcolor[HTML]{FF0000}} \color[HTML]{F1F1F1} -19.8\% & {\cellcolor[HTML]{389C38}} \color[HTML]{F1F1F1} 10.8\% & {\cellcolor[HTML]{FF0000}} \color[HTML]{F1F1F1} -30.5\% & {\cellcolor[HTML]{309830}} \color[HTML]{F1F1F1} 11.4\% & {\cellcolor[HTML]{2E962E}} \color[HTML]{F1F1F1} 11.5\% &\framebox[35pt]{ ~\hfill {\cellcolor[HTML]{2E962E}} \color[HTML]{F1F1F1} 11.5\% }\\
squares-7x7 & 1225 & 49 & {\cellcolor[HTML]{BBBBBB}} 2.36e+00 & {\cellcolor[HTML]{FF0000}} \color[HTML]{F1F1F1} -18.4\% & {\cellcolor[HTML]{4AA44A}} \color[HTML]{F1F1F1} 9.5\% & {\cellcolor[HTML]{FF0000}} \color[HTML]{F1F1F1} -29.9\% & {\cellcolor[HTML]{289328}} \color[HTML]{F1F1F1} 12.0\% & {\cellcolor[HTML]{269226}} \color[HTML]{F1F1F1} 12.1\% &\framebox[35pt]{ ~\hfill {\cellcolor[HTML]{269226}} \color[HTML]{F1F1F1} 12.1\% }\\
angles-3x3 & 1728 & 27 & {\cellcolor[HTML]{BBBBBB}} 4.01e+00 & {\cellcolor[HTML]{FF0808}} \color[HTML]{F1F1F1} -14.3\% & {\cellcolor[HTML]{70B870}} \color[HTML]{F1F1F1} 6.7\% & {\cellcolor[HTML]{FF0000}} \color[HTML]{F1F1F1} -34.3\% & {\cellcolor[HTML]{48A348}} \color[HTML]{F1F1F1} 9.6\% & {\cellcolor[HTML]{42A042}} \color[HTML]{F1F1F1} 10.2\% &\framebox[35pt]{ ~\hfill {\cellcolor[HTML]{42A042}} \color[HTML]{F1F1F1} 10.2\% }\\
angles-5x5 & 2700 & 75 & {\cellcolor[HTML]{BBBBBB}} 2.22e+00 & {\cellcolor[HTML]{FF0000}} \color[HTML]{F1F1F1} -15.2\% & {\cellcolor[HTML]{94CA94}} \color[HTML]{000000} 4.1\% & {\cellcolor[HTML]{FF0000}} \color[HTML]{F1F1F1} -33.8\% & {\cellcolor[HTML]{46A246}} \color[HTML]{F1F1F1} 9.8\% & {\cellcolor[HTML]{3C9E3C}} \color[HTML]{F1F1F1} 10.5\% &\framebox[35pt]{ ~\hfill {\cellcolor[HTML]{3C9E3C}} \color[HTML]{F1F1F1} 10.6\% }\\
angles-7x7 & 3675 & 147 & {\cellcolor[HTML]{BBBBBB}} 1.52e+00 & {\cellcolor[HTML]{FF0808}} \color[HTML]{F1F1F1} -14.2\% & {\cellcolor[HTML]{A6D2A6}} \color[HTML]{000000} 2.9\% & {\cellcolor[HTML]{FF0000}} \color[HTML]{F1F1F1} -35.7\% & {\cellcolor[HTML]{42A042}} \color[HTML]{F1F1F1} 10.2\% & {\cellcolor[HTML]{3C9E3C}} \color[HTML]{F1F1F1} 10.5\% &\framebox[35pt]{ ~\hfill {\cellcolor[HTML]{389C38}} \color[HTML]{F1F1F1} 10.8\% }\\
4squares-3x3 & 2304 & 36 & {\cellcolor[HTML]{BBBBBB}} 6.67e+00 & {\cellcolor[HTML]{FF3838}} \color[HTML]{F1F1F1} -10.9\% & {\cellcolor[HTML]{96CB96}} \color[HTML]{000000} 4.0\% & {\cellcolor[HTML]{FF0000}} \color[HTML]{F1F1F1} -25.4\% & {\cellcolor[HTML]{60B060}} \color[HTML]{F1F1F1} 7.9\% & {\cellcolor[HTML]{54AA54}} \color[HTML]{F1F1F1} 8.7\% &\framebox[35pt]{ ~\hfill {\cellcolor[HTML]{54AA54}} \color[HTML]{F1F1F1} 8.7\% }\\
4squares-5x5 & 3600 & 100 & {\cellcolor[HTML]{BBBBBB}} 3.34e+00 & {\cellcolor[HTML]{FF2424}} \color[HTML]{F1F1F1} -12.3\% & {\cellcolor[HTML]{B0D8B0}} \color[HTML]{000000} 2.1\% & {\cellcolor[HTML]{FF0000}} \color[HTML]{F1F1F1} -28.2\% & {\cellcolor[HTML]{62B062}} \color[HTML]{F1F1F1} 7.8\% & {\cellcolor[HTML]{54AA54}} \color[HTML]{F1F1F1} 8.8\% &\framebox[35pt]{ ~\hfill {\cellcolor[HTML]{54AA54}} \color[HTML]{F1F1F1} 8.8\% }\\
4squares-7x7 & 4900 & 196 & {\cellcolor[HTML]{BBBBBB}} 2.21e+00 & {\cellcolor[HTML]{FF2626}} \color[HTML]{F1F1F1} -12.0\% & {\cellcolor[HTML]{B8DCB8}} \color[HTML]{000000} 1.6\% & {\cellcolor[HTML]{FF0000}} \color[HTML]{F1F1F1} -28.2\% & {\cellcolor[HTML]{5AAC5A}} \color[HTML]{F1F1F1} 8.4\% & {\cellcolor[HTML]{50A850}} \color[HTML]{F1F1F1} 9.0\% &\framebox[35pt]{ ~\hfill {\cellcolor[HTML]{4AA44A}} \color[HTML]{F1F1F1} 9.4\% }\\
\midrule
Mean &  &  &  & {\cellcolor[HTML]{FF0000}} \color[HTML]{F1F1F1} -15.5\% & {\cellcolor[HTML]{86C286}} \color[HTML]{000000} 5.2\% & {\cellcolor[HTML]{FF0000}} \color[HTML]{F1F1F1} -30.4\% & {\cellcolor[HTML]{50A850}} \color[HTML]{F1F1F1} 9.1\% & {\cellcolor[HTML]{4AA44A}} \color[HTML]{F1F1F1} 9.6\% &\framebox[35pt]{ ~\hfill {\cellcolor[HTML]{48A348}} \color[HTML]{F1F1F1} 9.7\% }\\
\toprule
\end{tabular}

%% file: coltables/t-mse-realopt.tex
\begin{table}[t]
    \small
    \begin{center}
    \input{coltables/mse-realopt.tex}
    \caption{Problems with known optimum: SSE deviations from the optimum. \tabcom}
    \label{tab:mse-realopt}
    \end{center}
\end{table}
\renewcommand{\tabcom}{}

%% file: coltables/mse-realopt.tex
\begin{tabular}{lrrrrrrrrr}
\toprule
data set & n & k & \thead{greedy \\ km++} & \thead{vanilla \\ km++} & \thead{better \\ km++} & \thead{Hartigan- \\ Wong} & \thead{genetic \\ algorithm} & \thead{random \\ swap} & \thead{breathing \\ k-means} \\
\midrule
squares-3x3 & 225 & 9 & {\cellcolor[HTML]{9999FF}} {\cellcolor[HTML]{7777FF}} \color{white} 5.18\% & {\cellcolor[HTML]{9999FF}} {\cellcolor[HTML]{7777FF}} \color{white} {\cellcolor[HTML]{5555FF}} \color{white} {\cellcolor[HTML]{0000FF}} \color{white} 28.77\% &\framebox[35pt]{ ~\hfill {\cellcolor[HTML]{A3FBAF}} 0.00\% }& {\cellcolor[HTML]{9999FF}} {\cellcolor[HTML]{7777FF}} \color{white} {\cellcolor[HTML]{5555FF}} \color{white} {\cellcolor[HTML]{0000FF}} \color{white} 33.82\% & {\cellcolor[HTML]{9999FF}} 0.05\% &\framebox[35pt]{ ~\hfill {\cellcolor[HTML]{A3FBAF}} 0.00\% }&\framebox[35pt]{ ~\hfill {\cellcolor[HTML]{A3FBAF}} 0.00\% }\\
squares-5x5 & 625 & 25 & {\cellcolor[HTML]{9999FF}} {\cellcolor[HTML]{7777FF}} \color{white} {\cellcolor[HTML]{5555FF}} \color{white} 12.96\% & {\cellcolor[HTML]{9999FF}} {\cellcolor[HTML]{7777FF}} \color{white} {\cellcolor[HTML]{5555FF}} \color{white} {\cellcolor[HTML]{0000FF}} \color{white} 35.32\% & {\cellcolor[HTML]{9999FF}} 0.81\% & {\cellcolor[HTML]{9999FF}} {\cellcolor[HTML]{7777FF}} \color{white} {\cellcolor[HTML]{5555FF}} \color{white} {\cellcolor[HTML]{0000FF}} \color{white} 47.42\% & {\cellcolor[HTML]{9999FF}} 0.08\% &\framebox[35pt]{ ~\hfill {\cellcolor[HTML]{A3FBAF}} 0.00\% }&\framebox[35pt]{ ~\hfill {\cellcolor[HTML]{A3FBAF}} 0.00\% }\\
squares-7x7 & 1225 & 49 & {\cellcolor[HTML]{9999FF}} {\cellcolor[HTML]{7777FF}} \color{white} {\cellcolor[HTML]{5555FF}} \color{white} 13.76\% & {\cellcolor[HTML]{9999FF}} {\cellcolor[HTML]{7777FF}} \color{white} {\cellcolor[HTML]{5555FF}} \color{white} {\cellcolor[HTML]{0000FF}} \color{white} 34.67\% & {\cellcolor[HTML]{9999FF}} {\cellcolor[HTML]{7777FF}} \color{white} 2.90\% & {\cellcolor[HTML]{9999FF}} {\cellcolor[HTML]{7777FF}} \color{white} {\cellcolor[HTML]{5555FF}} \color{white} {\cellcolor[HTML]{0000FF}} \color{white} 47.79\% & {\cellcolor[HTML]{9999FF}} 0.10\% &\framebox[35pt]{ ~\hfill {\cellcolor[HTML]{A3FBAF}} 0.00\% }&\framebox[35pt]{ ~\hfill {\cellcolor[HTML]{A3FBAF}} 0.00\% }\\
angles-3x3 & 1728 & 27 & {\cellcolor[HTML]{9999FF}} {\cellcolor[HTML]{7777FF}} \color{white} {\cellcolor[HTML]{5555FF}} \color{white} 11.31\% & {\cellcolor[HTML]{9999FF}} {\cellcolor[HTML]{7777FF}} \color{white} {\cellcolor[HTML]{5555FF}} \color{white} {\cellcolor[HTML]{0000FF}} \color{white} 27.23\% & {\cellcolor[HTML]{9999FF}} {\cellcolor[HTML]{7777FF}} \color{white} 3.80\% & {\cellcolor[HTML]{9999FF}} {\cellcolor[HTML]{7777FF}} \color{white} {\cellcolor[HTML]{5555FF}} \color{white} {\cellcolor[HTML]{0000FF}} \color{white} 49.51\% & {\cellcolor[HTML]{9999FF}} 0.56\% &\framebox[35pt]{ ~\hfill {\cellcolor[HTML]{A3FBAF}} 0.00\% }&\framebox[35pt]{ ~\hfill {\cellcolor[HTML]{A3FBAF}} 0.00\% }\\
angles-5x5 & 2700 & 75 & {\cellcolor[HTML]{9999FF}} {\cellcolor[HTML]{7777FF}} \color{white} {\cellcolor[HTML]{5555FF}} \color{white} 11.81\% & {\cellcolor[HTML]{9999FF}} {\cellcolor[HTML]{7777FF}} \color{white} {\cellcolor[HTML]{5555FF}} \color{white} {\cellcolor[HTML]{0000FF}} \color{white} 28.75\% & {\cellcolor[HTML]{9999FF}} {\cellcolor[HTML]{7777FF}} \color{white} 7.17\% & {\cellcolor[HTML]{9999FF}} {\cellcolor[HTML]{7777FF}} \color{white} {\cellcolor[HTML]{5555FF}} \color{white} {\cellcolor[HTML]{0000FF}} \color{white} 49.63\% & {\cellcolor[HTML]{9999FF}} 0.83\% & {\cellcolor[HTML]{9999FF}} 0.06\% &\framebox[35pt]{ ~\hfill {\cellcolor[HTML]{A3FBAF}} 0.00\% }\\
angles-7x7 & 3675 & 147 & {\cellcolor[HTML]{9999FF}} {\cellcolor[HTML]{7777FF}} \color{white} {\cellcolor[HTML]{5555FF}} \color{white} 12.05\% & {\cellcolor[HTML]{9999FF}} {\cellcolor[HTML]{7777FF}} \color{white} {\cellcolor[HTML]{5555FF}} \color{white} {\cellcolor[HTML]{0000FF}} \color{white} 27.94\% & {\cellcolor[HTML]{9999FF}} {\cellcolor[HTML]{7777FF}} \color{white} 8.80\% & {\cellcolor[HTML]{9999FF}} {\cellcolor[HTML]{7777FF}} \color{white} {\cellcolor[HTML]{5555FF}} \color{white} {\cellcolor[HTML]{0000FF}} \color{white} 52.04\% & {\cellcolor[HTML]{9999FF}} 0.67\% & {\cellcolor[HTML]{9999FF}} 0.28\% &\framebox[35pt]{ ~\hfill {\cellcolor[HTML]{A3FBAF}} 0.00\% }\\
4squares-3x3 & 2304 & 36 & {\cellcolor[HTML]{9999FF}} {\cellcolor[HTML]{7777FF}} \color{white} 9.51\% & {\cellcolor[HTML]{9999FF}} {\cellcolor[HTML]{7777FF}} \color{white} {\cellcolor[HTML]{5555FF}} \color{white} {\cellcolor[HTML]{0000FF}} \color{white} 21.44\% & {\cellcolor[HTML]{9999FF}} {\cellcolor[HTML]{7777FF}} \color{white} 5.17\% & {\cellcolor[HTML]{9999FF}} {\cellcolor[HTML]{7777FF}} \color{white} {\cellcolor[HTML]{5555FF}} \color{white} {\cellcolor[HTML]{0000FF}} \color{white} 37.35\% & {\cellcolor[HTML]{9999FF}} 0.85\% &\framebox[35pt]{ ~\hfill {\cellcolor[HTML]{A3FBAF}} 0.00\% }&\framebox[35pt]{ ~\hfill {\cellcolor[HTML]{A3FBAF}} 0.00\% }\\
4squares-5x5 & 3600 & 100 & {\cellcolor[HTML]{9999FF}} {\cellcolor[HTML]{7777FF}} \color{white} 9.67\% & {\cellcolor[HTML]{9999FF}} {\cellcolor[HTML]{7777FF}} \color{white} {\cellcolor[HTML]{5555FF}} \color{white} {\cellcolor[HTML]{0000FF}} \color{white} 23.21\% & {\cellcolor[HTML]{9999FF}} {\cellcolor[HTML]{7777FF}} \color{white} 7.32\% & {\cellcolor[HTML]{9999FF}} {\cellcolor[HTML]{7777FF}} \color{white} {\cellcolor[HTML]{5555FF}} \color{white} {\cellcolor[HTML]{0000FF}} \color{white} 40.56\% & {\cellcolor[HTML]{9999FF}} {\cellcolor[HTML]{7777FF}} \color{white} 1.09\% & {\cellcolor[HTML]{9999FF}} 0.07\% &\framebox[35pt]{ ~\hfill {\cellcolor[HTML]{A3FBAF}} 0.00\% }\\
4squares-7x7 & 4900 & 196 & {\cellcolor[HTML]{9999FF}} {\cellcolor[HTML]{7777FF}} \color{white} {\cellcolor[HTML]{5555FF}} \color{white} 10.42\% & {\cellcolor[HTML]{9999FF}} {\cellcolor[HTML]{7777FF}} \color{white} {\cellcolor[HTML]{5555FF}} \color{white} {\cellcolor[HTML]{0000FF}} \color{white} 23.70\% & {\cellcolor[HTML]{9999FF}} {\cellcolor[HTML]{7777FF}} \color{white} 8.66\% & {\cellcolor[HTML]{9999FF}} {\cellcolor[HTML]{7777FF}} \color{white} {\cellcolor[HTML]{5555FF}} \color{white} {\cellcolor[HTML]{0000FF}} \color{white} 41.58\% & {\cellcolor[HTML]{9999FF}} {\cellcolor[HTML]{7777FF}} \color{white} 1.16\% & {\cellcolor[HTML]{9999FF}} 0.43\% &\framebox[35pt]{ ~\hfill {\cellcolor[HTML]{A3FBAF}} 0.00\% }\\
\midrule
Mean &  &  & {\cellcolor[HTML]{9999FF}} {\cellcolor[HTML]{7777FF}} \color{white} {\cellcolor[HTML]{5555FF}} \color{white} 10.74\% & {\cellcolor[HTML]{9999FF}} {\cellcolor[HTML]{7777FF}} \color{white} {\cellcolor[HTML]{5555FF}} \color{white} {\cellcolor[HTML]{0000FF}} \color{white} 27.89\% & {\cellcolor[HTML]{9999FF}} {\cellcolor[HTML]{7777FF}} \color{white} 4.96\% & {\cellcolor[HTML]{9999FF}} {\cellcolor[HTML]{7777FF}} \color{white} {\cellcolor[HTML]{5555FF}} \color{white} {\cellcolor[HTML]{0000FF}} \color{white} 44.41\% & {\cellcolor[HTML]{9999FF}} 0.60\% & {\cellcolor[HTML]{9999FF}} 0.09\% &\framebox[35pt]{ ~\hfill {\cellcolor[HTML]{A3FBAF}} 0.00\% }\\
\toprule
\end{tabular}

%% file: coltables/t-msee-fraenti.tex
\begin{table}[ht]
    \small
    \begin{center}
    \input{coltables/msee-fraenti.tex}
    \caption{Literature problems: SSE improvements relative to \gkmp. \tabcom}
    \label{tab:msee-fränti}
    \end{center}
\end{table}
\renewcommand{\tabcom}{}

%% file: coltables/msee-fraenti.tex
\begin{tabular}{lrrrrrrrrr}
\toprule
data set & n & k & \thead{$\Phi($greedy \\ km++$)$} & \thead{vanilla \\ km++} & \thead{better \\ km++} & \thead{Hartigan- \\ Wong} & \thead{genetic \\ algorithm} & \thead{random \\ swap} & \thead{breathing \\ k-means} \\
\midrule
aggregation & 788 & 200 & {\cellcolor[HTML]{BBBBBB}} 2.55e+02 & {\cellcolor[HTML]{FF5E5E}} \color[HTML]{F1F1F1} -8.1\% & {\cellcolor[HTML]{BCDEBC}} \color[HTML]{000000} 1.1\% & {\cellcolor[HTML]{BCDEBC}} \color[HTML]{000000} 1.4\% & {\cellcolor[HTML]{389C38}} \color[HTML]{F1F1F1} 10.9\% &\framebox[35pt]{ ~\hfill {\cellcolor[HTML]{2A942A}} \color[HTML]{F1F1F1} 11.8\% }& {\cellcolor[HTML]{5AAC5A}} \color[HTML]{F1F1F1} 8.4\% \\
compound & 399 & 50 & {\cellcolor[HTML]{BBBBBB}} 4.08e+02 & {\cellcolor[HTML]{FF8888}} \color[HTML]{F1F1F1} -5.0\% & {\cellcolor[HTML]{C0E0C0}} \color[HTML]{000000} 0.9\% & {\cellcolor[HTML]{FF4646}} \color[HTML]{F1F1F1} -9.8\% & {\cellcolor[HTML]{5EAE5E}} \color[HTML]{F1F1F1} 8.2\% &\framebox[35pt]{ ~\hfill {\cellcolor[HTML]{3C9E3C}} \color[HTML]{F1F1F1} 10.5\% }& {\cellcolor[HTML]{5EAE5E}} \color[HTML]{F1F1F1} 8.0\% \\
d31 & 3100 & 100 & {\cellcolor[HTML]{BBBBBB}} 1.39e+03 & {\cellcolor[HTML]{FF8A8A}} \color[HTML]{000000} -4.9\% & {\cellcolor[HTML]{BCDEBC}} \color[HTML]{000000} 1.2\% & {\cellcolor[HTML]{FF5858}} \color[HTML]{F1F1F1} -8.4\% & {\cellcolor[HTML]{98CC98}} \color[HTML]{000000} 3.8\% &\framebox[35pt]{ ~\hfill {\cellcolor[HTML]{80C080}} \color[HTML]{000000} 5.6\% }& {\cellcolor[HTML]{8AC48A}} \color[HTML]{000000} 4.9\% \\
flame & 240 & 80 & {\cellcolor[HTML]{BBBBBB}} 4.99e+01 & {\cellcolor[HTML]{FF5858}} \color[HTML]{F1F1F1} -8.5\% & {\cellcolor[HTML]{B4DAB4}} \color[HTML]{000000} 1.8\% & {\cellcolor[HTML]{6AB46A}} \color[HTML]{F1F1F1} 7.1\% & {\cellcolor[HTML]{188C18}} \color[HTML]{F1F1F1} 13.2\% &\framebox[35pt]{ ~\hfill {\cellcolor[HTML]{068206}} \color[HTML]{F1F1F1} 14.5\% }& {\cellcolor[HTML]{2C962C}} \color[HTML]{F1F1F1} 11.7\% \\
jain & 373 & 30 & {\cellcolor[HTML]{BBBBBB}} 6.31e+02 & {\cellcolor[HTML]{FF9292}} \color[HTML]{000000} -4.3\% & {\cellcolor[HTML]{A8D3A8}} \color[HTML]{000000} 2.7\% & {\cellcolor[HTML]{FF0000}} \color[HTML]{F1F1F1} -24.2\% & {\cellcolor[HTML]{68B368}} \color[HTML]{F1F1F1} 7.2\% &\framebox[35pt]{ ~\hfill {\cellcolor[HTML]{50A850}} \color[HTML]{F1F1F1} 9.0\% }& {\cellcolor[HTML]{66B266}} \color[HTML]{F1F1F1} 7.5\% \\
pathbased & 300 & 50 & {\cellcolor[HTML]{BBBBBB}} 3.00e+02 & {\cellcolor[HTML]{FF4C4C}} \color[HTML]{F1F1F1} -9.3\% & {\cellcolor[HTML]{A0D0A0}} \color[HTML]{000000} 3.3\% & {\cellcolor[HTML]{FF5858}} \color[HTML]{F1F1F1} -8.5\% & {\cellcolor[HTML]{2E962E}} \color[HTML]{F1F1F1} 11.5\% &\framebox[35pt]{ ~\hfill {\cellcolor[HTML]{1C8E1C}} \color[HTML]{F1F1F1} 12.8\% }& {\cellcolor[HTML]{42A042}} \color[HTML]{F1F1F1} 10.0\% \\
r15 & 600 & 30 & {\cellcolor[HTML]{BBBBBB}} 7.04e+01 & {\cellcolor[HTML]{FFA6A6}} \color[HTML]{000000} -2.9\% & {\cellcolor[HTML]{A2D0A2}} \color[HTML]{000000} 3.1\% & {\cellcolor[HTML]{FF7A7A}} \color[HTML]{F1F1F1} -6.0\% & {\cellcolor[HTML]{76BB76}} \color[HTML]{F1F1F1} 6.3\% &\framebox[35pt]{ ~\hfill {\cellcolor[HTML]{62B062}} \color[HTML]{F1F1F1} 7.7\% }& {\cellcolor[HTML]{72B872}} \color[HTML]{F1F1F1} 6.6\% \\
s2 & 5000 & 100 & {\cellcolor[HTML]{BBBBBB}} 2.70e+12 & {\cellcolor[HTML]{FFA4A4}} \color[HTML]{000000} -3.0\% & {\cellcolor[HTML]{C4E2C4}} \color[HTML]{000000} 0.7\% & {\cellcolor[HTML]{FF7A7A}} \color[HTML]{F1F1F1} -6.1\% & {\cellcolor[HTML]{AED6AE}} \color[HTML]{000000} 2.3\% &\framebox[35pt]{ ~\hfill {\cellcolor[HTML]{8EC68E}} \color[HTML]{000000} 4.5\% }& {\cellcolor[HTML]{9CCE9C}} \color[HTML]{000000} 3.6\% \\
spiral & 312 & 80 & {\cellcolor[HTML]{BBBBBB}} 1.31e+02 & {\cellcolor[HTML]{FF1C1C}} \color[HTML]{F1F1F1} -12.8\% & {\cellcolor[HTML]{BCDEBC}} \color[HTML]{000000} 1.3\% & {\cellcolor[HTML]{FF0000}} \color[HTML]{F1F1F1} -28.5\% & {\cellcolor[HTML]{68B368}} \color[HTML]{F1F1F1} 7.5\% &\framebox[35pt]{ ~\hfill {\cellcolor[HTML]{46A246}} \color[HTML]{F1F1F1} 9.8\% }& {\cellcolor[HTML]{6CB66C}} \color[HTML]{F1F1F1} 7.0\% \\
\midrule
Mean &  &  &  & {\cellcolor[HTML]{FF7272}} \color[HTML]{F1F1F1} -6.5\% & {\cellcolor[HTML]{B4DAB4}} \color[HTML]{000000} 1.8\% & {\cellcolor[HTML]{FF4E4E}} \color[HTML]{F1F1F1} -9.2\% & {\cellcolor[HTML]{60B060}} \color[HTML]{F1F1F1} 7.9\% &\framebox[35pt]{ ~\hfill {\cellcolor[HTML]{4AA44A}} \color[HTML]{F1F1F1} 9.6\% }& {\cellcolor[HTML]{66B266}} \color[HTML]{F1F1F1} 7.5\% \\
\toprule
\end{tabular}

%% file: coltables/t-msee-fr-mod.tex
\begin{table}[ht]
    \small
    \begin{center}
    \input{coltables/msee-fr-mod.tex}
    \caption{Modified literature problems: SSE improvements relative to \gkmp. \tabcom}
    \label{tab:msee-fr-mod}
    \end{center}
\end{table}
\renewcommand{\tabcom}{}

%% file: coltables/msee-fr-mod.tex
\begin{tabular}{lrrrrrrrrr}
\toprule
data set & n & k & \thead{$\Phi($greedy \\ km++$)$} & \thead{vanilla \\ km++} & \thead{better \\ km++} & \thead{Hartigan- \\ Wong} & \thead{genetic \\ algorithm} & \thead{random \\ swap} & \thead{breathing \\ k-means} \\
\midrule
aggregation-$\ast$ & 4200 & 200 & {\cellcolor[HTML]{BBBBBB}} 1.93e+00 & {\cellcolor[HTML]{FF0000}} \color[HTML]{F1F1F1} -47.2\% & {\cellcolor[HTML]{BCDEBC}} \color[HTML]{000000} 1.3\% & {\cellcolor[HTML]{FF0000}} \color[HTML]{F1F1F1} -107733.9\% & {\cellcolor[HTML]{FF0000}} \color[HTML]{F1F1F1} -1170.0\% & {\cellcolor[HTML]{FF0000}} \color[HTML]{F1F1F1} -1877.7\% &\framebox[35pt]{ ~\hfill {\cellcolor[HTML]{92C892}} \color[HTML]{000000} 4.3\% }\\
compound-$\ast$ & 4200 & 50 & {\cellcolor[HTML]{BBBBBB}} 1.40e+02 & {\cellcolor[HTML]{FF2E2E}} \color[HTML]{F1F1F1} -11.5\% & {\cellcolor[HTML]{A8D3A8}} \color[HTML]{000000} 2.8\% & {\cellcolor[HTML]{FF0000}} \color[HTML]{F1F1F1} -1850.0\% & {\cellcolor[HTML]{FFA6A6}} \color[HTML]{000000} -2.9\% & {\cellcolor[HTML]{9CCE9C}} \color[HTML]{000000} 3.6\% &\framebox[35pt]{ ~\hfill {\cellcolor[HTML]{389C38}} \color[HTML]{F1F1F1} 10.8\% }\\
d31-$\ast$ & 4200 & 100 & {\cellcolor[HTML]{BBBBBB}} 2.72e+01 & {\cellcolor[HTML]{FF0000}} \color[HTML]{F1F1F1} -30.0\% & {\cellcolor[HTML]{9ECE9E}} \color[HTML]{000000} 3.5\% & {\cellcolor[HTML]{FF0000}} \color[HTML]{F1F1F1} -10598.7\% & {\cellcolor[HTML]{FF0000}} \color[HTML]{F1F1F1} -27.2\% & {\cellcolor[HTML]{FF0000}} \color[HTML]{F1F1F1} -56.1\% &\framebox[35pt]{ ~\hfill {\cellcolor[HTML]{329832}} \color[HTML]{F1F1F1} 11.3\% }\\
flame-$\ast$ & 4200 & 80 & {\cellcolor[HTML]{BBBBBB}} 3.72e+01 & {\cellcolor[HTML]{FF3E3E}} \color[HTML]{F1F1F1} -10.4\% & {\cellcolor[HTML]{C6E2C6}} \color[HTML]{000000} 0.6\% & {\cellcolor[HTML]{FF0000}} \color[HTML]{F1F1F1} -1842.0\% & {\cellcolor[HTML]{389C38}} \color[HTML]{F1F1F1} 10.8\% & {\cellcolor[HTML]{FF9696}} \color[HTML]{000000} -4.1\% &\framebox[35pt]{ ~\hfill {\cellcolor[HTML]{148A14}} \color[HTML]{F1F1F1} 13.4\% }\\
jain-$\ast$ & 4200 & 30 & {\cellcolor[HTML]{BBBBBB}} 3.29e+02 & {\cellcolor[HTML]{FF5C5C}} \color[HTML]{F1F1F1} -8.2\% & {\cellcolor[HTML]{A8D3A8}} \color[HTML]{000000} 2.8\% & {\cellcolor[HTML]{FF0000}} \color[HTML]{F1F1F1} -1514.9\% & {\cellcolor[HTML]{FFBCBC}} \color[HTML]{000000} -1.2\% & {\cellcolor[HTML]{5EAE5E}} \color[HTML]{F1F1F1} 8.1\% &\framebox[35pt]{ ~\hfill {\cellcolor[HTML]{50A850}} \color[HTML]{F1F1F1} 9.1\% }\\
pathbased-$\ast$ & 4200 & 50 & {\cellcolor[HTML]{BBBBBB}} 1.68e+02 & {\cellcolor[HTML]{FF0000}} \color[HTML]{F1F1F1} -15.2\% & {\cellcolor[HTML]{92C892}} \color[HTML]{000000} 4.3\% & {\cellcolor[HTML]{FF0000}} \color[HTML]{F1F1F1} -1947.9\% & {\cellcolor[HTML]{7CBE7C}} \color[HTML]{000000} 5.9\% & {\cellcolor[HTML]{7EBE7E}} \color[HTML]{000000} 5.8\% &\framebox[35pt]{ ~\hfill {\cellcolor[HTML]{2C962C}} \color[HTML]{F1F1F1} 11.7\% }\\
r15-$\ast$ & 4200 & 30 & {\cellcolor[HTML]{BBBBBB}} 1.99e+01 & {\cellcolor[HTML]{FF7A7A}} \color[HTML]{F1F1F1} -6.0\% & {\cellcolor[HTML]{8CC68C}} \color[HTML]{000000} 4.7\% & {\cellcolor[HTML]{FF0000}} \color[HTML]{F1F1F1} -5888.4\% & {\cellcolor[HTML]{FF0000}} \color[HTML]{F1F1F1} -55.0\% & {\cellcolor[HTML]{54AA54}} \color[HTML]{F1F1F1} 8.8\% &\framebox[35pt]{ ~\hfill {\cellcolor[HTML]{50A850}} \color[HTML]{F1F1F1} 9.0\% }\\
s2-$\ast$ & 4200 & 100 & {\cellcolor[HTML]{BBBBBB}} 2.55e+10 & {\cellcolor[HTML]{FF0000}} \color[HTML]{F1F1F1} -28.9\% & {\cellcolor[HTML]{AAD4AA}} \color[HTML]{000000} 2.5\% & {\cellcolor[HTML]{FF0000}} \color[HTML]{F1F1F1} -13886.6\% & {\cellcolor[HTML]{FF0000}} \color[HTML]{F1F1F1} -32.3\% & {\cellcolor[HTML]{FF0000}} \color[HTML]{F1F1F1} -68.2\% &\framebox[35pt]{ ~\hfill {\cellcolor[HTML]{4EA64E}} \color[HTML]{F1F1F1} 9.2\% }\\
spiral-$\ast$ & 4200 & 80 & {\cellcolor[HTML]{BBBBBB}} 5.56e+01 & {\cellcolor[HTML]{FF0000}} \color[HTML]{F1F1F1} -19.0\% & {\cellcolor[HTML]{AED6AE}} \color[HTML]{000000} 2.3\% & {\cellcolor[HTML]{FF0000}} \color[HTML]{F1F1F1} -5802.3\% & {\cellcolor[HTML]{FF9696}} \color[HTML]{000000} -4.1\% & {\cellcolor[HTML]{FF0808}} \color[HTML]{F1F1F1} -14.2\% &\framebox[35pt]{ ~\hfill {\cellcolor[HTML]{369B36}} \color[HTML]{F1F1F1} 11.0\% }\\
\midrule
Mean &  &  &  & {\cellcolor[HTML]{FF0000}} \color[HTML]{F1F1F1} -19.6\% & {\cellcolor[HTML]{A8D3A8}} \color[HTML]{000000} 2.8\% & {\cellcolor[HTML]{FF0000}} \color[HTML]{F1F1F1} -16785.0\% & {\cellcolor[HTML]{FF0000}} \color[HTML]{F1F1F1} -141.8\% & {\cellcolor[HTML]{FF0000}} \color[HTML]{F1F1F1} -221.6\% &\framebox[35pt]{ ~\hfill {\cellcolor[HTML]{44A244}} \color[HTML]{F1F1F1} 10.0\% }\\
\toprule
\end{tabular}

%% file: coltables/t-msee-espiralX.tex
\begin{table}[ht]
    \small
    \begin{center}
    \input{coltables/msee-espiralX.tex}
    \caption{"Evil Spiral" problems: SSE improvements relative to \gkmp. \tabcom}
    \label{tab:msee-espiralX}
    \end{center}
\end{table}
\renewcommand{\tabcom}{}

%% file: coltables/msee-espiralX.tex
\begin{tabular}{lrrrrrrrrr}
\toprule
data set & n & k & \thead{$\Phi($greedy \\ km++$)$} & \thead{vanilla \\ km++} & \thead{better \\ km++} & \thead{Hartigan- \\ Wong} & \thead{genetic \\ algorithm} & \thead{random \\ swap} & \thead{breathing \\ k-means} \\
\midrule
espiral01 & 1000 & 100 & {\cellcolor[HTML]{BBBBBB}} 2.08e-01 & {\cellcolor[HTML]{FF1C1C}} \color[HTML]{F1F1F1} -12.8\% & {\cellcolor[HTML]{9ECE9E}} \color[HTML]{000000} 3.5\% & {\cellcolor[HTML]{FF0000}} \color[HTML]{F1F1F1} -325.8\% & {\cellcolor[HTML]{389C38}} \color[HTML]{F1F1F1} 10.8\% &\framebox[35pt]{ ~\hfill {\cellcolor[HTML]{309830}} \color[HTML]{F1F1F1} 11.3\% }& {\cellcolor[HTML]{44A244}} \color[HTML]{F1F1F1} 9.9\% \\
espiral04 & 2500 & 100 & {\cellcolor[HTML]{BBBBBB}} 2.17e-01 & {\cellcolor[HTML]{FF1212}} \color[HTML]{F1F1F1} -13.7\% & {\cellcolor[HTML]{AAD4AA}} \color[HTML]{000000} 2.5\% & {\cellcolor[HTML]{FF0000}} \color[HTML]{F1F1F1} -2490.5\% &\framebox[35pt]{ ~\hfill {\cellcolor[HTML]{48A348}} \color[HTML]{F1F1F1} 9.7\% }& {\cellcolor[HTML]{4CA64C}} \color[HTML]{F1F1F1} 9.4\% & {\cellcolor[HTML]{4CA64C}} \color[HTML]{F1F1F1} 9.3\% \\
espiral07 & 4000 & 100 & {\cellcolor[HTML]{BBBBBB}} 2.29e-01 & {\cellcolor[HTML]{FF2828}} \color[HTML]{F1F1F1} -12.0\% & {\cellcolor[HTML]{A6D2A6}} \color[HTML]{000000} 2.9\% & {\cellcolor[HTML]{FF0000}} \color[HTML]{F1F1F1} -3357.6\% & {\cellcolor[HTML]{68B368}} \color[HTML]{F1F1F1} 7.3\% & {\cellcolor[HTML]{56AB56}} \color[HTML]{F1F1F1} 8.6\% &\framebox[35pt]{ ~\hfill {\cellcolor[HTML]{46A246}} \color[HTML]{F1F1F1} 9.7\% }\\
espiral10 & 5500 & 100 & {\cellcolor[HTML]{BBBBBB}} 2.40e-01 & {\cellcolor[HTML]{FF2424}} \color[HTML]{F1F1F1} -12.3\% & {\cellcolor[HTML]{AAD4AA}} \color[HTML]{000000} 2.7\% & {\cellcolor[HTML]{FF0000}} \color[HTML]{F1F1F1} -3631.6\% & {\cellcolor[HTML]{98CC98}} \color[HTML]{000000} 3.8\% & {\cellcolor[HTML]{6CB66C}} \color[HTML]{F1F1F1} 7.1\% &\framebox[35pt]{ ~\hfill {\cellcolor[HTML]{52A852}} \color[HTML]{F1F1F1} 9.0\% }\\
espiral13 & 7000 & 100 & {\cellcolor[HTML]{BBBBBB}} 2.52e-01 & {\cellcolor[HTML]{FF2626}} \color[HTML]{F1F1F1} -12.0\% & {\cellcolor[HTML]{A8D3A8}} \color[HTML]{000000} 2.7\% & {\cellcolor[HTML]{FF0000}} \color[HTML]{F1F1F1} -2722.3\% & {\cellcolor[HTML]{CCE6CC}} \color[HTML]{000000} 0.1\% & {\cellcolor[HTML]{7ABC7A}} \color[HTML]{000000} 6.0\% &\framebox[35pt]{ ~\hfill {\cellcolor[HTML]{54AA54}} \color[HTML]{F1F1F1} 8.7\% }\\
espiral16 & 8500 & 100 & {\cellcolor[HTML]{BBBBBB}} 2.66e-01 & {\cellcolor[HTML]{FF3A3A}} \color[HTML]{F1F1F1} -10.7\% & {\cellcolor[HTML]{A4D2A4}} \color[HTML]{000000} 3.0\% & {\cellcolor[HTML]{FF0000}} \color[HTML]{F1F1F1} -1527.0\% & {\cellcolor[HTML]{FF9E9E}} \color[HTML]{000000} -3.4\% & {\cellcolor[HTML]{84C284}} \color[HTML]{000000} 5.3\% &\framebox[35pt]{ ~\hfill {\cellcolor[HTML]{50A850}} \color[HTML]{F1F1F1} 9.1\% }\\
espiral19 & 10000 & 100 & {\cellcolor[HTML]{BBBBBB}} 2.79e-01 & {\cellcolor[HTML]{FF3232}} \color[HTML]{F1F1F1} -11.2\% & {\cellcolor[HTML]{AED6AE}} \color[HTML]{000000} 2.3\% & {\cellcolor[HTML]{FF0000}} \color[HTML]{F1F1F1} -1251.7\% & {\cellcolor[HTML]{FF9C9C}} \color[HTML]{000000} -3.5\% & {\cellcolor[HTML]{94CA94}} \color[HTML]{000000} 4.1\% &\framebox[35pt]{ ~\hfill {\cellcolor[HTML]{54AA54}} \color[HTML]{F1F1F1} 8.7\% }\\
espiral22 & 11500 & 100 & {\cellcolor[HTML]{BBBBBB}} 2.94e-01 & {\cellcolor[HTML]{FF3838}} \color[HTML]{F1F1F1} -10.8\% & {\cellcolor[HTML]{ACD6AC}} \color[HTML]{000000} 2.4\% & {\cellcolor[HTML]{FF0000}} \color[HTML]{F1F1F1} -1001.2\% & {\cellcolor[HTML]{FF3232}} \color[HTML]{F1F1F1} -11.2\% & {\cellcolor[HTML]{AAD4AA}} \color[HTML]{000000} 2.6\% &\framebox[35pt]{ ~\hfill {\cellcolor[HTML]{58AC58}} \color[HTML]{F1F1F1} 8.5\% }\\
espiral25 & 13000 & 100 & {\cellcolor[HTML]{BBBBBB}} 3.12e-01 & {\cellcolor[HTML]{FF3A3A}} \color[HTML]{F1F1F1} -10.7\% & {\cellcolor[HTML]{A2D0A2}} \color[HTML]{000000} 3.1\% & {\cellcolor[HTML]{FF0000}} \color[HTML]{F1F1F1} -890.3\% & {\cellcolor[HTML]{FFA2A2}} \color[HTML]{000000} -3.1\% & {\cellcolor[HTML]{A0D0A0}} \color[HTML]{000000} 3.2\% &\framebox[35pt]{ ~\hfill {\cellcolor[HTML]{56AB56}} \color[HTML]{F1F1F1} 8.6\% }\\
\midrule
Mean &  &  &  & {\cellcolor[HTML]{FF2A2A}} \color[HTML]{F1F1F1} -11.8\% & {\cellcolor[HTML]{A8D3A8}} \color[HTML]{000000} 2.8\% & {\cellcolor[HTML]{FF0000}} \color[HTML]{F1F1F1} -1910.9\% & {\cellcolor[HTML]{BCDEBC}} \color[HTML]{000000} 1.2\% & {\cellcolor[HTML]{74BA74}} \color[HTML]{F1F1F1} 6.4\% &\framebox[35pt]{ ~\hfill {\cellcolor[HTML]{50A850}} \color[HTML]{F1F1F1} 9.1\% }\\
\toprule
\end{tabular}

%% file: coltables/t-msee-high-D.tex
\begin{table}[ht]
    \small
    \begin{center}
    \input{coltables/msee-high-D.tex}
    \caption{High-dimensional problems: SSE improvements relative to \gkmp. \tabcom}
    \label{tab:msee-high-D}
    \end{center}
\end{table}
\renewcommand{\tabcom}{}

%% file: coltables/msee-high-D.tex
\begin{tabular}{lrrrrrrrrr}
\toprule
data set & n & k & \thead{$\Phi($greedy \\ km++$)$} & \thead{vanilla \\ km++} & \thead{better \\ km++} & \thead{Hartigan- \\ Wong} & \thead{genetic \\ algorithm} & \thead{random \\ swap} & \thead{breathing \\ k-means} \\
\midrule
Norm25 & 10000 & 10 & {\cellcolor[HTML]{BBBBBB}} 1.15e+09 & {\cellcolor[HTML]{FF8484}} \color[HTML]{F1F1F1} -5.3\% & {\cellcolor[HTML]{B4DAB4}} \color[HTML]{000000} 1.9\% & {\cellcolor[HTML]{FF0404}} \color[HTML]{F1F1F1} -14.5\% &\framebox[35pt]{ ~\hfill {\cellcolor[HTML]{42A042}} \color[HTML]{F1F1F1} 10.2\% }& {\cellcolor[HTML]{4CA64C}} \color[HTML]{F1F1F1} 9.4\% & {\cellcolor[HTML]{5EAE5E}} \color[HTML]{F1F1F1} 8.1\% \\
Norm25 & 10000 & 25 & {\cellcolor[HTML]{BBBBBB}} 1.49e+05 &\framebox[35pt]{ ~\hfill {\cellcolor[HTML]{FFFF00}} \color[HTML]{000000} 0.0\% }&\framebox[35pt]{ ~\hfill {\cellcolor[HTML]{FFFF00}} \color[HTML]{000000} 0.0\% }& {\cellcolor[HTML]{FF0000}} \color[HTML]{F1F1F1} -285205.6\% &\framebox[35pt]{ ~\hfill {\cellcolor[HTML]{FFFF00}} \color[HTML]{000000} 0.0\% }&\framebox[35pt]{ ~\hfill {\cellcolor[HTML]{FFFF00}} \color[HTML]{000000} 0.0\% }&\framebox[35pt]{ ~\hfill {\cellcolor[HTML]{FFFF00}} \color[HTML]{000000} 0.0\% }\\
Norm25 & 10000 & 50 & {\cellcolor[HTML]{BBBBBB}} 1.42e+05 & {\cellcolor[HTML]{FFCCCC}} \color[HTML]{000000} -0.1\% & {\cellcolor[HTML]{FFFF00}} \color[HTML]{000000} 0.0\% & {\cellcolor[HTML]{FF0000}} \color[HTML]{F1F1F1} -48063.9\% & {\cellcolor[HTML]{BEDEBE}} \color[HTML]{000000} 1.1\% & {\cellcolor[HTML]{BCDEBC}} \color[HTML]{000000} 1.2\% &\framebox[35pt]{ ~\hfill {\cellcolor[HTML]{BCDEBC}} \color[HTML]{000000} 1.2\% }\\
Norm25 & 10000 & 100 & {\cellcolor[HTML]{BBBBBB}} 1.31e+05 & {\cellcolor[HTML]{FFCACA}} \color[HTML]{000000} -0.3\% & {\cellcolor[HTML]{FFFF00}} \color[HTML]{000000} 0.1\% & {\cellcolor[HTML]{FF0000}} \color[HTML]{F1F1F1} -1093.2\% & {\cellcolor[HTML]{C6E2C6}} \color[HTML]{000000} 0.6\% &\framebox[35pt]{ ~\hfill {\cellcolor[HTML]{BADCBA}} \color[HTML]{000000} 1.4\% }& {\cellcolor[HTML]{BCDEBC}} \color[HTML]{000000} 1.2\% \\
Norm25 & 10000 & 200 & {\cellcolor[HTML]{BBBBBB}} 1.18e+05 & {\cellcolor[HTML]{FFCACA}} \color[HTML]{000000} -0.3\% & {\cellcolor[HTML]{CCE6CC}} \color[HTML]{000000} 0.1\% & {\cellcolor[HTML]{C0E0C0}} \color[HTML]{000000} 0.9\% & {\cellcolor[HTML]{CCE6CC}} \color[HTML]{000000} 0.1\% &\framebox[35pt]{ ~\hfill {\cellcolor[HTML]{B6DBB6}} \color[HTML]{000000} 1.7\% }& {\cellcolor[HTML]{BCDEBC}} \color[HTML]{000000} 1.2\% \\
Cloud & 1024 & 10 & {\cellcolor[HTML]{BBBBBB}} 6.08e+06 & {\cellcolor[HTML]{FF9C9C}} \color[HTML]{000000} -3.6\% & {\cellcolor[HTML]{A8D3A8}} \color[HTML]{000000} 2.8\% & {\cellcolor[HTML]{FF7676}} \color[HTML]{F1F1F1} -6.3\% & {\cellcolor[HTML]{A8D3A8}} \color[HTML]{000000} 2.7\% &\framebox[35pt]{ ~\hfill {\cellcolor[HTML]{86C286}} \color[HTML]{000000} 5.2\% }& {\cellcolor[HTML]{8AC48A}} \color[HTML]{000000} 4.9\% \\
Cloud & 1024 & 25 & {\cellcolor[HTML]{BBBBBB}} 2.07e+06 & {\cellcolor[HTML]{FF9C9C}} \color[HTML]{000000} -3.6\% & {\cellcolor[HTML]{B4DAB4}} \color[HTML]{000000} 1.9\% & {\cellcolor[HTML]{FF0000}} \color[HTML]{F1F1F1} -32.9\% & {\cellcolor[HTML]{A4D2A4}} \color[HTML]{000000} 2.9\% &\framebox[35pt]{ ~\hfill {\cellcolor[HTML]{7EBE7E}} \color[HTML]{000000} 5.8\% }& {\cellcolor[HTML]{8AC48A}} \color[HTML]{000000} 4.9\% \\
Cloud & 1024 & 50 & {\cellcolor[HTML]{BBBBBB}} 1.12e+06 & {\cellcolor[HTML]{FF7C7C}} \color[HTML]{F1F1F1} -5.9\% & {\cellcolor[HTML]{B8DCB8}} \color[HTML]{000000} 1.6\% & {\cellcolor[HTML]{FF0000}} \color[HTML]{F1F1F1} -41.5\% & {\cellcolor[HTML]{C6E2C6}} \color[HTML]{000000} 0.6\% &\framebox[35pt]{ ~\hfill {\cellcolor[HTML]{82C082}} \color[HTML]{000000} 5.5\% }& {\cellcolor[HTML]{8CC68C}} \color[HTML]{000000} 4.8\% \\
Cloud & 1024 & 100 & {\cellcolor[HTML]{BBBBBB}} 6.21e+05 & {\cellcolor[HTML]{FF7676}} \color[HTML]{F1F1F1} -6.3\% & {\cellcolor[HTML]{BEDEBE}} \color[HTML]{000000} 1.1\% & {\cellcolor[HTML]{FF0000}} \color[HTML]{F1F1F1} -61.6\% & {\cellcolor[HTML]{A0D0A0}} \color[HTML]{000000} 3.3\% &\framebox[35pt]{ ~\hfill {\cellcolor[HTML]{6AB46A}} \color[HTML]{F1F1F1} 7.1\% }& {\cellcolor[HTML]{80C080}} \color[HTML]{000000} 5.7\% \\
Cloud & 1024 & 200 & {\cellcolor[HTML]{BBBBBB}} 3.06e+05 & {\cellcolor[HTML]{FF3E3E}} \color[HTML]{F1F1F1} -10.3\% & {\cellcolor[HTML]{C4E2C4}} \color[HTML]{000000} 0.8\% & {\cellcolor[HTML]{FF0000}} \color[HTML]{F1F1F1} -64.2\% & {\cellcolor[HTML]{7EBE7E}} \color[HTML]{000000} 5.7\% &\framebox[35pt]{ ~\hfill {\cellcolor[HTML]{5AAC5A}} \color[HTML]{F1F1F1} 8.4\% }& {\cellcolor[HTML]{76BB76}} \color[HTML]{F1F1F1} 6.3\% \\
Spam & 4601 & 10 & {\cellcolor[HTML]{BBBBBB}} 8.00e+07 & {\cellcolor[HTML]{FF0000}} \color[HTML]{F1F1F1} -18.0\% & {\cellcolor[HTML]{A0D0A0}} \color[HTML]{000000} 3.2\% & {\cellcolor[HTML]{FF0000}} \color[HTML]{F1F1F1} -111.8\% & {\cellcolor[HTML]{FF0000}} \color[HTML]{F1F1F1} -56.2\% & {\cellcolor[HTML]{FFA4A4}} \color[HTML]{000000} -3.0\% &\framebox[35pt]{ ~\hfill {\cellcolor[HTML]{98CC98}} \color[HTML]{000000} 3.8\% }\\
Spam & 4601 & 25 & {\cellcolor[HTML]{BBBBBB}} 1.63e+07 & {\cellcolor[HTML]{FF6868}} \color[HTML]{F1F1F1} -7.3\% & {\cellcolor[HTML]{A2D0A2}} \color[HTML]{000000} 3.1\% & {\cellcolor[HTML]{FF0000}} \color[HTML]{F1F1F1} -819.1\% & {\cellcolor[HTML]{FF0000}} \color[HTML]{F1F1F1} -309.6\% & {\cellcolor[HTML]{FF0000}} \color[HTML]{F1F1F1} -30.6\% &\framebox[35pt]{ ~\hfill {\cellcolor[HTML]{82C082}} \color[HTML]{000000} 5.5\% }\\
Spam & 4601 & 50 & {\cellcolor[HTML]{BBBBBB}} 6.14e+06 & {\cellcolor[HTML]{FF5656}} \color[HTML]{F1F1F1} -8.6\% & {\cellcolor[HTML]{A4D2A4}} \color[HTML]{000000} 3.0\% & {\cellcolor[HTML]{FF0000}} \color[HTML]{F1F1F1} -2249.1\% & {\cellcolor[HTML]{FF0000}} \color[HTML]{F1F1F1} -704.1\% & {\cellcolor[HTML]{FF0000}} \color[HTML]{F1F1F1} -29.7\% &\framebox[35pt]{ ~\hfill {\cellcolor[HTML]{7ABC7A}} \color[HTML]{000000} 6.1\% }\\
Spam & 4601 & 100 & {\cellcolor[HTML]{BBBBBB}} 2.13e+06 & {\cellcolor[HTML]{FF1A1A}} \color[HTML]{F1F1F1} -13.1\% & {\cellcolor[HTML]{B2D8B2}} \color[HTML]{000000} 2.0\% & {\cellcolor[HTML]{FF0000}} \color[HTML]{F1F1F1} -5563.6\% & {\cellcolor[HTML]{FF0000}} \color[HTML]{F1F1F1} -1522.5\% & {\cellcolor[HTML]{FF0000}} \color[HTML]{F1F1F1} -234.5\% &\framebox[35pt]{ ~\hfill {\cellcolor[HTML]{80C080}} \color[HTML]{000000} 5.7\% }\\
Spam & 4601 & 200 & {\cellcolor[HTML]{BBBBBB}} 6.78e+05 & {\cellcolor[HTML]{FF0404}} \color[HTML]{F1F1F1} -14.6\% & {\cellcolor[HTML]{BCDEBC}} \color[HTML]{000000} 1.2\% & {\cellcolor[HTML]{FF0000}} \color[HTML]{F1F1F1} -11310.0\% & {\cellcolor[HTML]{FF0000}} \color[HTML]{F1F1F1} -2462.4\% & {\cellcolor[HTML]{FF0000}} \color[HTML]{F1F1F1} -830.6\% &\framebox[35pt]{ ~\hfill {\cellcolor[HTML]{8EC68E}} \color[HTML]{000000} 4.7\% }\\
\midrule
Mean &  &  &  & {\cellcolor[HTML]{FF7474}} \color[HTML]{F1F1F1} -6.5\% & {\cellcolor[HTML]{B8DCB8}} \color[HTML]{000000} 1.5\% & {\cellcolor[HTML]{FF0000}} \color[HTML]{F1F1F1} -23642.4\% & {\cellcolor[HTML]{FF0000}} \color[HTML]{F1F1F1} -335.2\% & {\cellcolor[HTML]{FF0000}} \color[HTML]{F1F1F1} -72.2\% &\framebox[35pt]{ ~\hfill {\cellcolor[HTML]{92C892}} \color[HTML]{000000} 4.3\% }\\
\toprule
\end{tabular}

%% file: coltables/t-cpu-opt.tex
\begin{table}[ht]
    \small
    \begin{center}
    \input{coltables/cpu-opt.tex}
    \caption{Problems with known optimum: CPU time relative to \gkmp. \tabcom}
    \label{tab:cpu-opt}
    \end{center}
\end{table}
\renewcommand{\tabcom}{}

%% file: coltables/cpu-opt.tex
\begin{tabular}{lrrrrrrrrr}
\toprule
data set & n & k & \thead{t(greedy \\ km++)\\Python} & \thead{vanilla \\ km++\\Python} & \thead{better \\ km++\\ \textsf{Python}} & \thead{Hartigan- \\ Wong\\R/Fortran} & \thead{genetic \\ algorithm\\C} & \thead{random \\ swap\\C} & \thead{breathing \\ k-means\\Python} \\
\midrule
squares-3x3 & 225 & 9 & {\cellcolor{blue}} \color[HTML]{FFFFFF} 0.01s & {\cellcolor[HTML]{E8F4E8}} \color[HTML]{000000} 91.0\% & {\cellcolor[HTML]{FFF2F2}} \color[HTML]{000000} 358.5\% & {\cellcolor[HTML]{54AA54}} \color[HTML]{000000} 32.8\% & {\cellcolor[HTML]{FFFEFE}} \color[HTML]{000000} 129.4\% & {\cellcolor[HTML]{FF6A6A}} \color[HTML]{000000} 2955.1\% & {\cellcolor[HTML]{FFF5F5}} \color[HTML]{000000} 298.7\% \\
squares-5x5 & 625 & 25 & {\cellcolor{blue}} \color[HTML]{FFFFFF} 0.02s & {\cellcolor[HTML]{FFFFFF}} \color[HTML]{000000} 108.2\% & {\cellcolor[HTML]{FFE6E6}} \color[HTML]{000000} 579.9\% & {\cellcolor[HTML]{389C38}} \color[HTML]{FFFFFF} 21.9\% & {\cellcolor[HTML]{FFF9F9}} \color[HTML]{000000} 219.4\% & {\cellcolor[HTML]{FF3939}} \color[HTML]{FFFFFF} 3897.9\% & {\cellcolor[HTML]{FFF3F3}} \color[HTML]{000000} 339.3\% \\
squares-7x7 & 1225 & 49 & {\cellcolor{blue}} \color[HTML]{FFFFFF} 0.03s & {\cellcolor[HTML]{DCEEDC}} \color[HTML]{000000} 86.3\% & {\cellcolor[HTML]{FFE2E2}} \color[HTML]{000000} 667.9\% & {\cellcolor[HTML]{259225}} \color[HTML]{FFFFFF} 14.8\% & {\cellcolor[HTML]{FFF8F8}} \color[HTML]{000000} 244.9\% & {\cellcolor[HTML]{FF5151}} \color[HTML]{000000} 3439.5\% & {\cellcolor[HTML]{FFF7F7}} \color[HTML]{000000} 271.7\% \\
angles-3x3 & 1728 & 27 & {\cellcolor{blue}} \color[HTML]{FFFFFF} 0.02s & {\cellcolor[HTML]{EDF6ED}} \color[HTML]{000000} 92.9\% & {\cellcolor[HTML]{FFDCDC}} \color[HTML]{000000} 780.7\% & {\cellcolor[HTML]{339933}} \color[HTML]{FFFFFF} 20.1\% & {\cellcolor[HTML]{FFF5F5}} \color[HTML]{000000} 293.6\% & {\cellcolor[HTML]{FF0000}} \color[HTML]{FFFFFF} 7228.7\% & {\cellcolor[HTML]{FFEAEA}} \color[HTML]{000000} 514.2\% \\
angles-5x5 & 2700 & 75 & {\cellcolor{blue}} \color[HTML]{FFFFFF} 0.07s & {\cellcolor[HTML]{BDDEBD}} \color[HTML]{000000} 74.0\% & {\cellcolor[HTML]{FFDCDC}} \color[HTML]{000000} 773.4\% & {\cellcolor[HTML]{239123}} \color[HTML]{FFFFFF} 14.0\% & {\cellcolor[HTML]{FFF6F6}} \color[HTML]{000000} 289.8\% & {\cellcolor[HTML]{FF3F3F}} \color[HTML]{FFFFFF} 3781.3\% & {\cellcolor[HTML]{FFF2F2}} \color[HTML]{000000} 351.9\% \\
angles-7x7 & 3675 & 147 & {\cellcolor{blue}} \color[HTML]{FFFFFF} 0.12s & {\cellcolor[HTML]{BDDEBD}} \color[HTML]{000000} 74.2\% & {\cellcolor[HTML]{FFCFCF}} \color[HTML]{000000} 1030.7\% & {\cellcolor[HTML]{259225}} \color[HTML]{FFFFFF} 14.7\% & {\cellcolor[HTML]{FFF0F0}} \color[HTML]{000000} 393.8\% & {\cellcolor[HTML]{FF7070}} \color[HTML]{000000} 2839.7\% & {\cellcolor[HTML]{FFEFEF}} \color[HTML]{000000} 407.5\% \\
4squares-3x3 & 2304 & 36 & {\cellcolor{blue}} \color[HTML]{FFFFFF} 0.04s & {\cellcolor[HTML]{DCEEDC}} \color[HTML]{000000} 86.3\% & {\cellcolor[HTML]{FFDEDE}} \color[HTML]{000000} 743.4\% & {\cellcolor[HTML]{2C962C}} \color[HTML]{FFFFFF} 17.4\% & {\cellcolor[HTML]{FFF6F6}} \color[HTML]{000000} 288.8\% & {\cellcolor[HTML]{FF0000}} \color[HTML]{FFFFFF} 7577.4\% & {\cellcolor[HTML]{FFECEC}} \color[HTML]{000000} 475.0\% \\
4squares-5x5 & 3600 & 100 & {\cellcolor{blue}} \color[HTML]{FFFFFF} 0.10s & {\cellcolor[HTML]{BEDFBE}} \color[HTML]{000000} 74.6\% & {\cellcolor[HTML]{FFD7D7}} \color[HTML]{000000} 883.3\% & {\cellcolor[HTML]{259225}} \color[HTML]{FFFFFF} 14.6\% & {\cellcolor[HTML]{FFF3F3}} \color[HTML]{000000} 339.7\% & {\cellcolor[HTML]{FF1C1C}} \color[HTML]{FFFFFF} 4454.3\% & {\cellcolor[HTML]{FFF1F1}} \color[HTML]{000000} 373.8\% \\
4squares-7x7 & 4900 & 196 & {\cellcolor{blue}} \color[HTML]{FFFFFF} 0.22s & {\cellcolor[HTML]{ABD5AB}} \color[HTML]{000000} 66.9\% & {\cellcolor[HTML]{FFD2D2}} \color[HTML]{000000} 975.6\% & {\cellcolor[HTML]{219021}} \color[HTML]{FFFFFF} 13.2\% & {\cellcolor[HTML]{FFF0F0}} \color[HTML]{000000} 404.9\% & {\cellcolor[HTML]{FF7474}} \color[HTML]{000000} 2772.7\% & {\cellcolor[HTML]{FFEFEF}} \color[HTML]{000000} 409.3\% \\
\midrule
Mean &  &  & {\cellcolor{blue}} \color[HTML]{FFFFFF} 0.07s & {\cellcolor[HTML]{D6EBD6}} \color[HTML]{000000} 83.8\% & {\cellcolor[HTML]{FFDDDD}} \color[HTML]{000000} 754.8\% & {\cellcolor[HTML]{2E972E}} \color[HTML]{FFFFFF} 18.2\% & {\cellcolor[HTML]{FFF6F6}} \color[HTML]{000000} 289.4\% & {\cellcolor[HTML]{FF2323}} \color[HTML]{FFFFFF} 4327.4\% & {\cellcolor[HTML]{FFF1F1}} \color[HTML]{000000} 382.4\% \\
\toprule
\end{tabular}

%% file: coltables/t-cpu-fraenti.tex
\begin{table}[ht]
    \small
    \begin{center}
    \input{coltables/cpu-fraenti.tex}
    \caption{Literature problems: CPU time relative to \gkmp. \tabcom}
    \label{tab:cpu-fränti}
    \end{center}
\end{table}
\renewcommand{\tabcom}{}

%% file: coltables/cpu-fraenti.tex
\begin{tabular}{lrrrrrrrrr}
\toprule
data set & n & k & \thead{t(greedy \\ km++)\\Python} & \thead{vanilla \\ km++\\Python} & \thead{better \\ km++\\ \textsf{Python}} & \thead{Hartigan- \\ Wong\\R/Fortran} & \thead{genetic \\ algorithm\\C} & \thead{random \\ swap\\C} & \thead{breathing \\ k-means\\Python} \\
\midrule
aggregation & 788 & 200 & {\cellcolor{blue}} \color[HTML]{FFFFFF} 0.10s & {\cellcolor[HTML]{C2E1C2}} \color[HTML]{000000} 75.8\% & {\cellcolor[HTML]{FFE6E6}} \color[HTML]{000000} 581.5\% & {\cellcolor[HTML]{128912}} \color[HTML]{FFFFFF} 7.0\% & {\cellcolor[HTML]{FFF1F1}} \color[HTML]{000000} 376.9\% & {\cellcolor[HTML]{FFE1E1}} \color[HTML]{000000} 681.7\% & {\cellcolor[HTML]{FFE7E7}} \color[HTML]{000000} 577.3\% \\
compound & 399 & 50 & {\cellcolor{blue}} \color[HTML]{FFFFFF} 0.03s & {\cellcolor[HTML]{EEF7EE}} \color[HTML]{000000} 93.1\% & {\cellcolor[HTML]{FFE9E9}} \color[HTML]{000000} 530.7\% & {\cellcolor[HTML]{229122}} \color[HTML]{FFFFFF} 13.5\% & {\cellcolor[HTML]{FFFAFA}} \color[HTML]{000000} 213.7\% & {\cellcolor[HTML]{FFA4A4}} \color[HTML]{000000} 1842.7\% & {\cellcolor[HTML]{FFEEEE}} \color[HTML]{000000} 443.6\% \\
d31 & 3100 & 100 & {\cellcolor{blue}} \color[HTML]{FFFFFF} 0.11s & {\cellcolor[HTML]{C1E0C1}} \color[HTML]{000000} 75.6\% & {\cellcolor[HTML]{FFDFDF}} \color[HTML]{000000} 727.5\% & {\cellcolor[HTML]{229122}} \color[HTML]{FFFFFF} 13.6\% & {\cellcolor[HTML]{FFF5F5}} \color[HTML]{000000} 307.5\% & {\cellcolor[HTML]{FF2222}} \color[HTML]{FFFFFF} 4342.1\% & {\cellcolor[HTML]{FFECEC}} \color[HTML]{000000} 478.0\% \\
flame & 240 & 80 & {\cellcolor{blue}} \color[HTML]{FFFFFF} 0.04s & {\cellcolor[HTML]{D6EBD6}} \color[HTML]{000000} 83.9\% & {\cellcolor[HTML]{FFEBEB}} \color[HTML]{000000} 493.9\% & {\cellcolor[HTML]{198C19}} \color[HTML]{FFFFFF} 9.9\% & {\cellcolor[HTML]{FFFBFB}} \color[HTML]{000000} 189.3\% & {\cellcolor[HTML]{FFE4E4}} \color[HTML]{000000} 635.6\% & {\cellcolor[HTML]{FFEEEE}} \color[HTML]{000000} 439.2\% \\
jain & 373 & 30 & {\cellcolor{blue}} \color[HTML]{FFFFFF} 0.02s & {\cellcolor[HTML]{D3E9D3}} \color[HTML]{000000} 82.7\% & {\cellcolor[HTML]{FFEEEE}} \color[HTML]{000000} 429.4\% & {\cellcolor[HTML]{279327}} \color[HTML]{FFFFFF} 15.6\% & {\cellcolor[HTML]{FFFDFD}} \color[HTML]{000000} 150.3\% & {\cellcolor[HTML]{FF9393}} \color[HTML]{000000} 2168.7\% & {\cellcolor[HTML]{FFEEEE}} \color[HTML]{000000} 425.9\% \\
pathbased & 300 & 50 & {\cellcolor{blue}} \color[HTML]{FFFFFF} 0.03s & {\cellcolor[HTML]{E5F2E5}} \color[HTML]{000000} 89.6\% & {\cellcolor[HTML]{FFEBEB}} \color[HTML]{000000} 484.3\% & {\cellcolor[HTML]{219021}} \color[HTML]{FFFFFF} 13.2\% & {\cellcolor[HTML]{FFFCFC}} \color[HTML]{000000} 169.2\% & {\cellcolor[HTML]{FFBFBF}} \color[HTML]{000000} 1338.8\% & {\cellcolor[HTML]{FFF1F1}} \color[HTML]{000000} 374.4\% \\
r15 & 600 & 30 & {\cellcolor{blue}} \color[HTML]{FFFFFF} 0.02s & {\cellcolor[HTML]{D1E8D1}} \color[HTML]{000000} 81.7\% & {\cellcolor[HTML]{FFEAEA}} \color[HTML]{000000} 520.3\% & {\cellcolor[HTML]{299429}} \color[HTML]{FFFFFF} 16.4\% & {\cellcolor[HTML]{FFFBFB}} \color[HTML]{000000} 195.7\% & {\cellcolor[HTML]{FF7070}} \color[HTML]{000000} 2843.0\% & {\cellcolor[HTML]{FFEDED}} \color[HTML]{000000} 462.0\% \\
s2 & 5000 & 100 & {\cellcolor{blue}} \color[HTML]{FFFFFF} 0.14s & {\cellcolor[HTML]{D5EAD5}} \color[HTML]{000000} 83.3\% & {\cellcolor[HTML]{FFD9D9}} \color[HTML]{000000} 843.9\% & {\cellcolor[HTML]{319831}} \color[HTML]{FFFFFF} 19.3\% & {\cellcolor[HTML]{FFF4F4}} \color[HTML]{000000} 328.7\% & {\cellcolor[HTML]{FF0000}} \color[HTML]{FFFFFF} 7204.6\% & {\cellcolor[HTML]{FFE8E8}} \color[HTML]{000000} 551.5\% \\
spiral & 312 & 80 & {\cellcolor{blue}} \color[HTML]{FFFFFF} 0.04s & {\cellcolor[HTML]{D5EAD5}} \color[HTML]{000000} 83.6\% & {\cellcolor[HTML]{FFEAEA}} \color[HTML]{000000} 502.3\% & {\cellcolor[HTML]{188C18}} \color[HTML]{FFFFFF} 9.7\% & {\cellcolor[HTML]{FFFAFA}} \color[HTML]{000000} 202.7\% & {\cellcolor[HTML]{FFD6D6}} \color[HTML]{000000} 901.6\% & {\cellcolor[HTML]{FFF3F3}} \color[HTML]{000000} 340.1\% \\
\midrule
Mean &  &  & {\cellcolor{blue}} \color[HTML]{FFFFFF} 0.06s & {\cellcolor[HTML]{D5EAD5}} \color[HTML]{000000} 83.3\% & {\cellcolor[HTML]{FFE7E7}} \color[HTML]{000000} 568.2\% & {\cellcolor[HTML]{219021}} \color[HTML]{FFFFFF} 13.1\% & {\cellcolor[HTML]{FFF8F8}} \color[HTML]{000000} 237.1\% & {\cellcolor[HTML]{FF8585}} \color[HTML]{000000} 2439.9\% & {\cellcolor[HTML]{FFEDED}} \color[HTML]{000000} 454.7\% \\
\toprule
\end{tabular}

%% file: coltables/t-cpu-fr-mod.tex
\begin{table}[ht]
    \small
    \begin{center}
    \input{coltables/cpu-fr-mod.tex}
    \caption{Modified literature problems: CPU time relative to \gkmp. \tabcom}
    \label{tab:cpu-fr-mod}
    \end{center}
\end{table}
\renewcommand{\tabcom}{}

%% file: coltables/cpu-fr-mod.tex
\begin{tabular}{lrrrrrrrrr}
\toprule
data set & n & k & \thead{t(greedy \\ km++)\\Python} & \thead{vanilla \\ km++\\Python} & \thead{better \\ km++\\ \textsf{Python}} & \thead{Hartigan- \\ Wong\\R/Fortran} & \thead{genetic \\ algorithm\\C} & \thead{random \\ swap\\C} & \thead{breathing \\ k-means\\Python} \\
\midrule
aggregation-$\ast$ & 4200 & 200 & {\cellcolor{blue}} \color[HTML]{FFFFFF} 0.15s & {\cellcolor[HTML]{8DC68D}} \color[HTML]{000000} 55.2\% & {\cellcolor[HTML]{FFC5C5}} \color[HTML]{000000} 1222.3\% & {\cellcolor[HTML]{47A347}} \color[HTML]{FFFFFF} 28.0\% & {\cellcolor[HTML]{FFE7E7}} \color[HTML]{000000} 569.8\% & {\cellcolor[HTML]{FF0000}} \color[HTML]{FFFFFF} 11926.3\% & {\cellcolor[HTML]{FFEFEF}} \color[HTML]{000000} 410.1\% \\
compound-$\ast$ & 4200 & 50 & {\cellcolor{blue}} \color[HTML]{FFFFFF} 0.05s & {\cellcolor[HTML]{B8DCB8}} \color[HTML]{000000} 72.2\% & {\cellcolor[HTML]{FFCECE}} \color[HTML]{000000} 1052.4\% & {\cellcolor[HTML]{7BBD7B}} \color[HTML]{000000} 48.1\% & {\cellcolor[HTML]{FFEAEA}} \color[HTML]{000000} 506.3\% & {\cellcolor[HTML]{FF0000}} \color[HTML]{FFFFFF} 39361.3\% & {\cellcolor[HTML]{FFE7E7}} \color[HTML]{000000} 560.5\% \\
d31-$\ast$ & 4200 & 100 & {\cellcolor{blue}} \color[HTML]{FFFFFF} 0.09s & {\cellcolor[HTML]{9ACD9A}} \color[HTML]{000000} 60.3\% & {\cellcolor[HTML]{FFCBCB}} \color[HTML]{000000} 1102.8\% & {\cellcolor[HTML]{60B060}} \color[HTML]{000000} 37.8\% & {\cellcolor[HTML]{FFE9E9}} \color[HTML]{000000} 528.9\% & {\cellcolor[HTML]{FF0000}} \color[HTML]{FFFFFF} 21590.9\% & {\cellcolor[HTML]{FFEDED}} \color[HTML]{000000} 452.2\% \\
flame-$\ast$ & 4200 & 80 & {\cellcolor{blue}} \color[HTML]{FFFFFF} 0.07s & {\cellcolor[HTML]{A7D3A7}} \color[HTML]{000000} 65.5\% & {\cellcolor[HTML]{FFCBCB}} \color[HTML]{000000} 1113.7\% & {\cellcolor[HTML]{6CB66C}} \color[HTML]{000000} 42.5\% & {\cellcolor[HTML]{FFE5E5}} \color[HTML]{000000} 611.5\% & {\cellcolor[HTML]{FF0000}} \color[HTML]{FFFFFF} 24205.8\% & {\cellcolor[HTML]{FFE4E4}} \color[HTML]{000000} 624.7\% \\
jain-$\ast$ & 4200 & 30 & {\cellcolor{blue}} \color[HTML]{FFFFFF} 0.03s & {\cellcolor[HTML]{D2E9D2}} \color[HTML]{000000} 82.1\% & {\cellcolor[HTML]{FFCFCF}} \color[HTML]{000000} 1027.7\% & {\cellcolor[HTML]{8EC78E}} \color[HTML]{000000} 55.6\% & {\cellcolor[HTML]{FFE6E6}} \color[HTML]{000000} 579.2\% & {\cellcolor[HTML]{FF0000}} \color[HTML]{FFFFFF} 75741.6\% & {\cellcolor[HTML]{FFE5E5}} \color[HTML]{000000} 611.0\% \\
pathbased-$\ast$ & 4200 & 50 & {\cellcolor{blue}} \color[HTML]{FFFFFF} 0.05s & {\cellcolor[HTML]{B2D9B2}} \color[HTML]{000000} 69.7\% & {\cellcolor[HTML]{FFCCCC}} \color[HTML]{000000} 1080.7\% & {\cellcolor[HTML]{7EBF7E}} \color[HTML]{000000} 49.6\% & {\cellcolor[HTML]{FFE9E9}} \color[HTML]{000000} 524.8\% & {\cellcolor[HTML]{FF0000}} \color[HTML]{FFFFFF} 37964.6\% & {\cellcolor[HTML]{FFEBEB}} \color[HTML]{000000} 498.0\% \\
r15-$\ast$ & 4200 & 30 & {\cellcolor{blue}} \color[HTML]{FFFFFF} 0.03s & {\cellcolor[HTML]{C5E2C5}} \color[HTML]{000000} 77.2\% & {\cellcolor[HTML]{FFCACA}} \color[HTML]{000000} 1130.3\% & {\cellcolor[HTML]{9CCE9C}} \color[HTML]{000000} 61.1\% & {\cellcolor[HTML]{FFDFDF}} \color[HTML]{000000} 722.7\% & {\cellcolor[HTML]{FF0000}} \color[HTML]{FFFFFF} 70939.1\% & {\cellcolor[HTML]{FFE4E4}} \color[HTML]{000000} 623.6\% \\
s2-$\ast$ & 4200 & 100 & {\cellcolor{blue}} \color[HTML]{FFFFFF} 0.08s & {\cellcolor[HTML]{9CCE9C}} \color[HTML]{000000} 61.1\% & {\cellcolor[HTML]{FFCACA}} \color[HTML]{000000} 1126.0\% & {\cellcolor[HTML]{60B060}} \color[HTML]{000000} 37.6\% & {\cellcolor[HTML]{FFE9E9}} \color[HTML]{000000} 536.5\% & {\cellcolor[HTML]{FF0000}} \color[HTML]{FFFFFF} 21913.8\% & {\cellcolor[HTML]{FFECEC}} \color[HTML]{000000} 468.0\% \\
spiral-$\ast$ & 4200 & 80 & {\cellcolor{blue}} \color[HTML]{FFFFFF} 0.07s & {\cellcolor[HTML]{99CC99}} \color[HTML]{000000} 60.0\% & {\cellcolor[HTML]{FFCCCC}} \color[HTML]{000000} 1080.8\% & {\cellcolor[HTML]{67B367}} \color[HTML]{000000} 40.4\% & {\cellcolor[HTML]{FFE5E5}} \color[HTML]{000000} 603.4\% & {\cellcolor[HTML]{FF0000}} \color[HTML]{FFFFFF} 23579.3\% & {\cellcolor[HTML]{FFEDED}} \color[HTML]{000000} 460.5\% \\
\midrule
Mean &  &  & {\cellcolor{blue}} \color[HTML]{FFFFFF} 0.07s & {\cellcolor[HTML]{ABD5AB}} \color[HTML]{000000} 67.0\% & {\cellcolor[HTML]{FFCBCB}} \color[HTML]{000000} 1104.1\% & {\cellcolor[HTML]{71B871}} \color[HTML]{000000} 44.5\% & {\cellcolor[HTML]{FFE7E7}} \color[HTML]{000000} 575.9\% & {\cellcolor[HTML]{FF0000}} \color[HTML]{FFFFFF} 36358.1\% & {\cellcolor[HTML]{FFE9E9}} \color[HTML]{000000} 523.2\% \\
\toprule
\end{tabular}

%% file: coltables/t-cpu-espiralX.tex
\begin{table}[ht]
    \small
    \begin{center}
    \input{coltables/cpu-espiralX.tex}
    \caption{"Evil Spiral" problems: CPU time relative to \gkmp. \tabcom}
    \label{tab:cpu-espiralX}
    \end{center}
\end{table}
\renewcommand{\tabcom}{}

%% file: coltables/cpu-espiralX.tex
\begin{tabular}{lrrrrrrrrr}
\toprule
data set & n & k & \thead{t(greedy \\ km++)\\Python} & \thead{vanilla \\ km++\\Python} & \thead{better \\ km++\\ \textsf{Python}} & \thead{Hartigan- \\ Wong\\R/Fortran} & \thead{genetic \\ algorithm\\C} & \thead{random \\ swap\\C} & \thead{breathing \\ k-means\\Python} \\
\midrule
espiral01 & 1000 & 100 & {\cellcolor{blue}} \color[HTML]{FFFFFF} 0.05s & {\cellcolor[HTML]{CCE6CC}} \color[HTML]{000000} 80.1\% & {\cellcolor[HTML]{FFE1E1}} \color[HTML]{000000} 678.5\% & {\cellcolor[HTML]{1E8F1E}} \color[HTML]{FFFFFF} 11.9\% & {\cellcolor[HTML]{FFF7F7}} \color[HTML]{000000} 254.2\% & {\cellcolor[HTML]{FFA3A3}} \color[HTML]{000000} 1874.6\% & {\cellcolor[HTML]{FFEEEE}} \color[HTML]{000000} 428.9\% \\
espiral04 & 2500 & 100 & {\cellcolor{blue}} \color[HTML]{FFFFFF} 0.08s & {\cellcolor[HTML]{A8D4A8}} \color[HTML]{000000} 65.7\% & {\cellcolor[HTML]{FFD9D9}} \color[HTML]{000000} 838.6\% & {\cellcolor[HTML]{289428}} \color[HTML]{FFFFFF} 15.8\% & {\cellcolor[HTML]{FFF6F6}} \color[HTML]{000000} 284.9\% & {\cellcolor[HTML]{FF8585}} \color[HTML]{000000} 2437.0\% & {\cellcolor[HTML]{FFEDED}} \color[HTML]{000000} 457.7\% \\
espiral07 & 4000 & 100 & {\cellcolor{blue}} \color[HTML]{FFFFFF} 0.09s & {\cellcolor[HTML]{A4D2A4}} \color[HTML]{000000} 64.3\% & {\cellcolor[HTML]{FFCDCD}} \color[HTML]{000000} 1068.8\% & {\cellcolor[HTML]{369B36}} \color[HTML]{FFFFFF} 21.3\% & {\cellcolor[HTML]{FFF2F2}} \color[HTML]{000000} 351.3\% & {\cellcolor[HTML]{FF6464}} \color[HTML]{000000} 3077.9\% & {\cellcolor[HTML]{FFE8E8}} \color[HTML]{000000} 548.2\% \\
espiral10 & 5500 & 100 & {\cellcolor{blue}} \color[HTML]{FFFFFF} 0.10s & {\cellcolor[HTML]{9CCE9C}} \color[HTML]{000000} 61.1\% & {\cellcolor[HTML]{FFC4C4}} \color[HTML]{000000} 1243.7\% & {\cellcolor[HTML]{3E9F3E}} \color[HTML]{FFFFFF} 24.4\% & {\cellcolor[HTML]{FFF0F0}} \color[HTML]{000000} 399.0\% & {\cellcolor[HTML]{FF4E4E}} \color[HTML]{000000} 3495.4\% & {\cellcolor[HTML]{FFE7E7}} \color[HTML]{000000} 561.7\% \\
espiral13 & 7000 & 100 & {\cellcolor{blue}} \color[HTML]{FFFFFF} 0.10s & {\cellcolor[HTML]{A4D2A4}} \color[HTML]{000000} 64.2\% & {\cellcolor[HTML]{FFBABA}} \color[HTML]{000000} 1437.9\% & {\cellcolor[HTML]{48A448}} \color[HTML]{FFFFFF} 28.2\% & {\cellcolor[HTML]{FFECEC}} \color[HTML]{000000} 465.0\% & {\cellcolor[HTML]{FF3131}} \color[HTML]{FFFFFF} 4046.0\% & {\cellcolor[HTML]{FFE4E4}} \color[HTML]{000000} 620.4\% \\
espiral16 & 8500 & 100 & {\cellcolor{blue}} \color[HTML]{FFFFFF} 0.10s & {\cellcolor[HTML]{AFD7AF}} \color[HTML]{000000} 68.6\% & {\cellcolor[HTML]{FFACAC}} \color[HTML]{000000} 1707.7\% & {\cellcolor[HTML]{54AA54}} \color[HTML]{000000} 32.9\% & {\cellcolor[HTML]{FFE8E8}} \color[HTML]{000000} 553.5\% & {\cellcolor[HTML]{FF0C0C}} \color[HTML]{FFFFFF} 4754.2\% & {\cellcolor[HTML]{FFE1E1}} \color[HTML]{000000} 678.7\% \\
espiral19 & 10000 & 100 & {\cellcolor{blue}} \color[HTML]{FFFFFF} 0.11s & {\cellcolor[HTML]{A2D1A2}} \color[HTML]{000000} 63.4\% & {\cellcolor[HTML]{FFA7A7}} \color[HTML]{000000} 1787.6\% & {\cellcolor[HTML]{55AA55}} \color[HTML]{000000} 33.3\% & {\cellcolor[HTML]{FFE7E7}} \color[HTML]{000000} 575.2\% & {\cellcolor[HTML]{FF0404}} \color[HTML]{FFFFFF} 4920.0\% & {\cellcolor[HTML]{FFE1E1}} \color[HTML]{000000} 679.0\% \\
espiral22 & 11500 & 100 & {\cellcolor{blue}} \color[HTML]{FFFFFF} 0.12s & {\cellcolor[HTML]{A6D3A6}} \color[HTML]{000000} 65.0\% & {\cellcolor[HTML]{FFA2A2}} \color[HTML]{000000} 1885.9\% & {\cellcolor[HTML]{58AC58}} \color[HTML]{000000} 34.6\% & {\cellcolor[HTML]{FFE5E5}} \color[HTML]{000000} 601.4\% & {\cellcolor[HTML]{FF0000}} \color[HTML]{FFFFFF} 5098.8\% & {\cellcolor[HTML]{FFDEDE}} \color[HTML]{000000} 748.0\% \\
espiral25 & 13000 & 100 & {\cellcolor{blue}} \color[HTML]{FFFFFF} 0.14s & {\cellcolor[HTML]{9FCF9F}} \color[HTML]{000000} 62.4\% & {\cellcolor[HTML]{FFA0A0}} \color[HTML]{000000} 1936.9\% & {\cellcolor[HTML]{58AC58}} \color[HTML]{000000} 34.5\% & {\cellcolor[HTML]{FFE5E5}} \color[HTML]{000000} 605.9\% & {\cellcolor[HTML]{FF0000}} \color[HTML]{FFFFFF} 5011.3\% & {\cellcolor[HTML]{FFDFDF}} \color[HTML]{000000} 725.1\% \\
\midrule
Mean &  &  & {\cellcolor{blue}} \color[HTML]{FFFFFF} 0.10s & {\cellcolor[HTML]{A9D4A9}} \color[HTML]{000000} 66.1\% & {\cellcolor[HTML]{FFBCBC}} \color[HTML]{000000} 1398.4\% & {\cellcolor[HTML]{43A143}} \color[HTML]{FFFFFF} 26.3\% & {\cellcolor[HTML]{FFEDED}} \color[HTML]{000000} 454.5\% & {\cellcolor[HTML]{FF3B3B}} \color[HTML]{FFFFFF} 3857.2\% & {\cellcolor[HTML]{FFE5E5}} \color[HTML]{000000} 605.3\% \\
\toprule
\end{tabular}

%% file: coltables/t-cpu-high-D.tex
\begin{table}[ht]
    \small
    \begin{center}
    \input{coltables/cpu-high-D.tex}
    \caption{High-dimensional problems: CPU time relative to \gkmp. \tabcom}
    \label{tab:cpu-high-D}
    \end{center}
\end{table}
\renewcommand{\tabcom}{}

%% file: coltables/cpu-high-D.tex
\begin{tabular}{lrrrrrrrrr}
\toprule
data set & n & k & \thead{t(greedy \\ km++)\\Python} & \thead{vanilla \\ km++\\Python} & \thead{better \\ km++\\ \textsf{Python}} & \thead{Hartigan- \\ Wong\\R/Fortran} & \thead{genetic \\ algorithm\\C} & \thead{random \\ swap\\C} & \thead{breathing \\ k-means\\Python} \\
\midrule
Norm25 & 10000 & 10 & {\cellcolor{blue}} \color[HTML]{FFFFFF} 0.02s & {\cellcolor[HTML]{E1F0E1}} \color[HTML]{000000} 88.1\% & {\cellcolor[HTML]{FFC2C2}} \color[HTML]{000000} 1274.7\% & {\cellcolor[HTML]{C2E1C2}} \color[HTML]{000000} 76.0\% & {\cellcolor[HTML]{FF7171}} \color[HTML]{000000} 2826.7\% & {\cellcolor[HTML]{FF0000}} \color[HTML]{FFFFFF} 96394.3\% & {\cellcolor[HTML]{FFC5C5}} \color[HTML]{000000} 1220.0\% \\
Norm25 & 10000 & 25 & {\cellcolor{blue}} \color[HTML]{FFFFFF} 0.04s & {\cellcolor[HTML]{BCDEBC}} \color[HTML]{000000} 73.7\% & {\cellcolor[HTML]{FFB5B5}} \color[HTML]{000000} 1523.8\% & {\cellcolor[HTML]{E1F0E1}} \color[HTML]{000000} 88.0\% & {\cellcolor[HTML]{FFB8B8}} \color[HTML]{000000} 1470.6\% & {\cellcolor[HTML]{FF0000}} \color[HTML]{FFFFFF} 61781.7\% & {\cellcolor[HTML]{FFE8E8}} \color[HTML]{000000} 542.4\% \\
Norm25 & 10000 & 50 & {\cellcolor{blue}} \color[HTML]{FFFFFF} 0.06s & {\cellcolor[HTML]{BBDDBB}} \color[HTML]{000000} 73.4\% & {\cellcolor[HTML]{FFADAD}} \color[HTML]{000000} 1679.4\% & {\cellcolor[HTML]{EFF7EF}} \color[HTML]{000000} 93.5\% & {\cellcolor[HTML]{FFCACA}} \color[HTML]{000000} 1126.8\% & {\cellcolor[HTML]{FF0000}} \color[HTML]{FFFFFF} 70495.7\% & {\cellcolor[HTML]{FFC6C6}} \color[HTML]{000000} 1206.8\% \\
Norm25 & 10000 & 100 & {\cellcolor{blue}} \color[HTML]{FFFFFF} 0.14s & {\cellcolor[HTML]{AFD7AF}} \color[HTML]{000000} 68.4\% & {\cellcolor[HTML]{FFB4B4}} \color[HTML]{000000} 1539.3\% & {\cellcolor[HTML]{A9D4A9}} \color[HTML]{000000} 66.3\% & {\cellcolor[HTML]{FFD6D6}} \color[HTML]{000000} 890.5\% & {\cellcolor[HTML]{FF0000}} \color[HTML]{FFFFFF} 53017.9\% & {\cellcolor[HTML]{FFD3D3}} \color[HTML]{000000} 953.9\% \\
Norm25 & 10000 & 200 & {\cellcolor{blue}} \color[HTML]{FFFFFF} 0.33s & {\cellcolor[HTML]{A4D2A4}} \color[HTML]{000000} 64.3\% & {\cellcolor[HTML]{FFC3C3}} \color[HTML]{000000} 1253.8\% & {\cellcolor[HTML]{6CB66C}} \color[HTML]{000000} 42.5\% & {\cellcolor[HTML]{FFD9D9}} \color[HTML]{000000} 837.6\% & {\cellcolor[HTML]{FF0000}} \color[HTML]{FFFFFF} 24060.5\% & {\cellcolor[HTML]{FFD4D4}} \color[HTML]{000000} 929.2\% \\
Cloud & 1024 & 10 & {\cellcolor{blue}} \color[HTML]{FFFFFF} 0.01s & {\cellcolor[HTML]{FFFFFF}} \color[HTML]{000000} 117.2\% & {\cellcolor[HTML]{FFECEC}} \color[HTML]{000000} 472.1\% & {\cellcolor[HTML]{82C182}} \color[HTML]{000000} 51.1\% & {\cellcolor[HTML]{FFF2F2}} \color[HTML]{000000} 363.7\% & {\cellcolor[HTML]{FF0000}} \color[HTML]{FFFFFF} 19053.3\% & {\cellcolor[HTML]{FFD8D8}} \color[HTML]{000000} 854.1\% \\
Cloud & 1024 & 25 & {\cellcolor{blue}} \color[HTML]{FFFFFF} 0.02s & {\cellcolor[HTML]{FAFDFA}} \color[HTML]{000000} 97.8\% & {\cellcolor[HTML]{FFEAEA}} \color[HTML]{000000} 511.1\% & {\cellcolor[HTML]{50A850}} \color[HTML]{000000} 31.3\% & {\cellcolor[HTML]{FFF5F5}} \color[HTML]{000000} 307.3\% & {\cellcolor[HTML]{FF0000}} \color[HTML]{FFFFFF} 11720.1\% & {\cellcolor[HTML]{FFE6E6}} \color[HTML]{000000} 579.6\% \\
Cloud & 1024 & 50 & {\cellcolor{blue}} \color[HTML]{FFFFFF} 0.03s & {\cellcolor[HTML]{E0F0E0}} \color[HTML]{000000} 87.8\% & {\cellcolor[HTML]{FFE3E3}} \color[HTML]{000000} 650.0\% & {\cellcolor[HTML]{43A143}} \color[HTML]{FFFFFF} 26.5\% & {\cellcolor[HTML]{FFF1F1}} \color[HTML]{000000} 374.5\% & {\cellcolor[HTML]{FF0000}} \color[HTML]{FFFFFF} 8589.4\% & {\cellcolor[HTML]{FFE7E7}} \color[HTML]{000000} 572.7\% \\
Cloud & 1024 & 100 & {\cellcolor{blue}} \color[HTML]{FFFFFF} 0.07s & {\cellcolor[HTML]{C3E1C3}} \color[HTML]{000000} 76.4\% & {\cellcolor[HTML]{FFE7E7}} \color[HTML]{000000} 576.7\% & {\cellcolor[HTML]{309830}} \color[HTML]{FFFFFF} 18.8\% & {\cellcolor[HTML]{FFEFEF}} \color[HTML]{000000} 415.4\% & {\cellcolor[HTML]{FF3232}} \color[HTML]{FFFFFF} 4034.6\% & {\cellcolor[HTML]{FFEAEA}} \color[HTML]{000000} 508.4\% \\
Cloud & 1024 & 200 & {\cellcolor{blue}} \color[HTML]{FFFFFF} 0.12s & {\cellcolor[HTML]{BCDEBC}} \color[HTML]{000000} 73.7\% & {\cellcolor[HTML]{FFE6E6}} \color[HTML]{000000} 581.8\% & {\cellcolor[HTML]{279327}} \color[HTML]{FFFFFF} 15.5\% & {\cellcolor[HTML]{FFE8E8}} \color[HTML]{000000} 557.8\% & {\cellcolor[HTML]{FF9E9E}} \color[HTML]{000000} 1966.7\% & {\cellcolor[HTML]{FFE6E6}} \color[HTML]{000000} 585.1\% \\
Spam & 4601 & 10 & {\cellcolor{blue}} \color[HTML]{FFFFFF} 0.03s & {\cellcolor[HTML]{FFFFFF}} \color[HTML]{000000} 100.4\% & {\cellcolor[HTML]{FFE9E9}} \color[HTML]{000000} 534.5\% & {\cellcolor[HTML]{FFECEC}} \color[HTML]{000000} 481.8\% & {\cellcolor[HTML]{FF7171}} \color[HTML]{000000} 2835.9\% & {\cellcolor[HTML]{FF0000}} \color[HTML]{FFFFFF} 210704.4\% & {\cellcolor[HTML]{FFCCCC}} \color[HTML]{000000} 1076.4\% \\
Spam & 4601 & 25 & {\cellcolor{blue}} \color[HTML]{FFFFFF} 0.05s & {\cellcolor[HTML]{FFFFFF}} \color[HTML]{000000} 112.4\% & {\cellcolor[HTML]{FFE3E3}} \color[HTML]{000000} 648.8\% & {\cellcolor[HTML]{FFDDDD}} \color[HTML]{000000} 760.3\% & {\cellcolor[HTML]{FF7F7F}} \color[HTML]{000000} 2555.1\% & {\cellcolor[HTML]{FF0000}} \color[HTML]{FFFFFF} 156117.9\% & {\cellcolor[HTML]{FFDBDB}} \color[HTML]{000000} 797.9\% \\
Spam & 4601 & 50 & {\cellcolor{blue}} \color[HTML]{FFFFFF} 0.08s & {\cellcolor[HTML]{E8F4E8}} \color[HTML]{000000} 90.8\% & {\cellcolor[HTML]{FFDEDE}} \color[HTML]{000000} 737.9\% & {\cellcolor[HTML]{FFE6E6}} \color[HTML]{000000} 594.0\% & {\cellcolor[HTML]{FF7070}} \color[HTML]{000000} 2838.4\% & {\cellcolor[HTML]{FF0000}} \color[HTML]{FFFFFF} 149860.6\% & {\cellcolor[HTML]{FFDDDD}} \color[HTML]{000000} 760.4\% \\
Spam & 4601 & 100 & {\cellcolor{blue}} \color[HTML]{FFFFFF} 0.14s & {\cellcolor[HTML]{D7EBD7}} \color[HTML]{000000} 84.1\% & {\cellcolor[HTML]{FFDBDB}} \color[HTML]{000000} 801.3\% & {\cellcolor[HTML]{FFF5F5}} \color[HTML]{000000} 308.5\% & {\cellcolor[HTML]{FF6262}} \color[HTML]{000000} 3115.1\% & {\cellcolor[HTML]{FF0000}} \color[HTML]{FFFFFF} 119787.8\% & {\cellcolor[HTML]{FFD8D8}} \color[HTML]{000000} 848.3\% \\
Spam & 4601 & 200 & {\cellcolor{blue}} \color[HTML]{FFFFFF} 0.28s & {\cellcolor[HTML]{BADDBA}} \color[HTML]{000000} 72.8\% & {\cellcolor[HTML]{FFDDDD}} \color[HTML]{000000} 760.7\% & {\cellcolor[HTML]{FFF6F6}} \color[HTML]{000000} 279.3\% & {\cellcolor[HTML]{FF6565}} \color[HTML]{000000} 3057.3\% & {\cellcolor[HTML]{FF0000}} \color[HTML]{FFFFFF} 57533.6\% & {\cellcolor[HTML]{FFDDDD}} \color[HTML]{000000} 754.5\% \\
\midrule
Mean &  &  & {\cellcolor{blue}} \color[HTML]{FFFFFF} 0.10s & {\cellcolor[HTML]{DAEDDA}} \color[HTML]{000000} 85.4\% & {\cellcolor[HTML]{FFD6D6}} \color[HTML]{000000} 903.1\% & {\cellcolor[HTML]{FFFBFB}} \color[HTML]{000000} 195.6\% & {\cellcolor[HTML]{FFB3B3}} \color[HTML]{000000} 1571.5\% & {\cellcolor[HTML]{FF0000}} \color[HTML]{FFFFFF} 69674.6\% & {\cellcolor[HTML]{FFDADA}} \color[HTML]{000000} 812.7\% \\
\toprule
\end{tabular}

%% file: bkmshort.bbl
\begin{thebibliography}{37}
\providecommand{\natexlab}[1]{#1}
\providecommand{\url}[1]{\texttt{#1}}
\expandafter\ifx\csname urlstyle\endcsname\relax
  \providecommand{\doi}[1]{doi: #1}\else
  \providecommand{\doi}{doi: \begingroup \urlstyle{rm}\Url}\fi

\bibitem[Al-Sultan(1995)]{AlSultan1995}
Khaled~S. Al-Sultan.
\newblock A tabu search approach to the clustering problem.
\newblock \emph{Pattern Recognition}, 28, 1995.
\newblock ISSN 00313203.
\newblock \doi{10.1016/0031-3203(95)00022-R}.

\bibitem[Aloise et~al.(2009)Aloise, Deshpande, Hansen, and Popat]{Aloise2009}
Daniel Aloise, Amit Deshpande, Pierre Hansen, and Preyas Popat.
\newblock Np-hardness of euclidean sum-of-squares clustering.
\newblock \emph{Machine Learning}, 75:\penalty0 245--248, 5 2009.
\newblock ISSN 08856125.
\newblock \doi{10.1007/s10994-009-5103-0}.

\bibitem[Arthur and Vassilvitskii(2007)]{Arthur2007}
David Arthur and Sergei Vassilvitskii.
\newblock k-means++: the advantages of careful seeding.
\newblock pages 1027--1035, 2007.
\newblock ISBN 9780898716245.
\newblock \doi{10.1145/1283383.1283494}.

\bibitem[Bhattacharya et~al.(2020)Bhattacharya, Eube, Röglin, and
  Schmidt]{DBLP:conf/esa/BhattacharyaER020}
Anup Bhattacharya, Jan Eube, Heiko Röglin, and Melanie Schmidt.
\newblock Noisy, greedy and not so greedy k-means++.
\newblock volume 173, pages 18:1--18:21. Schloss Dagstuhl - Leibniz-Zentrum
  für Informatik, 2020.
\newblock \doi{10.4230/LIPIcs.ESA.2020.18}.
\newblock URL \url{https://doi.org/10.4230/LIPIcs.ESA.2020.18}.

\bibitem[Bradley and Fayyad(1998)]{Bradley1998}
P.S. Bradley and U.~Fayyad.
\newblock Refining initial points for k-means clustering.
\newblock pages 91--99, 1998.

\bibitem[Chang and Yeung(2008)]{Chang2008}
Hong Chang and Dit~Yan Yeung.
\newblock Robust path-based spectral clustering.
\newblock \emph{Pattern Recognition}, 41:\penalty0 191--203, 2008.
\newblock ISSN 00313203.
\newblock \doi{10.1016/j.patcog.2007.04.010}.

\bibitem[Dua and Graff(2017)]{Dua:2019}
Dheeru Dua and Casey Graff.
\newblock Uci machine learning repository, 2017.
\newblock URL \url{http://archive.ics.uci.edu/ml}.

\bibitem[Equitz(1989)]{Equitz1989}
William~H. Equitz.
\newblock A new vector quantization clustering algorithm.
\newblock \emph{IEEE Transactions on Acoustics, Speech, and Signal Processing},
  37, 1989.
\newblock ISSN 00963518.
\newblock \doi{10.1109/29.35395}.

\bibitem[Forgy(1965)]{Forgy65}
Edward Forgy.
\newblock Cluster analysis of multivariate data: efficiency vs.
  interpretability of classifications.
\newblock volume~21, page 768, 1965.
\newblock abstract.

\bibitem[Fritzke(1993)]{Fritzke93}
Bernd Fritzke.
\newblock Vector quantization with a growing and splitting elastic net.
\newblock pages 580--585. Springer, 1993.

\bibitem[Fritzke(1994)]{Fritzke94c}
Bernd Fritzke.
\newblock Growing cell structures-a self-organizing network for unsupervised
  and supervised learning.
\newblock \emph{Neural Networks}, 7:\penalty0 1441--1460, 1 1994.
\newblock ISSN 08936080.
\newblock \doi{10.1016/0893-6080(94)90091-4}.

\bibitem[Fritzke(1995)]{Fritzke95}
Bernd Fritzke.
\newblock A growing neural gas network learns topologies.
\newblock pages 625--632. MIT Press, 1995.

\bibitem[Fritzke(1997)]{Fritzke1997}
Bernd Fritzke.
\newblock The lbg-u method for vector quantization -- an improvement over lbg
  inspired from neural networks.
\newblock \emph{Neural Processing Letters}, 5:\penalty0 35--45, 1997.
\newblock ISSN 13704621.

\bibitem[Fränti(2000)]{Franti2000}
Pasi Fränti.
\newblock Genetic algorithm with deterministic crossover for vector
  quantization.
\newblock \emph{Pattern Recognition Letters}, 21, 2000.
\newblock ISSN 01678655.
\newblock \doi{10.1016/S0167-8655(99)00133-6}.

\bibitem[Fränti(2018)]{Franti2018}
Pasi Fränti.
\newblock Efficiency of random swap clustering.
\newblock \emph{Journal of Big Data}, 5, 2018.
\newblock ISSN 21961115.
\newblock \doi{10.1186/s40537-018-0122-y}.

\bibitem[Fränti and Sieranoja(2019)]{Franti2019b}
Pasi Fränti and Sami Sieranoja.
\newblock How much can k-means be improved by using better initialization and
  repeats?
\newblock \emph{Pattern Recognition}, 93:\penalty0 95--112, 9 2019.
\newblock ISSN 00313203.
\newblock \doi{10.1016/j.patcog.2019.04.014}.

\bibitem[Fränti and Virmajoki(2006)]{Franti2006}
Pasi Fränti and Olli Virmajoki.
\newblock Iterative shrinking method for clustering problems.
\newblock \emph{Pattern Recognition}, 39:\penalty0 761--775, 2006.
\newblock ISSN 00313203.
\newblock \doi{10.1016/j.patcog.2005.09.012}.

\bibitem[Fränti et~al.(1997)Fränti, Kaukoranta, and Nevalainen]{Franti1997}
Pasi Fränti, Timo Kaukoranta, and Olli Nevalainen.
\newblock On the splitting method for vector quantization codebook generation.
\newblock \emph{Optical Engineering}, 36, 1997.
\newblock ISSN 0091-3286.
\newblock \doi{10.1117/1.601531}.

\bibitem[Fränti et~al.(1998)Fränti, Kivijärvi, and Nevalainen]{Franti1998}
Pasi Fränti, Juha Kivijärvi, and Olli Nevalainen.
\newblock Tabu search algorithm for codebook generation in vector quantization.
\newblock \emph{Pattern Recognition}, 31, 1998.
\newblock ISSN 00313203.
\newblock \doi{10.1016/S0031-3203(97)00127-1}.

\bibitem[Fu and Medico(2007)]{Fu2007}
Limin Fu and Enzo Medico.
\newblock Flame, a novel fuzzy clustering method for the analysis of dna
  microarray data.
\newblock \emph{BMC Bioinformatics}, 8, 2007.
\newblock ISSN 14712105.
\newblock \doi{10.1186/1471-2105-8-3}.

\bibitem[Gionis et~al.(2007)Gionis, Mannila, and Tsaparas]{Gionis2007}
Aristides Gionis, Heikki Mannila, and Panayiotis Tsaparas.
\newblock Clustering aggregation.
\newblock \emph{ACM Transactions on Knowledge Discovery from Data}, 1, 2007.
\newblock ISSN 15564681.
\newblock \doi{10.1145/1217299.1217303}.

\bibitem[Gonzalez(1985)]{Gonzales1985}
Teofilo~F. Gonzalez.
\newblock Clustering to minimize the maximum intercluster distance.
\newblock \emph{Theoretical Computer Science}, 38, 1985.
\newblock ISSN 03043975.
\newblock \doi{10.1016/0304-3975(85)90224-5}.

\bibitem[Hartigan and Wong(1979)]{Hartigan1979}
J.A. Hartigan and M.A. Wong.
\newblock Algorithm as 136: A k-means clustering algorithm.
\newblock \emph{Journal of the Royal Statistical Society, Series C},
  28:\penalty0 100--108, 1979.

\bibitem[Jain and Law(2005)]{Jain2005}
Anil~K. Jain and Martin~H.C. Law.
\newblock Data clustering: A user's dilemma, 2005.
\newblock ISSN 16113349.

\bibitem[Kaukoranta et~al.(1998)Kaukoranta, Fränti, and
  Nevalainen]{Kaukoranta1998}
Timo Kaukoranta, Pasi Fränti, and Olli Nevalainen.
\newblock Iterative split-and-merge algorithm for vq codebook generation.
\newblock \emph{Optical Engineering}, 37:\penalty0 2726--2732, 1998.
\newblock ISSN 00913286.

\bibitem[Lattanzi and Sohler(2019)]{Lattanzi2019}
Silvio Lattanzi and Christian Sohler.
\newblock A better k-means++ algorithm via local search.
\newblock 2019.

\bibitem[Likas et~al.(2003)Likas, Vlassis, and Verbeek]{Likas2003}
Aristidis Likas, Nikos Vlassis, and Jakob~J. Verbeek.
\newblock The global k-means clustering algorithm.
\newblock \emph{Pattern Recognition}, 36:\penalty0 451--461, 2 2003.
\newblock ISSN 00313203.
\newblock \doi{10.1016/S0031-3203(02)00060-2}.

\bibitem[Linde et~al.(1980)Linde, Buzo, and Gray]{Linde1980}
Yoseph Linde, Andres Buzo, and Robert~M. Gray.
\newblock An algorithm for vector quantizer design.
\newblock \emph{IEEE Transactions on Communications}, 28:\penalty0 84--95,
  1980.
\newblock ISSN 00906778.
\newblock \doi{10.1109/TCOM.1980.1094577}.

\bibitem[Lloyd(1982)]{Lloyd1982}
Stuart~P. Lloyd.
\newblock Least squares quantization in pcm.
\newblock \emph{IEEE Transactions on Information Theory}, 28:\penalty0
  129--137, 1982.
\newblock ISSN 15579654.
\newblock \doi{10.1109/TIT.1982.1056489}.

\bibitem[MacQueen(1967)]{MacQueen1967}
J~B MacQueen.
\newblock Kmeans and analysis of multivariate observations.
\newblock pages 281--297, 1967.

\bibitem[Pedregosa et~al.(2011)Pedregosa, Varoquaux, Gramfort, Michel, Thirion,
  Grisel, Blondel, Prettenhofer, Weiss, Dubourg, Vanderplas, Passos,
  Cournapeau, Brucher, Perrot, and Édouard Duchesnay]{scikit}
Fabian Pedregosa, Gael Varoquaux, Alexandre Gramfort, Vincent Michel, Bertrand
  Thirion, Olivier Grisel, Mathieu Blondel, Peter Prettenhofer, Ron Weiss,
  Vincent Dubourg, Jake Vanderplas, Alexandre Passos, David Cournapeau,
  Matthieu Brucher, Matthieu Perrot, and Édouard Duchesnay.
\newblock Scikit-learn: Machine learning in python.
\newblock \emph{Journal of Machine Learning Research}, 12:\penalty0 2825--2830,
  10 2011.
\newblock ISSN 15324435.

\bibitem[\{R Core Team\}(2019)]{RCore2019}
\{R Core Team\}.
\newblock R: A language and environment for statistical computing., 2019.

\bibitem[Selim and Ismail(1984)]{Selim1984}
Shokri~Z. Selim and M.~A. Ismail.
\newblock K-means-type algorithms: A generalized convergence theorem and
  characterization of local optimality.
\newblock \emph{IEEE Transactions on Pattern Analysis and Machine
  Intelligence}, 6:\penalty0 81--87, 1984.
\newblock ISSN 01628828.
\newblock \doi{10.1109/TPAMI.1984.4767478}.

\bibitem[Steinbach et~al.(2000)Steinbach, Karypis, and Kumar]{steinbach2000}
Michael Steinbach, George Karypis, and Vipin Kumar.
\newblock A comparison of document clustering techniques.
\newblock \emph{IN KDD WORKSHOP ON TEXT MINING}, 2000.
\newblock URL
  \url{http://citeseer.ist.psu.edu/viewdoc/summary?doi=10.1.1.125.9225}.

\bibitem[Telgarsky and Vattani(2010)]{TelgarskyVattani2010}
Matus Telgarsky and Andrea Vattani.
\newblock Hartigan's method: k-means clustering without voronoi.
\newblock volume~9, pages 820--827. JMLR Workshop and Conference Proceedings, 3
  2010.

\bibitem[Veenman et~al.(2002)Veenman, Reinders, and Backer]{Veenman2002}
Cor~J. Veenman, Marcel~J.T. Reinders, and Eric Backer.
\newblock A maximum variance cluster algorithm.
\newblock \emph{IEEE Transactions on Pattern Analysis and Machine
  Intelligence}, 24:\penalty0 1273--1280, 2002.
\newblock ISSN 01628828.
\newblock \doi{10.1109/TPAMI.2002.1033218}.

\bibitem[Zahn(1971)]{Zahn1971}
Charles~T. Zahn.
\newblock Graph-theoretical methods for detecting and describing gestalt
  clusters.
\newblock \emph{IEEE Transactions on Computers}, C-20, 1971.
\newblock ISSN 00189340.
\newblock \doi{10.1109/T-C.1971.223083}.

\end{thebibliography}
